\pdfoutput=1

\documentclass[11pt]{article}

\IfFileExists{acl.sty}{%
  \usepackage[preprint]{acl}%
}{%
  \usepackage[margin=0.75in]{geometry}%
  \usepackage[numbers]{natbib}%
  \twocolumn%
}

\usepackage{times}
\usepackage{latexsym}

\usepackage[T1]{fontenc}

\usepackage[utf8]{inputenc}
\usepackage{xcolor}   
\usepackage{ulem}    

\usepackage{microtype}

\usepackage{graphicx}

\usepackage{graphicx}
\usepackage{color}
\usepackage{booktabs}
\usepackage{amssymb, amsmath}
\usepackage{multicol}
\usepackage{multirow}
\usepackage{hyperref}
\usepackage{colortbl}
\usepackage{rotating}
\usepackage{float}
\usepackage[caption=false,font=footnotesize]{subfig}
\usepackage{makecell}
\usepackage{xcolor}
\usepackage[most]{tcolorbox}
\usepackage[capitalize]{cleveref}
\usepackage{array}
\usepackage{enumitem}
\usepackage{longtable}
\usepackage{url}
\usepackage{tabularx}

\definecolor{heatE}{HTML}{E6F1FB}
\definecolor{heatA}{HTML}{B5D4F4}
\definecolor{heatB}{HTML}{378ADD}
\definecolor{heatC}{HTML}{185FA5}
\definecolor{heatD}{HTML}{0C447C}
\definecolor{dPos0}{HTML}{D9F0E4}
\definecolor{dPos1}{HTML}{9FE1CB}
\definecolor{dPos2}{HTML}{378ADD}
\definecolor{dPos3}{HTML}{0C447C}
\definecolor{dNegMid}{HTML}{E08080}
\definecolor{dNegLite}{HTML}{F4C8C8}
\definecolor{dGray}{HTML}{EEEEEE}
\definecolor{famFloor}{HTML}{D3D1C7}
\definecolor{famRecency}{HTML}{FAC775}
\definecolor{famFull}{HTML}{185FA5}
\definecolor{famRet}{HTML}{85B7EB}
\definecolor{famComp}{HTML}{AFA9EC}
\definecolor{famMem}{HTML}{5DCAA5}

\graphicspath{{../figures/new_figures/}{../figures/}{new_figures/}}

\title{ClinTraceBench: Source-Verifiable Longitudinal Clinical\\Reasoning over EHR-Derived Dialogues}

\author{
  Huimin Wang\textsuperscript{1} \quad
  Zhengyi Zhao\textsuperscript{2} \quad
  Yutian Zhao\textsuperscript{3} \thanks{\ Corresponding author.}\\
  \textsuperscript{1}Shenzhen University \quad
  \textsuperscript{2}The Chinese University of Hong Kong \quad
  \textsuperscript{3} Dealism \\
  \texttt{wanghm520@gmail.com} \quad
  \texttt{zyzhao@se.cuhk.edu.hk} \quad
  \texttt{rosezhao929@gmail.com}
}

\begin{document}
\maketitle

\begin{abstract}
Clinical LLM assistants must reason over multi-visit patient trajectories, 
yet whether the compact history representations used to scale them---retrieval, structured timelines, 
LLM summaries, agentic memory---preserve the longitudinal signal clinical reasoning needs has not been measured. 
We introduce ClinTraceBench: 385 MIMIC-IV-derived verified dialogues with event-ID provenance, 
a nine-task taxonomy (T1--T9), and L0--L4 deterministic + L5 human-audit validation (98.92\% agreement). 
We evaluate eight history representation strategies---a no-context floor, \textit{last-visit-only}, 
\textit{full-context}, BGE-M3 \textit{dense-retrieval}, two compression schemes, and two agentic-memory 
systems (\textit{Mem0}, \textit{A-Mem})---across four backbones (DeepSeek-V3, GPT-4o-mini, Haiku~4.5, 
Sonnet~4.6) on 6{,}271 questions: 32 cells, 200{,}672 predictions. 
Four findings: (SP4) a controlled T3 injection probe isolates compression-induced \textit{relation} 
loss---with the attribution sentence present \textit{before} construction, \textit{Mem0}, \textit{A-Mem} and 
\textit{llm-summary} still recover only 0--5.3\% of the injected positives; 
(SP1) compressed strategies pay an aggregation tax on multi-visit trends and cross-patient comparisons; 
(SP2) the blind-to-full gap spans $+29.8$~pp (GPT-4o-mini) to $+62.7$~pp (Haiku); 
(SP3) abstention scales non-monotonically with context length. 
On the Pareto frontier Haiku dominates Sonnet under \textit{full-context} (\$25.76 vs.\ \$106.21), 
inverting the ``biggest backbone wins'' heuristic.
\end{abstract}

\section{Introduction}

Clinical assistants built on LLMs must reason over long, multi-visit patient trajectories: 
retrieving labs, tracking trends across encounters, linking findings to problems within a visit, 
recalling treatment, comparing patients, and abstaining when the record is silent. 
As context windows grow~\citep{liu2024lost,bai2024longbench}, feeding the full chart is increasingly tractable 
but expensive, latency-bound, and---as we show---interacts non-monotonically with abstention. 
Practitioners therefore rely on compact history representations: 
retrieval-augmented generation~\citep{lewis2020rag,chen2024bge}, 
structured timelines, LLM-generated summaries, 
and agentic memory systems such as \textit{Mem0}~\citep{chhikara2025mem0} and \textit{A-Mem}~\citep{xu2025amem}. 
Whether these representations preserve enough signal for longitudinal clinical reasoning 
is the empirical question this paper addresses.

Three gaps make this question hard to answer with existing resources. 
First, EHR-derived clinical benchmarks~\citep{kweon2024ehrnoteqa,fleming2024medalign,ma2024clibench,benabacha2025medec} 
are dominated by single-encounter or single-document prompts; they do not stress multi-visit aggregation, 
encounter-local linkage, or abstention under unstated facts. 
Second, open-domain memory benchmarks~\citep{maharana2024evaluating,wu2024longmemeval,hu2025memoryagentbench} 
evaluate dialogue memory but lack clinical semantics and event-level source provenance, 
so a failure cannot be traced to a specific dialogue turn. 
Third, when frontier LLMs are evaluated on long clinical context, 
known long-context pathologies (lost-in-the-middle attention~\citep{liu2024lost}, 
miscalibrated abstention~\citep{kadavath2022know}) 
have not been characterized at the level of specific clinical cognitive operations.

\textbf{ClinTraceBench} fills these gaps with 385 MIMIC-IV-derived verified dialogues~\citep{johnson2023mimic} 
carrying event-ID provenance, a nine-task taxonomy covering 
fact recall, temporal trends, encounter-level linkage, retrospective propagation, 
contradiction, cross-patient comparison, treatment recall, observed treatment response, and abstention, 
and L0--L4 deterministic + L5 human-audit validation \footnote{L0--L5 denote six validation gates applied to each (anchor, gold) pair: L0 schema check, L1 source verification, L2 gold consistency, L3 deduplication, L4 stratification balance, and L5 human audit. Full criteria in Appendix~\ref{app:validation}.} (98.92\%). 
We evaluate eight representation strategies against four frontier backbones 
(DeepSeek-V3, GPT-4o-mini, Haiku~4.5, Sonnet~4.6) on a fixed 6{,}271-question set: 
32 cells, 200{,}672 predictions, \$527.85. 
Contributions:

\begin{enumerate}[leftmargin=*, nosep]
    \item \textbf{Benchmark construction.} A reproducible pipeline turning real MIMIC-IV trajectories into verified, source-anchored dialogues with L0--L4 deterministic + L5 human spot-check validation---linking every gold answer to a specific dialogue turn, a property absent from prior memory benchmarks~\citep{maharana2024evaluating,wu2024longmemeval}.
    
    \item \textbf{Task taxonomy.} Nine tasks covering longitudinal axes existing clinical and memory benchmarks miss, organized by the cognitive operation a clinician performs over a chart (T1--T9, defined in §3).
    
    \item \textbf{Controlled preservation probe.} A sentence-level injection probe (T3), run with the sentence inserted both before construction (equal input) and after it (update/staleness), isolating signal preservation from backbone capacity---a methodology generalizable beyond clinical settings.
    
    \item \textbf{Findings.} Full context wins on pooled accuracy but is not uniformly best; dense retrieval is competitive at lower cost; compressed and agentic memory representations exhibit task-specific failures (aggregation tax, loss of the finding--diagnosis relation even under equal input, abstention drift); the cost--quality Pareto frontier inverts the ``biggest backbone wins'' heuristic.
\end{enumerate}

\begin{figure*}[htbp]
\centering
\includegraphics[width=0.99\textwidth]{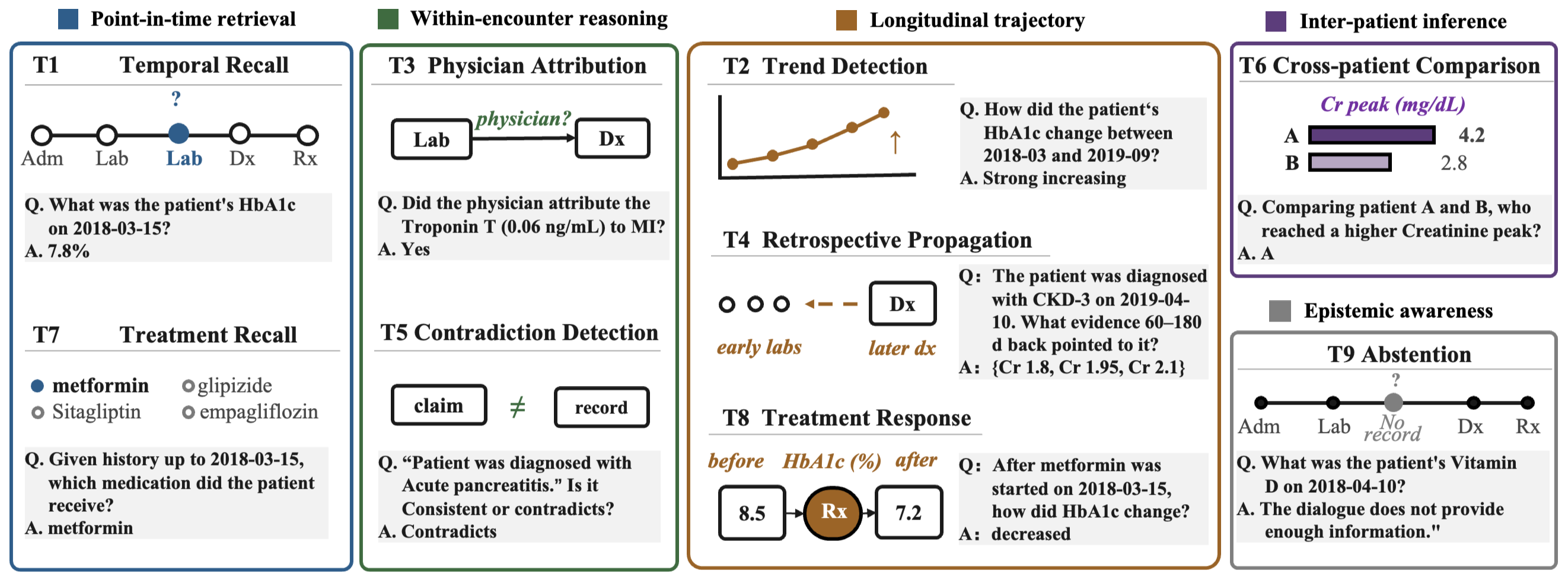}
\caption{ClinTraceBench task taxonomy. The nine tasks (T1--T9) are grouped by five cognitive operations: point-in-time retrieval (T1, T7), within-encounter reasoning (T3, T5), longitudinal trajectory (T2, T4, T8), inter-patient comparison (T6), and epistemic awareness (T9). Each card shows the task identifier, a schematic of the source anchor, and one example question (Q) with its deterministic gold answer (A). Per-task question counts are broken out in \Cref{tab:main}.}
\label{fig:1}
\end{figure*}
\section{Related work}
\label{sec:related}

\textbf{EHR-derived clinical evaluation.} 
\textsc{EHRNoteQA}~\citep{kweon2024ehrnoteqa} curates MIMIC-IV discharge-summary QA pairs validated by clinicians; 
\textsc{MedAlign}~\citep{fleming2024medalign} similarly grounds clinician-written instructions in real EHR notes; 
\textsc{CliBench}~\citep{ma2024clibench} extends MIMIC-IV to multi-decision clinical tasks 
(diagnosis, procedure, lab, prescription); 
\textsc{MEDEC}~\citep{benabacha2025medec} probes medical error detection on clinical notes; 
\textsc{DR.BENCH}~\citep{gao2023drbench} targets progress-note diagnostic reasoning over multi-encounter MIMIC-III records; 
\textsc{MedHELM}~\citep{bedi2024medhelm} provides a multi-capability evaluation framework for medical LLMs. 
These benchmarks evaluate models against real chart data, but each is dominated by 
single-document or single-encounter prompts. None stress multi-visit aggregation, 
encounter-local attribution, or abstention under unstated facts.

\textbf{Temporal and longitudinal clinical reasoning.} 
The i2b2 temporal challenges~\citep{sun2013i2b2} established temporal-relation extraction on clinical notes; 
\textsc{TIMER}~\citep{cui2025timer} introduces temporal instruction tuning over multi-encounter EHR data; 
and \citet{kruse2025temporal} evaluate LLM temporal reasoning for longitudinal clinical summarization and prediction. 
None of these provide event-level source-provenance from the answer back to a specific dialogue turn, 
nor a controlled probe that isolates signal-preservation failure modes from backbone reasoning capacity.

\textbf{Long-context and retrieval evaluation in NLP.} 
Long-context benchmarks evaluate retrieval and reasoning across extended inputs: 
\textsc{LongBench}~\citep{bai2024longbench} and needle-in-a-haystack probes~\citep{kamradt2023niah} 
characterize attention drift; \citet{liu2024lost} document the lost-in-the-middle phenomenon. 
Retrieval-augmented generation~\citep{lewis2020rag} and \textit{dense-retrieval} embeddings~\citep{chen2024bge} 
form the foundation of one strategy family we evaluate. 
\citet{kadavath2022know} establish calibration and abstention as a measurable LLM capability, 
which our T9 task operationalizes in a clinical setting.

\textbf{Memory benchmarks and agentic memory.} 
\textsc{LoCoMo}~\citep{maharana2024evaluating} evaluates multi-session conversational memory; 
\textsc{LongMemEval}~\citep{wu2024longmemeval} extends to longer, more diverse memory horizons; 
\textsc{MemoryAgentBench}~\citep{hu2025memoryagentbench} evaluates incremental multi-turn memory; 
a recent survey~\citep{zhang2025memory_survey} reviews memory mechanisms in LLM agents. 
The open-domain memory lineage motivates agentic-memory methods we evaluate Mem0~\citep{chhikara2025mem0}, A\-MEM~\citep{xu2025amem}, MemGPT~\citep{packer2023memgpt}. 
We re-purpose this lineage to ask whether memory-style representations preserve enough longitudinal 
clinical signal to compete with \textit{full-context} feeding on tasks designed around real EHR trajectories.

\textbf{Position of our work.} 
We are not the first MIMIC-IV~\citep{johnson2023mimic} benchmark, 
not the first temporal benchmark, and not the first memory benchmark. 
Our contribution is the integration: real EHR-derived longitudinal trajectories, 
verified dialogue substrate with event-level provenance, 
a nine-task taxonomy covering fact / temporal / linkage / retrospective / contradiction / 
comparison / treatment / response / abstention, 
L0--L4 deterministic + L5 human-audit validation, 
and a controlled-injection probe that isolates representation-level signal preservation 
from backbone capacity---enabling head-to-head paired comparison of history representation strategies.
\section{Benchmark design}
\label{sec:design}
\begin{figure*}[htbp]
\centering
\includegraphics[width=0.99\textwidth]{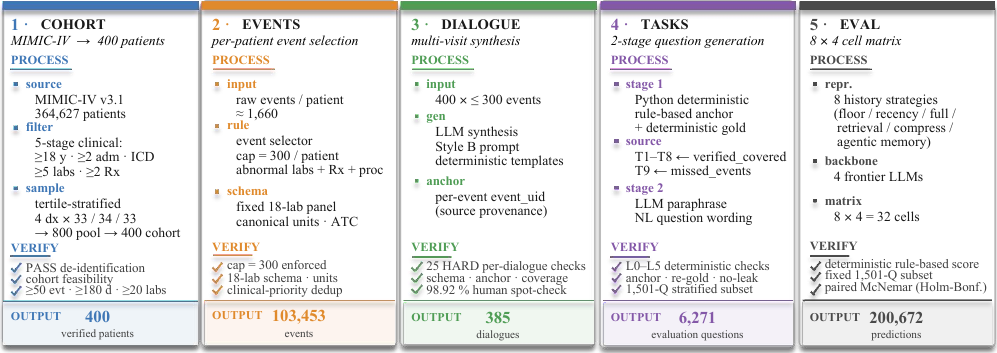}
\caption{ClinTraceBench construction pipeline. 1.\ Cohort: MIMIC-IV $364{,}627 \to 400$, 4-disease balanced + complexity-tertile stratified. 2.\ Events: fixed 18-lab panel, deterministic event IDs for provenance. 3.\ Dialogue: Sonnet~4.6 synthesis + DeepSeek/Qwen3-Max rescue; verifier v3.2 (10 HARD + 3 SOFT); 385/400 pass. 4.\ Tasks: Python-generated anchor+gold, Sonnet paraphrases to questions; L0--L4 deterministic + L5 human audit yields the 6{,}271-question set (98.92\% agreement). 5.\ Evaluation: $8 \times 4 = 32$ cells, 200{,}672 predictions. Each panel lists its \textsc{Process}, \textsc{Verify}, \textsc{Output}.}
\label{fig:pipeline}
\end{figure*}

\Cref{fig:1} shows the T1--T9 task taxonomy organized by the cognitive operation a clinician performs when reading a longitudinal record; \Cref{fig:pipeline} shows the source-verifiable construction pipeline, whose per-stage Output bands report the full-cohort headline counts ($400$ patients yielding $385$ verified dialogues, $6{,}271$ evaluation questions, $200{,}672$ predictions) and per-stage validation gates.

\textbf{Cohort.} We sample 400 MIMIC-IV (hospital-table) patients, balanced across four target diseases (diabetes, hypertension, CKD, CAD; 100 each) and stratified within each disease into low / medium / high complexity tertiles (33/34/33) by event, admission, span, and medication counts. Inclusion requires $\geq 2$ admissions $>7$ days apart, $\geq 5$ labs, $\geq 2$ prescriptions, and $\geq 2$ ICD codes in the index disease; sampling is deterministic. 15 patients fail dialogue invariants during synthesis and are dropped, yielding 385 verified dialogues. Lab extraction uses a fixed 18-lab panel with canonical units and clinical-priority deduplication (full schema in Appendix). Each record is transformed into a multi-visit patient--clinician dialogue, preserving structured fields (dates, lab values, ICD / procedure codes, medications) inside narrative turns. Consistent with prior evidence that frontier LLMs encode clinical knowledge~\citep{singhal2023large}, we use a held-out frontier generator with deterministic templates for structured fields. MIMIC-IV is HIPAA-compliant and we introduce no additional identifiers.

\textbf{Task layers.} The nine tasks make unequal demands on longitudinal reasoning and fall into three layers: \textbf{preservation / access} (T1, T3, T7), which tests whether a representation retains the evidence at all---a prerequisite for cross-visit reasoning rather than an instance of it; \textbf{integrative} (T2, T4, T6, T8), which requires synthesis across visits or patients; and \textbf{diagnostic / epistemic probes} (T5, T9), which characterize contradiction handling and abstention and carry trivial-baseline ceilings. We reserve \textit{reasoning} claims for the integrative layer and describe T1 / T3 / T7 results as evidence preservation or representation fidelity.

\textbf{Task suite.} The nine tasks are:
\begin{itemize}[leftmargin=*, nosep]
    \item \textbf{T1 --- Numeric lab retrieval.} Given a lab name and date, return the value.
    \item \textbf{T2 --- Trend classification.} Direction (increased / decreased / stable) and magnitude across two-to-five visits.
    \item \textbf{T3 --- Controlled attribution detection.} For selected same-encounter finding--problem pairs, the dialogue either contains or omits a physician-attribution sentence of the form ``physician noted: finding is consistent with problem'' (balanced yes / no). T3 runs in two settings. \textbf{(a) Post-construction} (\textit{staleness probe}): the sentence is injected after each representation is built, so the four upstream-built strategies (Mem0, A-Mem, \textit{llm-summary}, \textit{structured-timeline}) cannot recover it by construction---this measures staleness under update, not discarding. \textbf{(b) Pre-construction equal input} (\textit{preservation probe}, our primary setting): the sentence enters the dialogue \textit{before} Mem0, A-Mem or the \textit{llm-summary} is built, so every constructor sees it; it is scored by positive-class recall on the injected positives rather than balanced accuracy (\Cref{sec:sp4}).
    \item \textbf{T4 --- Retrospective propagation.} Given a later diagnosis, list the earlier findings constituting its evidence. Scored as binarized set-F1 on event UIDs (1 iff exact match), so T4 pools identically with accuracy. Cross-visit and longitudinal.
    \item \textbf{T5 --- Controlled contradiction detection (exploratory).} Does the chart contain a controlled self-contradictory statement? Contradicts-skewed by construction (455 / 40), giving a trivial majority baseline of 0.919 that no cell beats; reported only for relative ordering.
    \item \textbf{T6 --- Cross-patient comparison.} Three subtypes T6a / T6b / T6c (number of diagnoses, max lab value, delta value).
    \item \textbf{T7 --- Treatment fact recall.} Four-option clinical-management question grounded in documented treatment events; pool split between medication and procedure subtypes to avoid within-task class bias.
    \item \textbf{T8 --- Observed treatment response (exploratory).} Did a lab increase, decrease, or stay similar after a documented treatment event? Exploratory: concurrent-medication annotations were not retained, so T8 tracks post-treatment lab change, not causal response.
    \item \textbf{T9 --- Abstention.} All gold answers are ``insufficient information''; T9 accuracy equals the abstention rate, with over-answer rate $= 1 - \text{abstention}$. A trivial always-abstain classifier scores 100\% by construction, so T9 is used as a relative ordering, not an absolute benchmark.
\end{itemize}


\textbf{Evaluation set.} The L0--L4 + L5 pipeline yields a 6{,}271-question stratified set (T1 $=$ 1500, T2 $=$ 889, T3 $=$ 600, T4 $=$ 53, T5 $=$ 495, T6 $=$ 700, T7 $=$ 400, T8 $=$ 134, T9 $=$ 1500), fixed across all 32 cells to support paired McNemar tests. At $p = 0.5$, the pooled Wilson 95\% CI is $\pm 0.6$~pp; per-task CIs range from $\pm 1.3$~pp (T1, T9) to $\pm 9$~pp (T4); per-task minimum-detectable effects span $5.1$--$27.2$~pp, leaving T8 ($17.1$~pp) weakly powered and T4 ($27.2$~pp) under-powered (Appendix~\ref{app:newresults}). Pairwise headline comparisons use Holm--Bonferroni at $\alpha = 0.05$ ($m = 12$); all 12 remain significant after correction.

\textbf{Headline metric.} We report \textit{pooled} accuracy (over all 6{,}271 questions) and \textit{macro} accuracy (mean of the nine per-task accuracies). Because per-task sizes are unbalanced ($n_{\text{T1}} = n_{\text{T9}} = 1500$ vs.\ $n_{\text{T4}} = 53$), pooled accuracy is T1/T9-dominated; macro accuracy and the per-task surface (\Cref{tab:main}) should be consulted for representation comparisons that should not be task-mix-weighted.

%

\definecolor{heatE}{HTML}{F6F9FC}  
\definecolor{heatA}{HTML}{E2EAF3}  
\definecolor{heatB}{HTML}{C7D5E4}  
\definecolor{heatC}{HTML}{AABCD3}  
\definecolor{heatD}{HTML}{8AA4C3}  

\definecolor{dPos0}{HTML}{F2F8F4}
\definecolor{dPos1}{HTML}{E0EFE6}
\definecolor{dPos2}{HTML}{C0DDC8}
\definecolor{dPos3}{HTML}{8ABFA0}

\begin{table*}[!htbp]
    \centering
    \footnotesize
    \setlength{\tabcolsep}{2.0pt}
    \renewcommand{\arraystretch}{1.2}
    \resizebox{\textwidth}{!}{%
    \begin{tabular}{@{}l *{8}{cccc}@{}}
    \toprule
     & \multicolumn{4}{c}{\textbf{blind}} & \multicolumn{4}{c}{\textbf{last-vis}} & \multicolumn{4}{c}{\textbf{full}} & \multicolumn{4}{c}{\textbf{dense}} & \multicolumn{4}{c}{\textbf{struct.}} & \multicolumn{4}{c}{\textbf{summ.}} & \multicolumn{4}{c}{\textbf{Mem0}} & \multicolumn{4}{c}{\textbf{A-MEM}} \\
    \cmidrule(lr){2-5}\cmidrule(lr){6-9}\cmidrule(lr){10-13}\cmidrule(lr){14-17}\cmidrule(lr){18-21}\cmidrule(lr){22-25}\cmidrule(lr){26-29}\cmidrule(lr){30-33}
    \textbf{Task} & DS & GPT & Hk & Sn & DS & GPT & Hk & Sn & DS & GPT & Hk & Sn & DS & GPT & Hk & Sn & DS & GPT & Hk & Sn & DS & GPT & Hk & Sn & DS & GPT & Hk & Sn & DS & GPT & Hk & Sn \\
    \midrule
    T1 & \cellcolor{heatE}\textcolor{black}{0} & \cellcolor{heatE}\textcolor{black}{0} & \cellcolor{heatE}\textcolor{black}{1} & \cellcolor{heatE}\textcolor{black}{0} & \cellcolor{heatE}\textcolor{black}{16} & \cellcolor{heatE}\textcolor{black}{15} & \cellcolor{heatE}\textcolor{black}{13} & \cellcolor{heatE}\textcolor{black}{12} & \cellcolor{heatD}\textcolor{black}{81} & \cellcolor{heatD}\textcolor{black}{78} & \cellcolor{heatD}\textcolor{black}{81} & \cellcolor{heatD}\textcolor{black}{84} & \cellcolor{heatD}\textcolor{black}{80} & \cellcolor{heatD}\textcolor{black}{74} & \cellcolor{heatD}\textcolor{black}{78} & \cellcolor{heatD}\textcolor{black}{79} & \cellcolor{heatD}\textcolor{black}{89} & \cellcolor{heatD}\textcolor{black}{89} & \cellcolor{heatD}\textcolor{black}{92} & \cellcolor{heatD}\textcolor{black}{92} & \cellcolor{heatA}\textcolor{black}{36} & \cellcolor{heatA}\textcolor{black}{36} & \cellcolor{heatA}\textcolor{black}{35} & \cellcolor{heatA}\textcolor{black}{36} & \cellcolor{heatB}\textcolor{black}{52} & \cellcolor{heatB}\textcolor{black}{52} & \cellcolor{heatB}\textcolor{black}{51} & \cellcolor{heatB}\textcolor{black}{49} & \cellcolor{heatB}\textcolor{black}{49} & \cellcolor{heatA}\textcolor{black}{42} & \cellcolor{heatA}\textcolor{black}{43} & \cellcolor{heatB}\textcolor{black}{49} \\
    T2 & \cellcolor{heatE}\textcolor{black}{17} & \cellcolor{heatA}\textcolor{black}{33} & \cellcolor{heatE}\textcolor{black}{0} & \cellcolor{heatA}\textcolor{black}{30} & \cellcolor{heatA}\textcolor{black}{33} & \cellcolor{heatA}\textcolor{black}{35} & \cellcolor{heatE}\textcolor{black}{18} & \cellcolor{heatA}\textcolor{black}{35} & \cellcolor{heatA}\textcolor{black}{38} & \cellcolor{heatE}\textcolor{black}{26} & \cellcolor{heatA}\textcolor{black}{35} & \cellcolor{heatB}\textcolor{black}{47} & \cellcolor{heatA}\textcolor{black}{35} & \cellcolor{heatE}\textcolor{black}{27} & \cellcolor{heatA}\textcolor{black}{43} & \cellcolor{heatA}\textcolor{black}{43} & \cellcolor{heatE}\textcolor{black}{30} & \cellcolor{heatE}\textcolor{black}{21} & \cellcolor{heatE}\textcolor{black}{29} & \cellcolor{heatA}\textcolor{black}{41} & \cellcolor{heatE}\textcolor{black}{26} & \cellcolor{heatA}\textcolor{black}{33} & \cellcolor{heatA}\textcolor{black}{32} & \cellcolor{heatA}\textcolor{black}{34} & \cellcolor{heatA}\textcolor{black}{33} & \cellcolor{heatE}\textcolor{black}{30} & \cellcolor{heatA}\textcolor{black}{35} & \cellcolor{heatA}\textcolor{black}{39} & \cellcolor{heatA}\textcolor{black}{33} & \cellcolor{heatA}\textcolor{black}{32} & \cellcolor{heatA}\textcolor{black}{37} & \cellcolor{heatA}\textcolor{black}{38} \\
    T3 & \cellcolor{heatB}\textcolor{black}{50} & \cellcolor{heatB}\textcolor{black}{49} & \cellcolor{heatB}\textcolor{black}{50} & \cellcolor{heatB}\textcolor{black}{50} & \cellcolor{heatB}\textcolor{black}{56} & \cellcolor{heatB}\textcolor{black}{54} & \cellcolor{heatB}\textcolor{black}{55} & \cellcolor{heatB}\textcolor{black}{51} & \cellcolor{heatD}\textcolor{black}{94} & \cellcolor{heatD}\textcolor{black}{93} & \cellcolor{heatD}\textcolor{black}{87} & \cellcolor{heatD}\textcolor{black}{89} & \cellcolor{heatD}\textcolor{black}{83} & \cellcolor{heatD}\textcolor{black}{85} & \cellcolor{heatD}\textcolor{black}{87} & \cellcolor{heatD}\textcolor{black}{77} & \cellcolor{heatB}\textcolor{black}{51} & \cellcolor{heatB}\textcolor{black}{51} & \cellcolor{heatB}\textcolor{black}{50} & \cellcolor{heatB}\textcolor{black}{50} & \cellcolor{heatB}\textcolor{black}{50} & \cellcolor{heatB}\textcolor{black}{49} & \cellcolor{heatB}\textcolor{black}{50} & \cellcolor{heatB}\textcolor{black}{50} & \cellcolor{heatB}\textcolor{black}{49} & \cellcolor{heatB}\textcolor{black}{48} & \cellcolor{heatB}\textcolor{black}{50} & \cellcolor{heatB}\textcolor{black}{50} & \cellcolor{heatB}\textcolor{black}{50} & \cellcolor{heatB}\textcolor{black}{48} & \cellcolor{heatB}\textcolor{black}{48} & \cellcolor{heatB}\textcolor{black}{50} \\
    T4 & \cellcolor{heatE}\textcolor{black}{13} & \cellcolor{heatE}\textcolor{black}{8} & \cellcolor{heatE}\textcolor{black}{19} & \cellcolor{heatE}\textcolor{black}{11} & \cellcolor{heatE}\textcolor{black}{6} & \cellcolor{heatE}\textcolor{black}{11} & \cellcolor{heatE}\textcolor{black}{17} & \cellcolor{heatE}\textcolor{black}{11} & \cellcolor{heatE}\textcolor{black}{23} & \cellcolor{heatE}\textcolor{black}{28} & \cellcolor{heatE}\textcolor{black}{19} & \cellcolor{heatE}\textcolor{black}{25} & \cellcolor{heatE}\textcolor{black}{21} & \cellcolor{heatE}\textcolor{black}{23} & \cellcolor{heatE}\textcolor{black}{15} & \cellcolor{heatE}\textcolor{black}{6} & \cellcolor{heatE}\textcolor{black}{6} & \cellcolor{heatE}\textcolor{black}{21} & \cellcolor{heatE}\textcolor{black}{25} & \cellcolor{heatE}\textcolor{black}{17} & \cellcolor{heatE}\textcolor{black}{15} & \cellcolor{heatE}\textcolor{black}{17} & \cellcolor{heatE}\textcolor{black}{19} & \cellcolor{heatE}\textcolor{black}{11} & \cellcolor{heatE}\textcolor{black}{4} & \cellcolor{heatE}\textcolor{black}{13} & \cellcolor{heatE}\textcolor{black}{9} & \cellcolor{heatE}\textcolor{black}{15} & \cellcolor{heatE}\textcolor{black}{8} & \cellcolor{heatE}\textcolor{black}{11} & \cellcolor{heatE}\textcolor{black}{9} & \cellcolor{heatE}\textcolor{black}{13} \\
    T5 & \cellcolor{heatE}\textcolor{black}{25} & \cellcolor{heatA}\textcolor{black}{39} & \cellcolor{heatE}\textcolor{black}{17} & \cellcolor{heatA}\textcolor{black}{38} & \cellcolor{heatD}\textcolor{black}{79} & \cellcolor{heatB}\textcolor{black}{58} & \cellcolor{heatA}\textcolor{black}{37} & \cellcolor{heatC}\textcolor{black}{64} & \cellcolor{heatD}\textcolor{black}{86} & \cellcolor{heatD}\textcolor{black}{90} & \cellcolor{heatD}\textcolor{black}{85} & \cellcolor{heatD}\textcolor{black}{84} & \cellcolor{heatD}\textcolor{black}{87} & \cellcolor{heatD}\textcolor{black}{89} & \cellcolor{heatD}\textcolor{black}{86} & \cellcolor{heatD}\textcolor{black}{84} & \cellcolor{heatD}\textcolor{black}{75} & \cellcolor{heatD}\textcolor{black}{78} & \cellcolor{heatD}\textcolor{black}{87} & \cellcolor{heatD}\textcolor{black}{86} & \cellcolor{heatD}\textcolor{black}{83} & \cellcolor{heatC}\textcolor{black}{67} & \cellcolor{heatC}\textcolor{black}{64} & \cellcolor{heatC}\textcolor{black}{63} & \cellcolor{heatD}\textcolor{black}{80} & \cellcolor{heatD}\textcolor{black}{73} & \cellcolor{heatD}\textcolor{black}{72} & \cellcolor{heatD}\textcolor{black}{75} & \cellcolor{heatD}\textcolor{black}{78} & \cellcolor{heatC}\textcolor{black}{69} & \cellcolor{heatD}\textcolor{black}{74} & \cellcolor{heatD}\textcolor{black}{78} \\
    T6 & \cellcolor{heatA}\textcolor{black}{36} & \cellcolor{heatA}\textcolor{black}{31} & \cellcolor{heatE}\textcolor{black}{0} & \cellcolor{heatE}\textcolor{black}{10} & \cellcolor{heatA}\textcolor{black}{42} & \cellcolor{heatB}\textcolor{black}{47} & \cellcolor{heatB}\textcolor{black}{47} & \cellcolor{heatA}\textcolor{black}{42} & \cellcolor{heatB}\textcolor{black}{54} & \cellcolor{heatB}\textcolor{black}{50} & \cellcolor{heatB}\textcolor{black}{51} & \cellcolor{heatB}\textcolor{black}{52} & \cellcolor{heatB}\textcolor{black}{47} & \cellcolor{heatA}\textcolor{black}{39} & \cellcolor{heatA}\textcolor{black}{44} & \cellcolor{heatB}\textcolor{black}{46} & \cellcolor{heatC}\textcolor{black}{61} & \cellcolor{heatB}\textcolor{black}{53} & \cellcolor{heatB}\textcolor{black}{52} & \cellcolor{heatB}\textcolor{black}{56} & \cellcolor{heatA}\textcolor{black}{44} & \cellcolor{heatA}\textcolor{black}{39} & \cellcolor{heatA}\textcolor{black}{42} & \cellcolor{heatA}\textcolor{black}{42} & \cellcolor{heatB}\textcolor{black}{46} & \cellcolor{heatA}\textcolor{black}{42} & \cellcolor{heatB}\textcolor{black}{45} & \cellcolor{heatA}\textcolor{black}{43} & \cellcolor{heatB}\textcolor{black}{50} & \cellcolor{heatB}\textcolor{black}{46} & \cellcolor{heatA}\textcolor{black}{44} & \cellcolor{heatB}\textcolor{black}{48} \\
    T7 & \cellcolor{heatE}\textcolor{black}{26} & \cellcolor{heatE}\textcolor{black}{24} & \cellcolor{heatE}\textcolor{black}{1} & \cellcolor{heatE}\textcolor{black}{20} & \cellcolor{heatA}\textcolor{black}{36} & \cellcolor{heatA}\textcolor{black}{30} & \cellcolor{heatE}\textcolor{black}{14} & \cellcolor{heatA}\textcolor{black}{37} & \cellcolor{heatD}\textcolor{black}{94} & \cellcolor{heatD}\textcolor{black}{87} & \cellcolor{heatD}\textcolor{black}{97} & \cellcolor{heatD}\textcolor{black}{95} & \cellcolor{heatD}\textcolor{black}{90} & \cellcolor{heatD}\textcolor{black}{86} & \cellcolor{heatD}\textcolor{black}{94} & \cellcolor{heatD}\textcolor{black}{93} & \cellcolor{heatC}\textcolor{black}{64} & \cellcolor{heatC}\textcolor{black}{62} & \cellcolor{heatB}\textcolor{black}{51} & \cellcolor{heatC}\textcolor{black}{69} & \cellcolor{heatB}\textcolor{black}{51} & \cellcolor{heatB}\textcolor{black}{48} & \cellcolor{heatA}\textcolor{black}{35} & \cellcolor{heatB}\textcolor{black}{45} & \cellcolor{heatD}\textcolor{black}{73} & \cellcolor{heatC}\textcolor{black}{64} & \cellcolor{heatB}\textcolor{black}{54} & \cellcolor{heatC}\textcolor{black}{66} & \cellcolor{heatC}\textcolor{black}{65} & \cellcolor{heatC}\textcolor{black}{62} & \cellcolor{heatA}\textcolor{black}{44} & \cellcolor{heatC}\textcolor{black}{62} \\
    T8 & \cellcolor{heatE}\textcolor{black}{23} & \cellcolor{heatE}\textcolor{black}{25} & \cellcolor{heatE}\textcolor{black}{17} & \cellcolor{heatE}\textcolor{black}{26} & \cellcolor{heatA}\textcolor{black}{31} & \cellcolor{heatA}\textcolor{black}{35} & \cellcolor{heatE}\textcolor{black}{18} & \cellcolor{heatE}\textcolor{black}{23} & \cellcolor{heatA}\textcolor{black}{44} & \cellcolor{heatE}\textcolor{black}{27} & \cellcolor{heatB}\textcolor{black}{46} & \cellcolor{heatB}\textcolor{black}{47} & \cellcolor{heatA}\textcolor{black}{39} & \cellcolor{heatE}\textcolor{black}{28} & \cellcolor{heatA}\textcolor{black}{40} & \cellcolor{heatA}\textcolor{black}{40} & \cellcolor{heatA}\textcolor{black}{45} & \cellcolor{heatA}\textcolor{black}{40} & \cellcolor{heatB}\textcolor{black}{54} & \cellcolor{heatB}\textcolor{black}{49} & \cellcolor{heatE}\textcolor{black}{23} & \cellcolor{heatE}\textcolor{black}{25} & \cellcolor{heatE}\textcolor{black}{19} & \cellcolor{heatE}\textcolor{black}{22} & \cellcolor{heatA}\textcolor{black}{37} & \cellcolor{heatA}\textcolor{black}{36} & \cellcolor{heatA}\textcolor{black}{31} & \cellcolor{heatE}\textcolor{black}{30} & \cellcolor{heatA}\textcolor{black}{37} & \cellcolor{heatA}\textcolor{black}{34} & \cellcolor{heatE}\textcolor{black}{29} & \cellcolor{heatA}\textcolor{black}{37} \\
    T9 & \cellcolor{heatA}\textcolor{black}{36} & \cellcolor{heatC}\textcolor{black}{69} & \cellcolor{heatE}\textcolor{black}{0} & \cellcolor{heatE}\textcolor{black}{20} & \cellcolor{heatD}\textcolor{black}{88} & \cellcolor{heatD}\textcolor{black}{90} & \cellcolor{heatD}\textcolor{black}{99} & \cellcolor{heatD}\textcolor{black}{93} & \cellcolor{heatD}\textcolor{black}{70} & \cellcolor{heatC}\textcolor{black}{70} & \cellcolor{heatD}\textcolor{black}{91} & \cellcolor{heatD}\textcolor{black}{75} & \cellcolor{heatD}\textcolor{black}{77} & \cellcolor{heatD}\textcolor{black}{71} & \cellcolor{heatD}\textcolor{black}{92} & \cellcolor{heatD}\textcolor{black}{79} & \cellcolor{heatB}\textcolor{black}{56} & \cellcolor{heatC}\textcolor{black}{66} & \cellcolor{heatB}\textcolor{black}{48} & \cellcolor{heatA}\textcolor{black}{44} & \cellcolor{heatD}\textcolor{black}{80} & \cellcolor{heatD}\textcolor{black}{78} & \cellcolor{heatD}\textcolor{black}{94} & \cellcolor{heatD}\textcolor{black}{74} & \cellcolor{heatD}\textcolor{black}{81} & \cellcolor{heatD}\textcolor{black}{82} & \cellcolor{heatD}\textcolor{black}{94} & \cellcolor{heatD}\textcolor{black}{73} & \cellcolor{heatD}\textcolor{black}{82} & \cellcolor{heatD}\textcolor{black}{78} & \cellcolor{heatD}\textcolor{black}{97} & \cellcolor{heatD}\textcolor{black}{73} \\
    \midrule
    Pool. & \cellcolor{heatE}\textcolor{black}{24} & \cellcolor{heatA}\textcolor{black}{33} & \cellcolor{heatE}\textcolor{black}{8} & \cellcolor{heatE}\textcolor{black}{20} & \cellcolor{heatB}\textcolor{black}{47} & \cellcolor{heatB}\textcolor{black}{46} & \cellcolor{heatA}\textcolor{black}{42} & \cellcolor{heatA}\textcolor{black}{45} & \cellcolor{heatC}\textcolor{black}{67} & \cellcolor{heatC}\textcolor{black}{63} & \cellcolor{heatD}\textcolor{black}{71} & \cellcolor{heatC}\textcolor{black}{70} & \cellcolor{heatC}\textcolor{black}{66} & \cellcolor{heatC}\textcolor{black}{60} & \cellcolor{heatC}\textcolor{black}{70} & \cellcolor{heatC}\textcolor{black}{66} & \cellcolor{heatB}\textcolor{black}{59} & \cellcolor{heatB}\textcolor{black}{59} & \cellcolor{heatB}\textcolor{black}{58} & \cellcolor{heatB}\textcolor{black}{60} & \cellcolor{heatB}\textcolor{black}{48} & \cellcolor{heatB}\textcolor{black}{47} & \cellcolor{heatB}\textcolor{black}{50} & \cellcolor{heatB}\textcolor{black}{46} & \cellcolor{heatB}\textcolor{black}{55} & \cellcolor{heatB}\textcolor{black}{54} & \cellcolor{heatB}\textcolor{black}{56} & \cellcolor{heatB}\textcolor{black}{52} & \cellcolor{heatB}\textcolor{black}{55} & \cellcolor{heatB}\textcolor{black}{51} & \cellcolor{heatB}\textcolor{black}{54} & \cellcolor{heatB}\textcolor{black}{53} \\
    Macro & \cellcolor{heatE}\textcolor{black}{25} & \cellcolor{heatA}\textcolor{black}{31} & \cellcolor{heatE}\textcolor{black}{12} & \cellcolor{heatE}\textcolor{black}{23} & \cellcolor{heatA}\textcolor{black}{43} & \cellcolor{heatA}\textcolor{black}{42} & \cellcolor{heatA}\textcolor{black}{35} & \cellcolor{heatA}\textcolor{black}{41} & \cellcolor{heatC}\textcolor{black}{65} & \cellcolor{heatC}\textcolor{black}{61} & \cellcolor{heatC}\textcolor{black}{66} & \cellcolor{heatC}\textcolor{black}{67} & \cellcolor{heatC}\textcolor{black}{62} & \cellcolor{heatB}\textcolor{black}{58} & \cellcolor{heatC}\textcolor{black}{64} & \cellcolor{heatC}\textcolor{black}{61} & \cellcolor{heatB}\textcolor{black}{53} & \cellcolor{heatB}\textcolor{black}{53} & \cellcolor{heatB}\textcolor{black}{54} & \cellcolor{heatB}\textcolor{black}{56} & \cellcolor{heatB}\textcolor{black}{45} & \cellcolor{heatA}\textcolor{black}{44} & \cellcolor{heatA}\textcolor{black}{43} & \cellcolor{heatA}\textcolor{black}{42} & \cellcolor{heatB}\textcolor{black}{51} & \cellcolor{heatB}\textcolor{black}{49} & \cellcolor{heatB}\textcolor{black}{49} & \cellcolor{heatB}\textcolor{black}{49} & \cellcolor{heatB}\textcolor{black}{50} & \cellcolor{heatB}\textcolor{black}{47} & \cellcolor{heatB}\textcolor{black}{47} & \cellcolor{heatB}\textcolor{black}{50} \\
    $\Delta_{\text{blind}}$ & \textcolor{gray}{---} & \textcolor{gray}{---} & \textcolor{gray}{---} & \textcolor{gray}{---} & \cellcolor{dPos1}\textcolor{black}{18} & \cellcolor{dPos1}\textcolor{black}{11} & \cellcolor{dPos2}\textcolor{black}{24} & \cellcolor{dPos1}\textcolor{black}{18} & \cellcolor{dPos2}\textcolor{black}{40} & \cellcolor{dPos2}\textcolor{black}{30} & \cellcolor{dPos3}\textcolor{black}{54} & \cellcolor{dPos3}\textcolor{black}{44} & \cellcolor{dPos2}\textcolor{black}{37} & \cellcolor{dPos2}\textcolor{black}{27} & \cellcolor{dPos3}\textcolor{black}{53} & \cellcolor{dPos2}\textcolor{black}{38} & \cellcolor{dPos2}\textcolor{black}{28} & \cellcolor{dPos2}\textcolor{black}{22} & \cellcolor{dPos3}\textcolor{black}{43} & \cellcolor{dPos2}\textcolor{black}{33} & \cellcolor{dPos2}\textcolor{black}{20} & \cellcolor{dPos1}\textcolor{black}{13} & \cellcolor{dPos2}\textcolor{black}{32} & \cellcolor{dPos1}\textcolor{black}{19} & \cellcolor{dPos2}\textcolor{black}{26} & \cellcolor{dPos1}\textcolor{black}{18} & \cellcolor{dPos2}\textcolor{black}{37} & \cellcolor{dPos2}\textcolor{black}{26} & \cellcolor{dPos2}\textcolor{black}{25} & \cellcolor{dPos1}\textcolor{black}{16} & \cellcolor{dPos2}\textcolor{black}{36} & \cellcolor{dPos2}\textcolor{black}{27} \\
    \bottomrule
    \end{tabular}%
    }
   \caption{ClinTraceBench main results: 8 strategies $\times$ 4 backbones $\times$ 9 tasks.
Per-cell accuracy (\%) for every (strategy, backbone, task) triple. 
Bottom rows: Pool. = accuracy pooled across tasks weighted by subset size; 
Macro = equal-task-weight mean; 
$\boldsymbol{\Delta_{\text{blind}}}$ = Macro minus the same-backbone \textit{no-context-blind} Macro (pp).
Task layers (\Cref{sec:design}): T1, T3, T7 are \textbf{preservation/access} tasks;
T2, T4, T6, T8 are \textbf{integrative} tasks requiring cross-visit or cross-patient synthesis;
T5 and T9 are \textbf{diagnostic/epistemic} probes with trivial-baseline ceilings.
Backbones: DS = DeepSeek-V3, GPT = GPT-4o-mini, Hk = Claude Haiku 4.5, Sn = Claude Sonnet 4.6. 
Strategies: blind = \textit{no-context-blind}; last-vis = \textit{last-visit-only}; 
full = \textit{full-context}; dense = \textit{dense-retrieval}; 
struct. = \textit{structured-timeline}; summ. = \textit{llm-summary}. 
Blue heat: accuracy 
(\colorbox{heatE}{$<\!30$}~\colorbox{heatA}{30--45}~\colorbox{heatB}{45--60}~\colorbox{heatC}{60--70}~\colorbox{heatD}{$\geq\!70$}).
Green heat ($\Delta_{\text{blind}}$ row only):
(\colorbox{dPos1}{10--20}~\colorbox{dPos2}{20--40}~\colorbox{dPos3}{$\geq\!40$}).}
    \label{tab:main}
\end{table*}

\section{History representation strategies and backbones}
\label{sec:strategies}

\textbf{History representation strategies.} Eight strategies span the principal context-access choices, grouped into six families:
\begin{itemize}[leftmargin=*, nosep]
    \item \textbf{Floor.} \textit{no-context-blind} --- the model sees only the question.
    \item \textbf{Recency.} \textit{last-visit-only} --- only the most recent encounter.
    \item \textbf{Full.} \textit{full-context} --- entire multi-visit dialogue fed verbatim; upper bound for context coverage.
    \item \textbf{Retrieval.} \textit{dense-retrieval} --- BGE-M3 chunk embeddings per patient, top-$K{=}5$, the chunk unit being one complete clinical visit. Reranking and hybrid retrieval are not evaluated; a $K \in \{3,5,10\}$ ablation shows no consistent gain from larger $K$ (Appendix~\ref{app:newresults}).
    \item \textbf{Compression.} \textit{structured-timeline} (per-encounter table) and \textit{llm-summary} (single 500-token DeepSeek summary; doubling the budget to 1{,}000 tokens does not remove the aggregation tax, Appendix~\ref{app:newresults}).
    \item \textbf{Agentic memory.} Mem0 (extracted facts, DeepSeek-prepared, top-$K{=}5$) and A-Mem (Zettelkasten-style atomic notes, DeepSeek-prepared).
\end{itemize}

\textbf{Backbones.} Inference is routed through OpenRouter: DeepSeek-V3 (deepseek-chat-v3.1, 64k), GPT-4o-mini (gpt-4o-mini, 128k), Claude Haiku~4.5 (claude-haiku-4.5, 200k), and Claude Sonnet~4.6 (claude-sonnet-4.6, 200k). Prompt-prefix caching is enabled across all backbones (90.7~M cache-read tokens, a 17.1\% cache-read hit rate); output-token caps are 120 for T8 and 80 for T9. Provider caching settings, T8 / T9 output-token caps, and the T9 abstention parser are in Appendix~\ref{app:appA8}.

\textbf{Memory prep model.} Both agentic-memory strategies extract artifacts (facts or atomic notes) from the dialogue once before any question is asked. We hold the prep model fixed at DeepSeek-V3 across all four answering backbones, isolating answering-side use of memory from construction-side extraction. A consequence: non-DeepSeek Mem0 and A-Mem cells run a heterogeneous pipeline (DeepSeek-extracted artifacts answered by GPT-4o-mini / Haiku / Sonnet); we discuss the SP2 impact in §Limitations.

\section{Results}
\label{sec:results}
\subsection{Main accuracy}
\label{sec:results-main}

\Cref{tab:main} reports the full strategy-$\times$-backbone-$\times$-task cube; the three right-margin columns give pooled, macro, and $\Delta_\text{blind}$ (macro lift over the same-backbone blind). \textit{full-context} is best on every backbone (pooled 67.2 / 63.0 / 70.6 / 69.8 for DeepSeek / GPT-4o-mini / Haiku / Sonnet). \textit{dense-retrieval} follows on every backbone (60.4--69.7), within 1--4~pp of \textit{full-context}---BGE-M3 with $K{=}5$ captures most but not all of what \textit{full-context} provides. \textit{structured-timeline} sits third (58.4--59.6) and is strikingly backbone-invariant ($\pm 0.6$~pp). \textit{Mem0} and \textit{A-Mem} cluster at 50.8--55.9; \textit{llm-summary} and \textit{last-visit-only} trail at 41.6--49.5. Macro accuracy preserves the same ordering as pooled. The per-task surface exposes the difficulty structure: even under \textit{full-context}, T2 / T4 / T6 / T8 / T9 remain hard while T1 / T3 / T5 / T7 are tractable; compressed strategies show the T3 constant-No band of 50.0 cells under post-construction injection (\Cref{sec:sp4}) and the aggregation tax on T2 / T6 / T8.

Three checks show this ordering is not an artifact of task mix, scoring metric, or prep model (Appendix~\ref{app:newresults}). It survives recomputing macro accuracy without T5 / T8 / T9 and without the T3 probe. Under macro-F1, \textit{full-context}, \textit{dense-retrieval} and \textit{structured-timeline} stay above the compressed strategies on T5 (0.695 / 0.736 / 0.656 vs.\ a 0.478 majority baseline) and \textit{structured-timeline} leads on T8 (0.523 vs.\ 0.247), so the skewed tasks do separate representations. And rebuilding \textit{Mem0} / \textit{A-Mem} with each answering backbone gives six matched cells, none better than DeepSeek-prepared ($\Delta$ $-0.045$ to $-0.244$).

\Cref{tab:context-delta} decomposes the context lift (vs.\ same-backbone blind) for three regimes (\textit{full-context}, \textit{dense-retrieval}, and the agentic-memory mean of \textit{Mem0} and \textit{A-Mem}); blue shades positive lift, red the rare cells where context hurts.

%
%

\definecolor{dNegMid}{HTML}{E5A399}    
\definecolor{dNegLite}{HTML}{F4DAD4}   
\definecolor{dGray}{HTML}{ECECEC}      
\definecolor{dPos0}{HTML}{F4F8FC}      
\definecolor{dPos1}{HTML}{DEE6F0}      
\definecolor{dPos2}{HTML}{BCCDE0}      
\definecolor{dPos3}{HTML}{95B0CD}      

\begin{table}[t]
    \centering
    \scriptsize
    \setlength{\tabcolsep}{3.0pt}
    \renewcommand{\arraystretch}{1.2}
    \resizebox{\columnwidth}{!}{%
    \begin{tabular}{@{}>{\bfseries}l l *{9}{c} !{\vrule width 0.5pt} c@{}}
    \toprule
    \textbf{Strategy} & \textbf{BB} & \textbf{T1} & \textbf{T2} & \textbf{T3} & \textbf{T4} & \textbf{T5} & \textbf{T6} & \textbf{T7} & \textbf{T8} & \textbf{T9} & \textbf{Macro} \\
    \midrule
    \multirow{4}{*}{full} & DS & \cellcolor{dPos3}\textcolor{black}{+81} & \cellcolor{dPos1}\textcolor{black}{+20} & \cellcolor{dPos2}\textcolor{black}{+44} & \cellcolor{dPos0}\textcolor{black}{+9} & \cellcolor{dPos3}\textcolor{black}{+61} & \cellcolor{dPos1}\textcolor{black}{+19} & \cellcolor{dPos3}\textcolor{black}{+69} & \cellcolor{dPos1}\textcolor{black}{+21} & \cellcolor{dPos2}\textcolor{black}{+35} & \cellcolor{dPos2}\textcolor{black}{+40} \\
     & GPT & \cellcolor{dPos3}\textcolor{black}{+78} & \cellcolor{dNegMid}\textcolor{black}{-8} & \cellcolor{dPos2}\textcolor{black}{+44} & \cellcolor{dPos1}\textcolor{black}{+21} & \cellcolor{dPos2}\textcolor{black}{+51} & \cellcolor{dPos1}\textcolor{black}{+19} & \cellcolor{dPos3}\textcolor{black}{+63} & \cellcolor{dGray}\textcolor{black}{+2} & \cellcolor{dGray}\textcolor{black}{+1} & \cellcolor{dPos1}\textcolor{black}{+30} \\
     & Hk & \cellcolor{dPos3}\textcolor{black}{+81} & \cellcolor{dPos2}\textcolor{black}{+35} & \cellcolor{dPos2}\textcolor{black}{+37} & \cellcolor{dGray}\textcolor{black}{0} & \cellcolor{dPos3}\textcolor{black}{+68} & \cellcolor{dPos2}\textcolor{black}{+51} & \cellcolor{dPos3}\textcolor{black}{+95} & \cellcolor{dPos1}\textcolor{black}{+29} & \cellcolor{dPos3}\textcolor{black}{+91} & \cellcolor{dPos2}\textcolor{black}{+54} \\
     & Sn & \cellcolor{dPos3}\textcolor{black}{+84} & \cellcolor{dPos1}\textcolor{black}{+17} & \cellcolor{dPos2}\textcolor{black}{+39} & \cellcolor{dPos0}\textcolor{black}{+13} & \cellcolor{dPos2}\textcolor{black}{+46} & \cellcolor{dPos2}\textcolor{black}{+42} & \cellcolor{dPos3}\textcolor{black}{+76} & \cellcolor{dPos1}\textcolor{black}{+21} & \cellcolor{dPos2}\textcolor{black}{+54} & \cellcolor{dPos2}\textcolor{black}{+44} \\
    \cmidrule(l{2pt}r{2pt}){1-12}
    \multirow{4}{*}{dense} & DS & \cellcolor{dPos3}\textcolor{black}{+80} & \cellcolor{dPos1}\textcolor{black}{+18} & \cellcolor{dPos2}\textcolor{black}{+33} & \cellcolor{dPos0}\textcolor{black}{+8} & \cellcolor{dPos3}\textcolor{black}{+62} & \cellcolor{dPos0}\textcolor{black}{+11} & \cellcolor{dPos3}\textcolor{black}{+64} & \cellcolor{dPos1}\textcolor{black}{+16} & \cellcolor{dPos2}\textcolor{black}{+42} & \cellcolor{dPos2}\textcolor{black}{+37} \\
     & GPT & \cellcolor{dPos3}\textcolor{black}{+74} & \cellcolor{dNegMid}\textcolor{black}{-7} & \cellcolor{dPos2}\textcolor{black}{+35} & \cellcolor{dPos1}\textcolor{black}{+15} & \cellcolor{dPos2}\textcolor{black}{+50} & \cellcolor{dPos0}\textcolor{black}{+8} & \cellcolor{dPos3}\textcolor{black}{+62} & \cellcolor{dGray}\textcolor{black}{+2} & \cellcolor{dGray}\textcolor{black}{+3} & \cellcolor{dPos1}\textcolor{black}{+27} \\
     & Hk & \cellcolor{dPos3}\textcolor{black}{+77} & \cellcolor{dPos2}\textcolor{black}{+43} & \cellcolor{dPos2}\textcolor{black}{+37} & \cellcolor{dNegLite}\textcolor{black}{-4} & \cellcolor{dPos3}\textcolor{black}{+69} & \cellcolor{dPos2}\textcolor{black}{+44} & \cellcolor{dPos3}\textcolor{black}{+93} & \cellcolor{dPos1}\textcolor{black}{+23} & \cellcolor{dPos3}\textcolor{black}{+92} & \cellcolor{dPos2}\textcolor{black}{+53} \\
     & Sn & \cellcolor{dPos3}\textcolor{black}{+78} & \cellcolor{dPos0}\textcolor{black}{+13} & \cellcolor{dPos1}\textcolor{black}{+27} & \cellcolor{dNegMid}\textcolor{black}{-6} & \cellcolor{dPos2}\textcolor{black}{+46} & \cellcolor{dPos2}\textcolor{black}{+36} & \cellcolor{dPos3}\textcolor{black}{+73} & \cellcolor{dPos0}\textcolor{black}{+13} & \cellcolor{dPos2}\textcolor{black}{+59} & \cellcolor{dPos2}\textcolor{black}{+38} \\
    \cmidrule(l{2pt}r{2pt}){1-12}
    \multirow{4}{*}{\makecell[c]{mem\\avg}} & DS & \cellcolor{dPos2}\textcolor{black}{+50} & \cellcolor{dPos1}\textcolor{black}{+16} & \cellcolor{dNegLite}\textcolor{black}{0} & \cellcolor{dNegMid}\textcolor{black}{-8} & \cellcolor{dPos2}\textcolor{black}{+54} & \cellcolor{dPos0}\textcolor{black}{+12} & \cellcolor{dPos2}\textcolor{black}{+44} & \cellcolor{dPos0}\textcolor{black}{+14} & \cellcolor{dPos2}\textcolor{black}{+46} & \cellcolor{dPos1}\textcolor{black}{+25} \\
     & GPT & \cellcolor{dPos2}\textcolor{black}{+47} & \cellcolor{dNegLite}\textcolor{black}{-2} & \cellcolor{dNegLite}\textcolor{black}{-2} & \cellcolor{dGray}\textcolor{black}{+5} & \cellcolor{dPos2}\textcolor{black}{+32} & \cellcolor{dPos0}\textcolor{black}{+13} & \cellcolor{dPos2}\textcolor{black}{+38} & \cellcolor{dPos0}\textcolor{black}{+9} & \cellcolor{dPos0}\textcolor{black}{+12} & \cellcolor{dPos1}\textcolor{black}{+17} \\
     & Hk & \cellcolor{dPos2}\textcolor{black}{+46} & \cellcolor{dPos2}\textcolor{black}{+36} & \cellcolor{dNegLite}\textcolor{black}{-1} & \cellcolor{dNegMid}\textcolor{black}{-9} & \cellcolor{dPos2}\textcolor{black}{+56} & \cellcolor{dPos2}\textcolor{black}{+45} & \cellcolor{dPos2}\textcolor{black}{+48} & \cellcolor{dPos0}\textcolor{black}{+13} & \cellcolor{dPos3}\textcolor{black}{+95} & \cellcolor{dPos2}\textcolor{black}{+36} \\
     & Sn & \cellcolor{dPos2}\textcolor{black}{+48} & \cellcolor{dPos0}\textcolor{black}{+8} & \cellcolor{dGray}\textcolor{black}{0} & \cellcolor{dGray}\textcolor{black}{+3} & \cellcolor{dPos2}\textcolor{black}{+38} & \cellcolor{dPos2}\textcolor{black}{+35} & \cellcolor{dPos2}\textcolor{black}{+44} & \cellcolor{dPos0}\textcolor{black}{+8} & \cellcolor{dPos2}\textcolor{black}{+53} & \cellcolor{dPos1}\textcolor{black}{+26} \\
    \bottomrule
    \end{tabular}%
    }
    \caption{Per-cell context-lift over \textit{no-context-blind}. 
Accuracy differences (pp) between three context regimes (\textit{full-context}, 
\textit{dense-retrieval}, mem-avg = mean of Mem0 and A-MEM) and the same-backbone, 
same-task \textit{no-context-blind} baseline. 
Macro = mean over T1--T9. 
Diverging color encodes lift 
(\colorbox{dNegMid}{$\leq\!-5$}~\colorbox{dNegLite}{$-5$ to $0$}~\colorbox{dGray}{$0$--$5$}~\colorbox{dPos0}{$5$--$15$}~\colorbox{dPos1}{$15$--$30$}~\colorbox{dPos2}{$30$--$60$}~\colorbox{dPos3}{$\geq\!60$}).
See Appendix~\ref{app:appA1} for heatmap visualization.}
    \label{tab:context-delta}
    \end{table}




\subsection{SP1: The aggregation bottleneck}
\label{sec:sp1}

We index findings along two complementary axes: SP1 (task $\times$ strategy, this subsection) and SP2 (backbone $\times$ strategy, \Cref{sec:sp2}). The two are entangled---GPT-4o-mini's $-7.8$~pp on T2 with summary memory is both an SP1 observation (T2 demands aggregation; summaries destroy it) and an SP2 one (GPT-4o-mini has the narrowest blind-to-full gap).

\begin{figure}[t]
\centering
\includegraphics[width=\columnwidth]{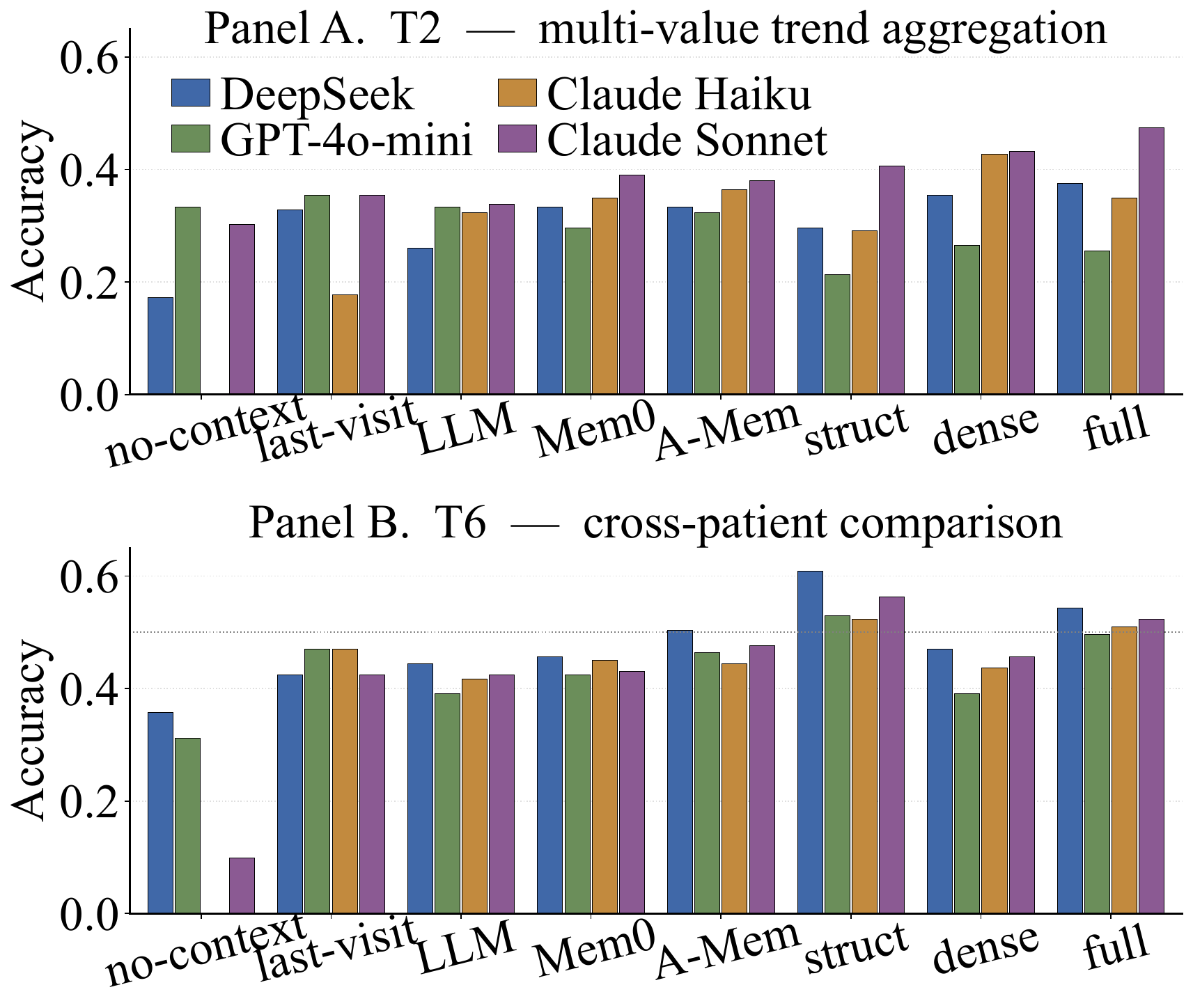}
\caption{Accuracy on two longitudinal-reasoning bottlenecks, grouped by strategy (x-axis) and backbone (color). (A) T2 trend aggregation (889 q/cell). (B) T6 cross-patient comparison (700 q/cell, subtypes T6a/T6b/T6c). Dotted line in Panel B = 0.5 reference (suppressed in A where all bars sit below 0.5). Missing Haiku no-context and Sonnet's near-zero on the same column are parser-strictness artifacts on verbose abstentions. Even \textit{full-context} stays well below ceiling: T2/T6 are reasoning bottlenecks distinct from point-fact recall.}
\label{fig:7}
\end{figure}

Compressed representations lose to \textit{full-context} because preprocessing discards signal needed at answer time. The penalty is sharpest on multi-value aggregation---T2 (trend across visits), T6b / T6c (max and delta across patients), and exploratory T8. On T8, \textit{full-context} reaches 0.269--0.470 across backbones while \textit{Mem0}, \textit{A-Mem}, and \textit{llm-summary} fall to 0.291--0.373, 0.291--0.373, and 0.194--0.254 respectively. T7 makes the contrast cleanest by rewarding single-fact recall over aggregation: \textit{full-context} reaches 0.942--0.965 (DeepSeek / Haiku / Sonnet), while \textit{Mem0} drops to 0.535--0.733 and \textit{llm-summary} to 0.349--0.512. \Cref{fig:7} visualizes the T2 / T6 bottleneck. The mechanism is structural: agentic-memory representations never store per-visit value pairs, and a 500-token \textit{llm-summary} cannot encode a multi-visit trajectory. Dense retrieval mitigates this---original text remains addressable---but still fails when aggregation crosses non-contiguous chunks.

\subsection{SP2: Backbone-specific extraction profiles}
\label{sec:sp2}

The blind-to-full gap diagnoses how much of the answer each backbone can extract from the chart when given full access:
\begin{itemize}[leftmargin=*, nosep]
    \item \textbf{Haiku}: $0.079 \to 0.706$, gap $+62.7$~pp (largest).
    \item \textbf{Sonnet}: $0.203 \to 0.698$, gap $+49.6$~pp.
    \item \textbf{DeepSeek-V3}: $0.236 \to 0.672$, gap $+43.6$~pp.
    \item \textbf{GPT-4o-mini}: $0.332 \to 0.630$, gap $+29.8$~pp (smallest).
\end{itemize}

Haiku exhibits the cleanest context utilization: lowest blind (0.079), tied top under \textit{full-context}, largest 62.7~pp lift. Sonnet and DeepSeek-V3 occupy the middle. GPT-4o-mini is the outlier: highest blind (0.332), lowest \textit{full-context} (0.630), narrowest lift---a pattern consistent with a hallucination floor that subsequent SP3 evidence on T9 reinforces. ``How much a model uses given context'' therefore varies more across backbones than overall accuracy suggests, and backbone choice cannot be reduced to a single capacity dimension.

\subsection{SP3: Non-monotonic abstention under longer context}
\label{sec:sp3}

\begin{figure}[t]
\centering
\includegraphics[width=\columnwidth]{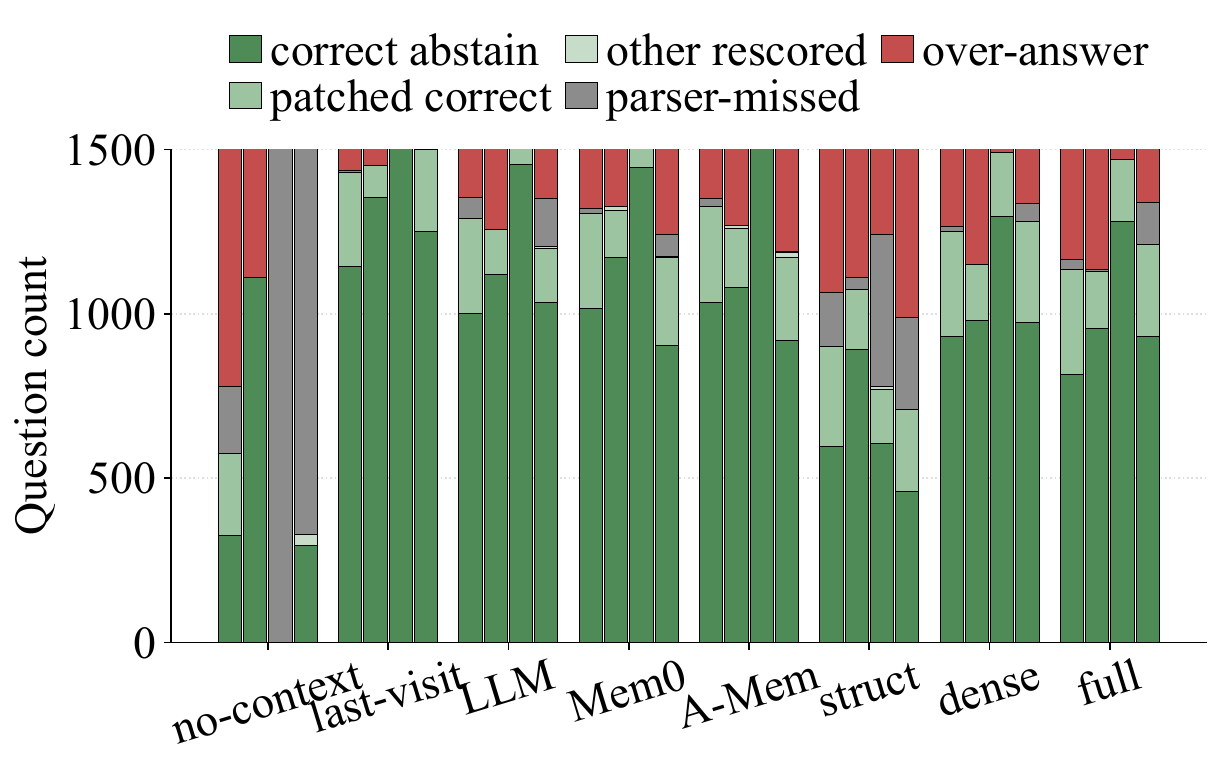}
\caption{T9 per-cell decomposition into five outcomes: 
correct abstain (refused as expected), 
patched correct / other rescored (post-hoc parser recovered abstention via synonyms), 
parser-missed (verbose abstention not captured), 
over-answer (confident hallucination); 
1500 q/cell, bars are DeepSeek/GPT-4o-mini/Haiku/Sonnet left to right within each strategy cluster. 
Abstention rate $=$ correct + patched + parser-missed. 
Over-answer grows from \textit{last-visit-only} to \textit{full-context} and \textit{structured-timeline}: abstention is non-monotonic in context length.}
\label{fig:5}
\end{figure}

T9 probes abstention over unstated facts; all 1{,}500 gold answers are ``insufficient information'', so accuracy equals the abstention rate and over-answer rate $= 1 - \text{abstention}$. Per-cell:
\begin{itemize}[leftmargin=*, nosep]
    \item \textit{no-context-blind}: abstention 0.355 / 0.685 / 0.000 / 0.204 (DeepSeek / GPT-4o-mini / Haiku / Sonnet); over-answer 0.645 / 0.315 / 1.000 / 0.796.
    \item \textit{full-context}: abstention 0.701 / 0.698 / 0.907 / 0.747; over-answer 0.299 / 0.302 / 0.093 / 0.253.
    \item \textit{last-visit-only}: abstention 0.883 / 0.895 / 0.994 / 0.926; over-answer 0.117 / 0.105 / 0.006 / 0.074.
\end{itemize}
Two findings. First, \textit{full-context} beats blind on every backbone---any patient context lifts abstention. Second, abstention is non-monotonic: \textit{last-visit-only} dominates \textit{full-context} on every backbone by 5--30~pp (Haiku 0.994 vs.\ 0.907; Sonnet 0.926 vs.\ 0.747; DeepSeek 0.883 vs.\ 0.701; GPT-4o-mini 0.895 vs.\ 0.698). Adding dialogue beyond the most-recent encounter actively degrades abstention. \Cref{fig:5} shows the over-answer fraction climbing from \textit{last-visit-only} through \textit{full-context} to \textit{structured-timeline}. GPT-4o-mini's anomalous blind T9 of 0.685 (vs.\ its pooled-blind 0.332) reflects a default abstention prior under empty context---a backbone-specific blind hedging. Per the trivial-baseline caveat (§3), T9 is a relative ordering, not an absolute measure.
The \textit{last-visit-only} advantage is significant on paired data (0.924 vs.\ 0.763 pooled; $\Delta = 0.161$, McNemar $p = 1.7\times10^{-38}$) and survives parser strictness: a lenient parser leaves it ahead (0.930 vs.\ 0.838), narrowing the gap to 0.092 without changing direction. Of the 307 \textit{full-context} non-abstentions, fabricated values make up $\approx 54\%$ of genuine over-answers once parser-missed verbose abstentions are discounted---plausible-but-unsupported detail, not cross-visit confusion, is the dominant mode (Appendix~\ref{app:newresults}).
\subsection{SP4: Controlled preservation probe: compressed representations drop the finding--diagnosis relation even under equal input}
\label{sec:sp4}

T3 isolates preprocessing-induced signal loss. Each ``yes'' instance carries one physician-attribution sentence linking a same-encounter finding to its problem; ``no'' instances are unmodified. We report the equal-input setting first, since it is the one in which every representation receives the same text.

\textbf{Equal input (preservation probe).} On a stratified 20-patient subset of the cohort---47 T3 questions, 19 injected-positive and 28 negative---the attribution sentence is inserted \emph{before} \textit{Mem0}, \textit{A-Mem} or the 500-token \textit{llm-summary} is constructed, so every constructor sees it. \Cref{tab:sp4-equal} gives positive-class recall on the 19 positives: at best 1/19 ($5.3\%$) for \textit{Mem0} and \textit{A-Mem}, and 0/19 for every \textit{llm-summary} cell. Rebuilding the memory with the answering backbone rather than DeepSeek-V3 changes nothing (lower block), so neither injection order nor prep-model mismatch explains it.

Artifact inspection shows what is lost: the underlying lab fact is usually retained, the explicit finding--diagnosis attribution usually omitted. The failure is \emph{relation-level}, not \emph{access-level}---compression keeps the facts and drops the link.

\begin{table}[t]
\centering
\small
\setlength{\tabcolsep}{4.5pt}
\begin{tabular}{lcccc}
\toprule
& DS & GPT & Hk & Sn \\
\midrule
\multicolumn{5}{l}{\textit{DeepSeek-V3--prepared}} \\
\textit{Mem0}         & 0.053 & 0.053 & 0.000 & 0.000 \\
\textit{A-Mem}        & 0.053 & 0.053 & 0.000 & 0.000 \\
\textit{llm-summary}  & 0.000 & 0.000 & 0.000 & 0.000 \\
\midrule
\multicolumn{5}{l}{\textit{Backbone-matched prep}} \\
\textit{Mem0}         & --- & 0.053 & 0.000 & 0.000 \\
\textit{A-Mem}        & --- & 0.053 & 0.000 & 0.000 \\
\bottomrule
\end{tabular}
\caption{SP4 \textbf{equal-input} preservation probe: positive-class recall on the 19 injected-positive T3 instances (stratified 20-patient subset), with the attribution sentence present before the representation is built. DS / GPT / Hk / Sn = DeepSeek-V3 / GPT-4o-mini / Haiku~4.5 / Sonnet~4.6; for DS the DeepSeek-prepared cell is already backbone-matched. \textit{structured-timeline} was not run in this setting (\textit{Scope}).}
\label{tab:sp4-equal}
\end{table}

\begin{figure}[t]
\centering
\includegraphics[width=\columnwidth]{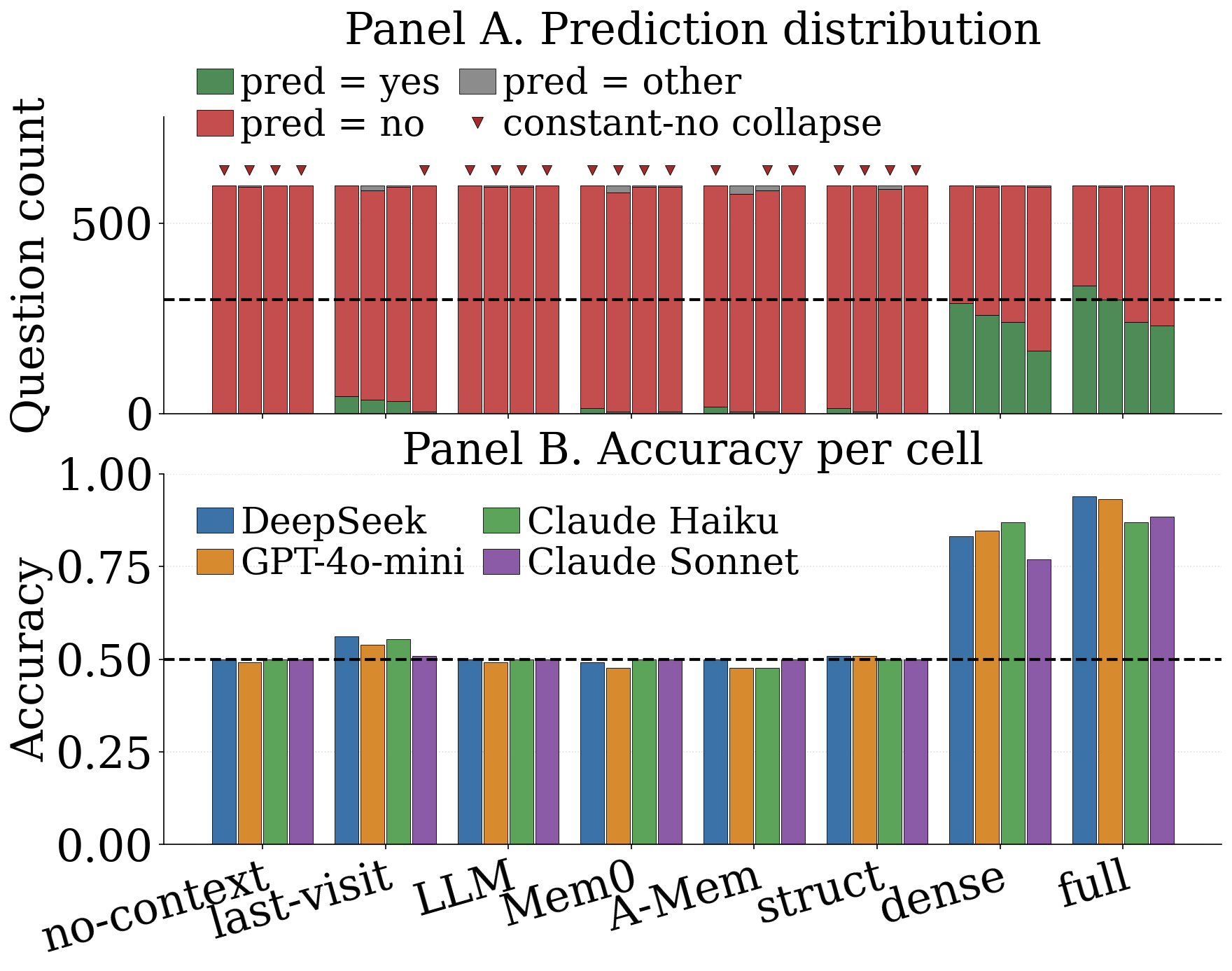}
\caption{T3 constant-No collapse under \textbf{post-construction} injection (update/staleness probe). 
(A) Per-cell yes/no/other predictions (600 q/cell); dashed line at 300 = balanced gold; 
red triangles flag cells with pred~$=$~no $\geq 575/600$. (B) Per-cell accuracy; dashed 0.5 = constant-No floor. 
The four upstream-built strategies flatten to 0.50 across all 16 cells. 
Primary preservation evidence is the equal-input probe (\Cref{tab:sp4-equal}).}
\label{fig:6}
\end{figure}

\textbf{Post-construction injection (update/staleness probe).} Injecting the sentence \emph{after} construction leaves \textit{full-context} (0.869--0.938), \textit{last-visit-only} and \textit{dense-retrieval} (0.769--0.869) intact, while all four upstream-built strategies answer ``No'' on all 600 T3 questions in all 16 cells, giving exactly 0.500 on the balanced gold---the constant-No floor of \Cref{fig:6}. A derived representation cannot contain a sentence introduced after it was built, so this measures staleness under update, not discarding; we report it as a secondary probe.

Neither setting depends on backbone capacity: GPT-4o-mini, the weakest backbone overall, still reaches 0.931 post-construction under full context, and under equal input the strongest backbones recover no more of the signal than the weakest.

\textbf{Scope.} The equal-input conclusion covers \textit{Mem0}, \textit{A-Mem} and \textit{llm-summary} only; \textit{structured-timeline} has not been run under pre-construction injection and enters only under the post-construction setting. Neither setting is an apples-to-apples ranking across all eight strategies, and what the probe measures is signal preservation, not unsupervised linkage reasoning.

\subsection{Retrieval, cost, and Pareto efficiency}
\label{sec:cost}

\begin{figure}[t]
\centering
\includegraphics[width=\columnwidth]{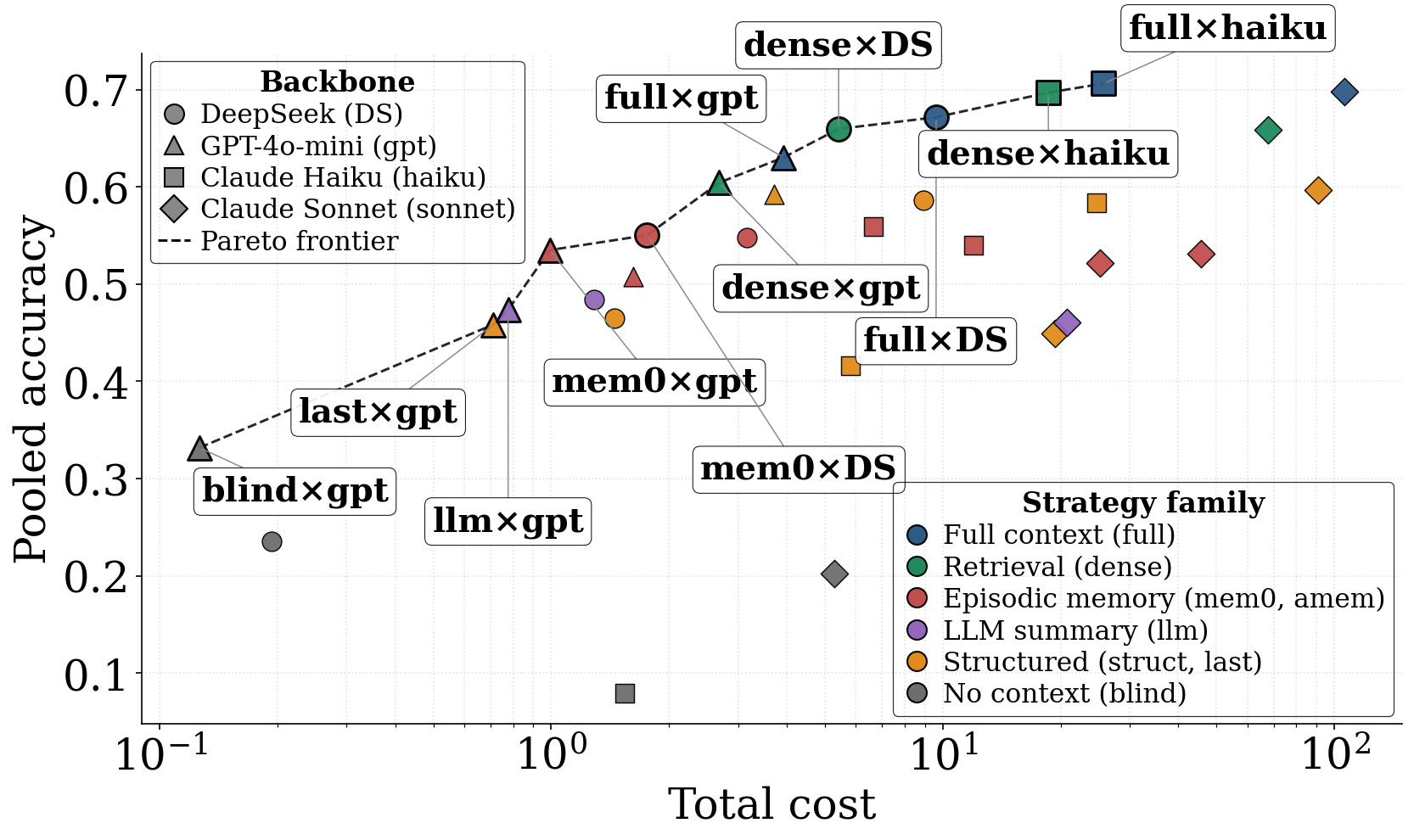}
\caption{Cost vs.\ pooled accuracy across 32 cells. Cost: USD per 6{,}271-question cell (log scale). Color = strategy family; shape = backbone. Dashed curve: Pareto frontier (11/32). Haiku dominates Sonnet on \textit{full-context}; \textit{structured-timeline} is off-frontier on cost despite competitive accuracy.}
\label{fig:8}
\end{figure}

Dense retrieval matches \textit{full-context} accuracy at a fraction of the cost. BGE-M3 hit rates over 4{,}771 retrieval-eligible questions: hit@1 = 0.595, hit@3 = 0.852, hit@5 = 0.926, hit@10 = 0.979 (Appendix~\ref{app:appA10}). \Cref{fig:8} places every cell in (cost, accuracy) space and traces the 11-cell Pareto frontier. Two consequences. First, \textit{full-context} $\times$ Sonnet (\$106.21, 0.698) is dominated by \textit{full-context} $\times$ Haiku (\$25.76, 0.706), inverting the ``biggest backbone wins'' heuristic. Second, \textit{structured-timeline} never reaches the frontier despite competitive accuracy (0.58--0.60)---the structured tokens fail to compress enough to offset per-call cost.

\subsection{Illustrative failure modes}
\label{sec:cases}
Aggregate accuracy collapses categorically distinct errors into a single number. Four case studies in Appendix~\ref{app:appA12cases} (Figure~\ref{fig:9}) make the distinction concrete: T3 constant-No collapse (Case~1), T9 over-answering (Case~2), Haiku-specific T2 trend failure (Case~3), and T6 cross-patient confusion (Case~4). The underlying errors differ in kind---conflating them obscures the diagnostic signal that each representation strategy fails in a characteristic way.

\section{Discussion}
\label{sec:discussion}

\textbf{What compact representations get right.} Dense retrieval lands within 1--4~pp of \textit{full-context} pooled and within 0--9~pp on every task except T4 (small-$n$); when answers are chunk-localizable, BGE-M3 surfaces them reliably. Structured timelines beat \textit{full-context} on numeric retrieval (T1 0.886--0.923) by reducing noise around the cell. \textit{Mem0} and \textit{A-Mem} hold their own on T4 and T9 where compression incidentally aligns with gold density.

\textbf{What compact representations get wrong.} (i) Aggregation (T2/T6b/T6c, exploratory T8): preprocessing discards per-visit value pairs needed for trends/deltas. (ii) Relation preservation (T3): even with the attribution sentence present at construction time, \textit{Mem0}, \textit{A-Mem} and \textit{llm-summary} recover 0--5.3\% of injected positives while typically keeping the underlying lab fact---the loss is of the finding--diagnosis \emph{relation}, not of access. Post-construction, those strategies plus \textit{structured-timeline} sit at the constant-No floor, showing additionally that derived artifacts go stale. (iii) Contradiction (T5): trivial always-contradicts scores 91.9\%; no cell beats it. (iv) Non-monotonic abstention (T9): \textit{full-context} helps over blind but degrades vs.\ \textit{last-visit-only }on every backbone, and trivial always-abstain scores 100\% by construction.

\textbf{What the dialogue substrate contributes.} In a small \textit{full-context} pilot the model received either the dialogue or a bare structured-event table over the same events. The table keeps the events but does not explicitly represent the narrative relations T3, T4 and T5 target, so those tasks cannot be evaluated equivalently from it (Appendix~\ref{app:newresults}).

\textbf{Cost flips ``biggest backbone wins''.} On the Pareto frontier (\Cref{fig:8}), Haiku not Sonnet sits on it under \textit{full-context}: cheaper (\$25.76 vs.\ \$106.21) and slightly more accurate (0.706 vs.\ 0.698). Memory-equipped clinical agents should be benchmarked on the cost--quality frontier.

\textbf{Memory size $\neq$ accuracy.} Total \textit{Mem0} character budget is uncorrelated with patient-level T1 \textit{Mem0} accuracy (Pearson $r=-0.129$; Appendix~\ref{app:appA11}). Architectural choice (atomic facts vs.\ visit-level notes vs.\ prose) matters more than budget; more facts can introduce retrieval noise.

\textbf{Future work.} Multi-EHR replication, longer horizons, dynamic memory updates, evaluation on real longitudinal notes and conversations, and extending the equal-input probe to \textit{structured-timeline} and the full cohort. A future release will rebalance T5 and enlarge T8 to remove the trivial-class ceilings.

\section{Conclusion}
\label{sec:conclusion}

ClinTraceBench provides the first source-verifiable benchmark for longitudinal clinical reasoning over EHR-derived dialogues: 9 tasks, 385 verified dialogues, 8 history representation strategies, 4 frontier backbones, 6{,}271 questions, and 200{,}672 predictions. Three high-level findings emerge. First, full context remains the accuracy upper bound, and the gap to compressed representations is driven not by retrieval failure but by a preprocessing tax on relational signal: when the finding--diagnosis attribution sentence is present in the input \emph{before} the representation is built, compressed strategies still typically retain the underlying fact while dropping the relation, and no scaling of backbone capacity recovers it. Second, abstention discipline does not improve monotonically with context length, complicating the case for feeding longer charts. Third, the cost--quality Pareto frontier inverts the ``biggest backbone wins'' heuristic, with smaller backbones dominating under full-context.

\section*{Limitations}
\label{sec:limitations}

\textbf{Synthetic dialogues.} The 385 dialogues are LLM-generated from de-identified structured MIMIC-IV fields. What the 98.92\% audit and the deterministic verifier establish is \emph{source fidelity and gold correctness}, not conversational realism: key labs, admissions and discharges, procedures, and diagnoses are covered at $\geq 98\%$; hallucination checks on values, reference ranges, ICD codes, and drugs pass at 99.8--100\%; independent gold recomputation mismatches on $<0.5\%$ of items. Patients are also stratified by source-record complexity, but none of this shows that synthetic dialogues reproduce the ambiguity, redundancy, temporal inconsistency, and missing documentation of real clinical text. All eight strategies are compared over the same dialogues, so within-benchmark comparisons are supported, but transfer to real clinical conversations is not established and we do not claim it; validating whether the strategy ordering persists on real longitudinal notes and conversations is future work.

\textbf{Scope.} (i)~Single EHR source (MIMIC-IV, hospital-wide admissions rather than an ICU-only cohort); evaluation is English-only and generalization to other settings, languages, and pediatric populations is untested. (ii)~Single time horizon (385 dialogues); multi-year follow-up and dynamic memory updates are deferred. (iii)~T5 is controlled rather than natural-EHR contradiction; T8 measures post-treatment lab change rather than causal efficacy (concurrent-medication annotations were not retained); T9 covers unstated lab and medication facts only. (iv)~\textit{full-context} serves as a coverage upper bound, not a deployment target, given its cost and latency profile.

\textbf{Residual artifacts.} Approximately 1{,}400 empty API responses (0.7\% of 200{,}672; Appendix~\ref{app:diagnostic}) are retained in headline numbers: excluding them would inflate Sonnet / Haiku \textit{full-context} by at most $+0.2$~pp, well below all headline contrasts, and the sensitivity appendix confirms that every SP1--SP4 contrast holds ($\Delta < 0.5$~pp) under either policy. Roughly 13\% of Sonnet T8 rows hit the original 120-token cap before we extended it, so residual truncation may persist on a small subset. Long-context degradation is uniform across backbones ($-13$ to $-14$~pp from shortest to longest quartile), but length and disease severity remain entangled.

\textbf{Agentic-memory prep confound.} The headline cube fixes the prep model at DeepSeek-V3, so non-DeepSeek \textit{Mem0} / \textit{A-Mem} cells run a heterogeneous pipeline and SP2's gap on those cells partly conflates answering- and construction-side effects. We therefore ran six backbone-matched cells (GPT-4o-mini, Haiku, Sonnet $\times$ \textit{Mem0}, \textit{A-Mem}) on the stratified 20-patient subset: matching prep to the answering backbone did \emph{not} improve accuracy in any of the six ($\Delta$ from $-0.045$ to $-0.244$) and left the T3 attribution signal almost entirely lost ($\leq 1/19$ positives recovered), so the prep model does not explain the weak agentic-memory results. Being confined to that subset, this check corroborates the full-cohort ordering in \Cref{tab:main} only at that scale.

\textbf{Task-design limits and planned revisions.} Trivial baselines leave T5 / T6 / T8 / T9 not discriminable from majority-class predictors at $\alpha = 0.05$ in accuracy terms, so T1 / T3 / T4 / T7 carry the cleanest representation signal and the rest serve as supporting probes; macro-F1 recovers usable signal on T5 and T8 (Appendix~\ref{app:newresults}). T4's 53 questions leave it under-powered at a 27.2~pp minimum detectable effect. A future release will rebalance the T5 contradiction classes and enlarge the T8 pool; neither change is part of the present benchmark version.

\section*{Ethics statement}

\textbf{Data access and de-identification.} ClinTraceBench is derived from MIMIC-IV~\citep{johnson2023mimic}, a HIPAA-compliant, de-identified electronic health record dataset released under the PhysioNet Credentialed Health Data License; all authors completed the required CITI training and signed the PhysioNet Data Use Agreement before accessing it. No re-identification was attempted, and the synthetic dialogues are generated from structured fields only (dates, lab values, ICD codes, drugs), so they contain no protected health information.

\textbf{Scope and intended use.} The 385 verified dialogues are LLM-generated doctor--patient conversations synthesized from de-identified structured records; they are not real clinical communications and are not clinically validated. ClinTraceBench evaluates information retrieval and reasoning fidelity over EHR-derived dialogues: it does not measure clinical safety, diagnostic accuracy, treatment appropriateness, or patient outcomes, and must not be used as a proxy for clinical readiness, for clinical decision support, for deployment-oriented model training, or in any patient-facing application.

\textbf{Release and redistribution.} The question set with event-ID provenance and gold answers, all 200{,}672 predictions, the scoring harness, and the controlled-injection generator are released at \url{https://github.com/HathyHuimin/ClinTraceBench}. The dialogues inherit MIMIC-IV's redistribution restrictions: their text is not redistributed, and regenerating it requires PhysioNet credentialing and MIMIC-IV access.

\bibliography{custom}
\clearpage

%
%

\appendix

\section{Validation gates}
\label{app:validation}

Each (anchor, gold) pair passes through five deterministic levels (L0--L4)
  plus a final human spot-check (L5) before entering the 6{,}271-question
  evaluation set:

  \begin{itemize}[leftmargin=*, nosep]
      \item \textbf{L0 --- Schema.} JSONL fields and types match the
      task-specific schema; threshold 0 failure.
      \item \textbf{L1 --- Anchor source.} Anchors for T1--T8 must lie in
      \texttt{verified\_covered\_events}; T9 anchors must lie in
      \texttt{missed\_events}; threshold $<1\%$.
      \item \textbf{L2 --- Gold recompute.} The gold is independently
      regenerated from the spec and the raw structured fields, and must
      equal the candidate gold; threshold $<0.5\%$.
      \item \textbf{L3 --- No gold leakage.} The question text must not
      contain the gold answer (critical for T2 trend labels and T9
      abstention cues); threshold 0 failure.
      \item \textbf{L4 --- Distribution coverage.} Per-task disease,
      complexity-tertile, patient, and event-type distributions are
      checked against the construction targets; reported as a soft
      warning rather than a hard fail.
      \item \textbf{L5 --- Human spot-check.} A stratified sample of 370
      questions (30/task low-risk $\times$ 4 + 50/task high-risk
      $\times$ 5) was hand-audited by the authors; 366/370 =
      \textbf{98.92\%} agreement with the automated gold. 7/9 tasks
      scored 0\% FAIL; T7\,=\,2\% (distractor substring leak),
      T9\,=\,6\% (Secondary-pool textual templating). Both below the
      10\% fix threshold.
  \end{itemize}

L0--L4 are deterministic and run on all generated questions; L5 is a human verification pass on a stratified subsample.

\section{Per-task subtype diagnostics}
\label{app:subtypes}

The headline tables pool over every within-task subtype for compactness, but
each of our nine evaluation tasks has at least one principled axis along
which difficulty varies by construction. This appendix reopens those axes
one task at a time, so the reader can see which question variants drive the
aggregate task scores in the main paper. Three recurring patterns are worth
flagging up front. First, tasks whose subtypes are categorical anchors
(T3 attribution class, T7 medication vs procedure, T9 silent-evidence
kind) split cleanly into separate distributions. Second, tasks with an
explicit controlled-vs-natural design (T5) collapse onto that family axis
and the headline finding lives there. Third, tasks with a
magnitude $\times$ direction lattice (T8) need both axes to read off
difficulty; small-magnitude bins are categorically harder than larger ones
across every backbone.

\subsection{Subtype overview}
\begin{figure}[htbp]
\centering
\includegraphics[width=\columnwidth]{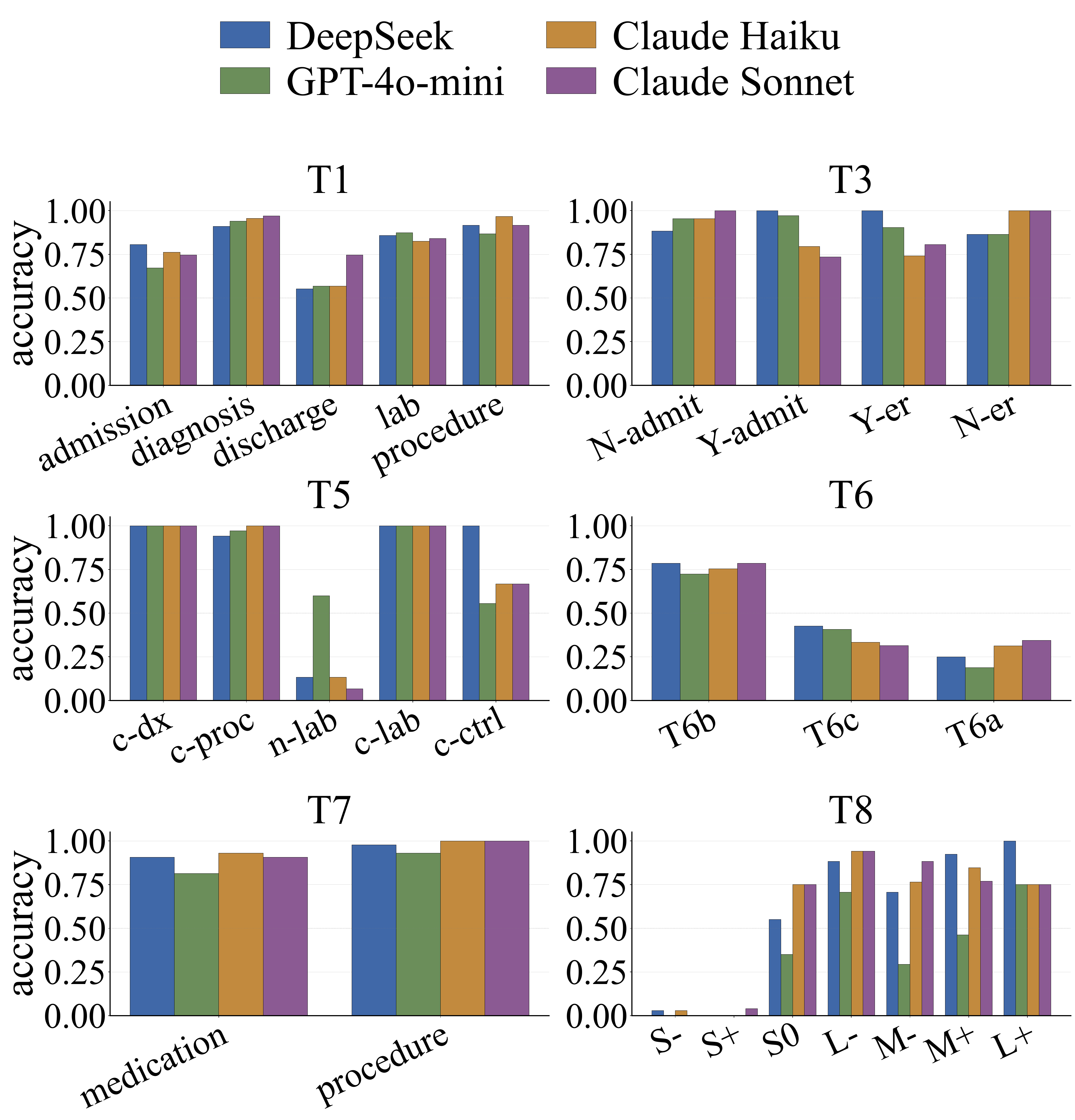}
\caption{Per-task subtype accuracy under full-context across four backbones.
T5 short codes: c-dx / c-proc / c-lab / c-ctrl $=$
controlled\_dx\_presence / \_procedure / \_lab\_value / \_control;
n-lab $=$ natural\_c2\_lab\_non\_glucose.
T8 short codes: L/M/S $=$ large/medium/small magnitude;
$-$/$+$/$0$ $=$ negative/positive/flat direction.}
\label{fig:appA2}
\end{figure}

\Cref{fig:appA2} is the six-panel overview: T1, T3, T5, T6, T7, T8 under
full-context across all four backbones. T2 / T4 / T9 are deferred to their
own dedicated figures below because their subtype axes do not match the
generic layout. Per-task subtype $n$ counts are listed inside the figure.

\subsection{T3 four-class attribution detail}
\begin{figure}[htbp]
\centering
\includegraphics[width=\columnwidth]{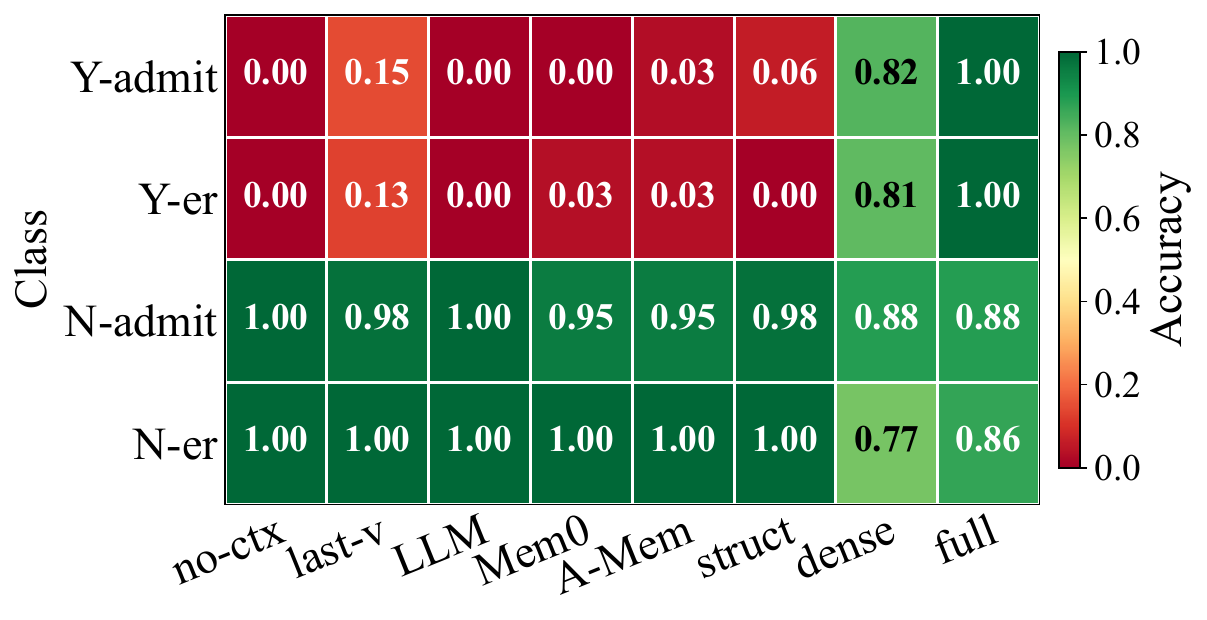}
\caption{T3 attribution-detection accuracy by 4-class stratification
(DeepSeek backbone, representative). $n$: Y-admit 34, Y-er 31, N-admit 43,
N-er 22.}
\label{fig:appA3}
\end{figure}

T3's design crosses injection presence (\textbf{Y} / \textbf{N}) with
encounter category (\textbf{admit} / \textbf{er}) to produce four classes.
The DeepSeek panel in \Cref{fig:appA3} is the representative split; the
other three backbones show the same qualitative pattern. The story is the
same in every backbone we tested: injected cells (Y-$*$) collapse to
${\sim}0$ under compressed memories and are recovered by \texttt{dense} and
\texttt{full-context}; not-injected cells (N-$*$) sit near ceiling for every
strategy except the dense / full-context strategies, where additional
context occasionally adds noise.

\subsection{T5 controlled vs.\ natural contradiction}
\begin{figure}[htbp]
\centering
\includegraphics[width=\columnwidth]{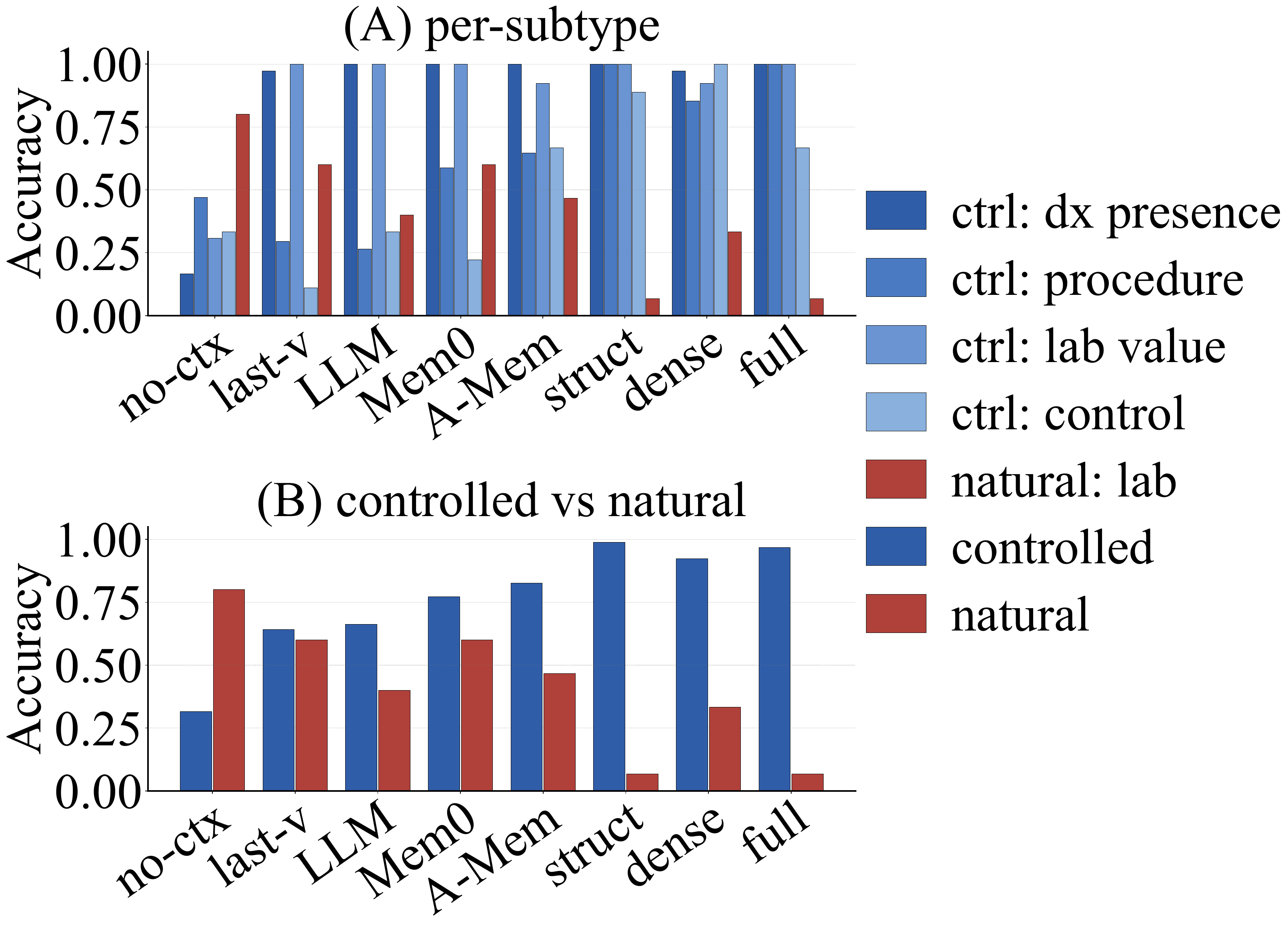}
\caption{T5 controlled-vs-natural contradiction breakdown (Sonnet
backbone exemplar). (A) per-subtype accuracy across the 8 memory baselines;
(B) the same data collapsed onto the controlled-vs-natural family axis.}
\label{fig:appA4}
\end{figure}

T5 is designed as a controlled contradiction probe rather than a natural
EHR-contradiction mining task. \Cref{fig:appA4} makes that explicit:
the five subtypes collapse cleanly onto a single controlled-vs-natural
axis -- four controlled subtypes against a single natural one -- and the
two families move in opposite directions as memory becomes richer. Controlled subtypes
climb from $0.32$ at \texttt{no-ctx} to $0.97$ under \texttt{full-context},
while the lone natural subtype (lab non-glucose) goes the other way
($0.80 \rightarrow 0.07$), because \texttt{dense} and \texttt{full-context}
introduce confounding lab evidence that triggers spurious contradictions.

\subsection{T4 retrospective evidence retrieval}
\begin{figure}[htbp]
\centering
\includegraphics[width=\columnwidth]{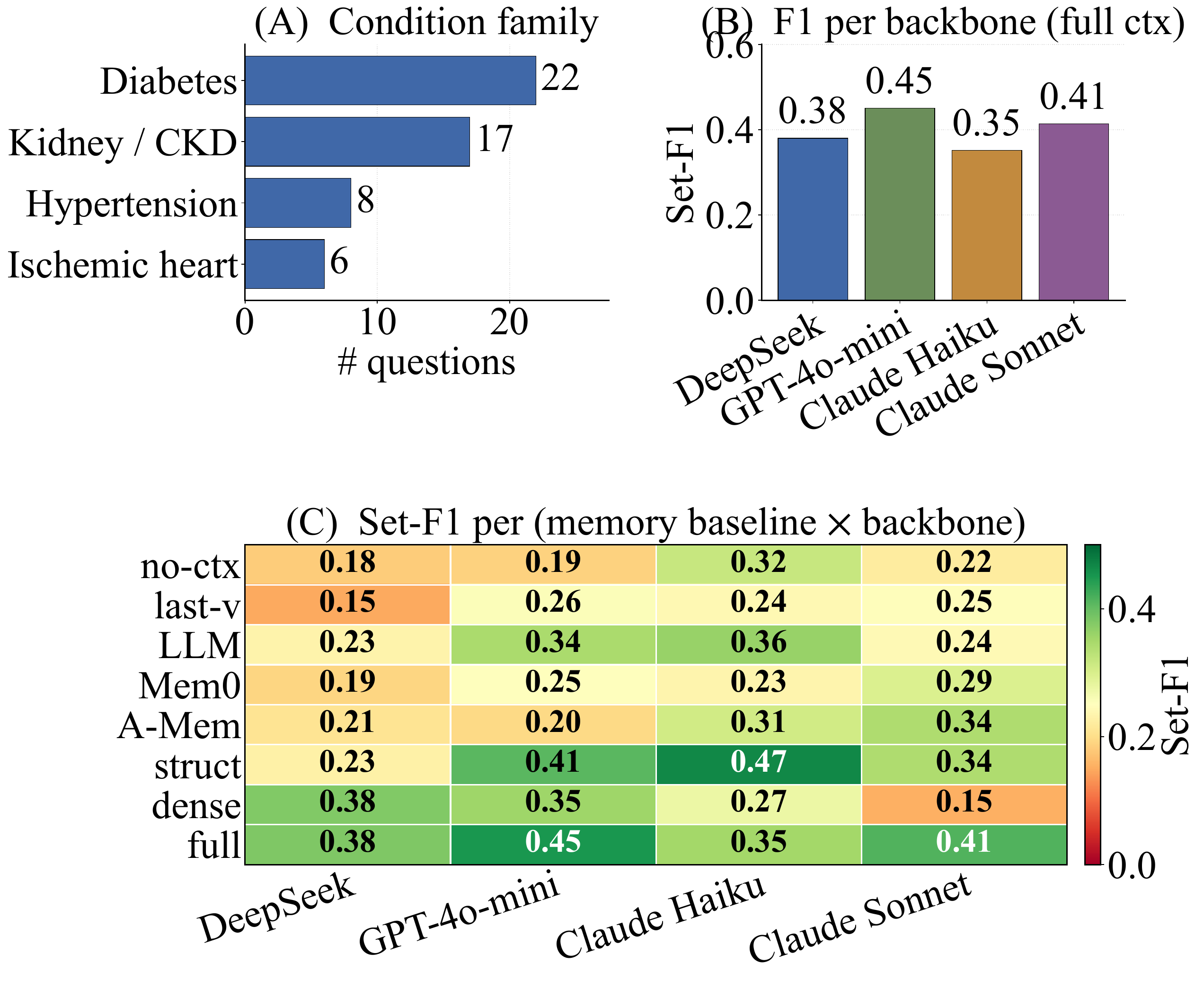}
\caption{T4 retrospective evidence retrieval. (A) condition-family
distribution of the $n=53$ questions; (B) per-backbone set-F1 under
full-context; (C) set-F1 heatmap across 8 memory baselines $\times$ 4
backbones.}
\label{fig:appA5}
\end{figure}

T4 scores strict set-F1 on event UIDs and is the smallest task in the suite
($n=53$); the question pool is biased toward a handful of condition
families. \Cref{fig:appA5} reports the pool composition, per-backbone
full-context set-F1, and the full $8 \times 4$ cell heatmap. The
small-$n$ profile is intentional and is the reason T4 is treated as a
limited rather than headline finding throughout the paper.

\subsection{T8 observed treatment-response bins}
\begin{figure}[htbp]
\centering
\includegraphics[width=\columnwidth]{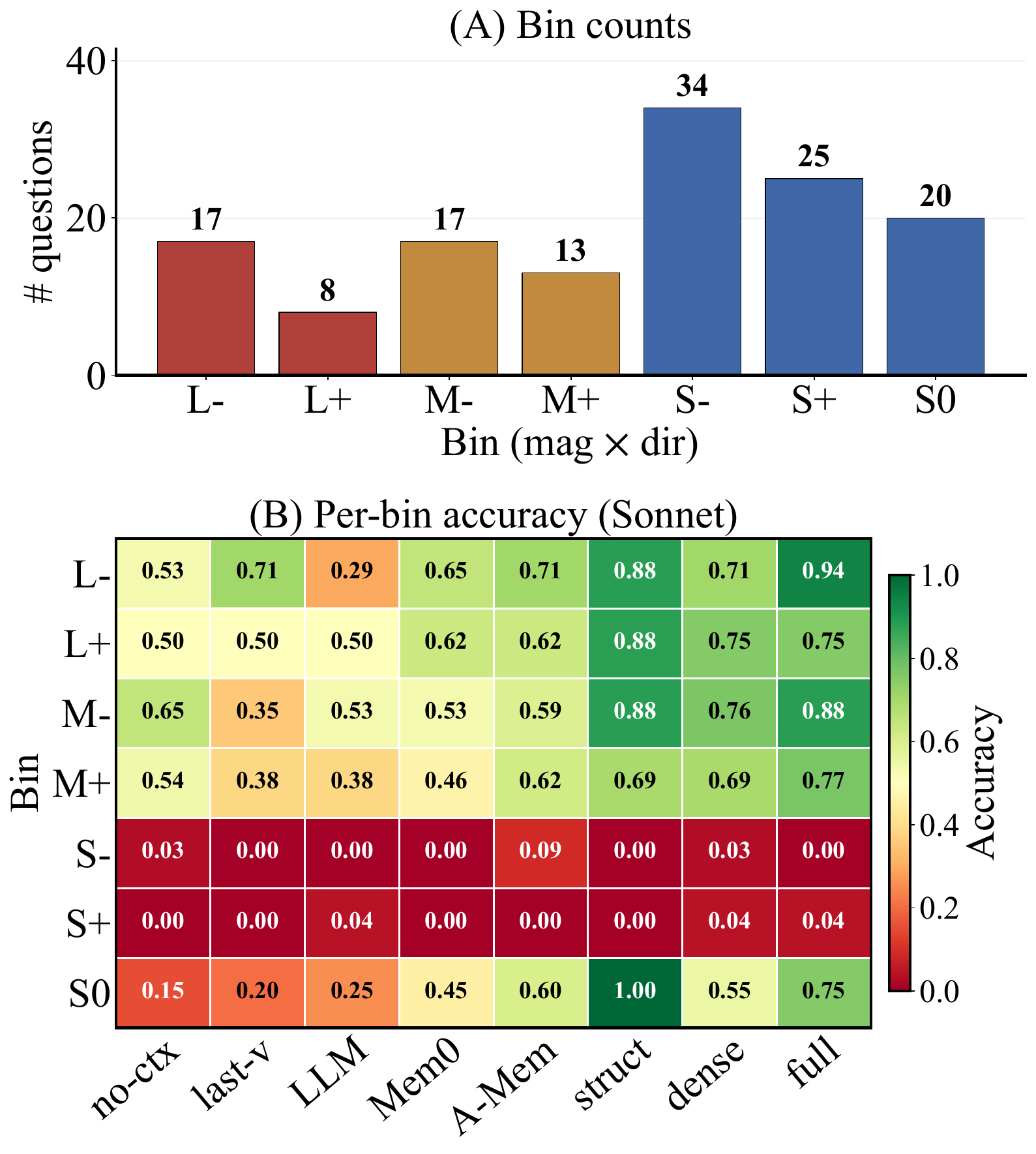}
\caption{T8 per-bin breakdown (Sonnet exemplar). (A) question counts in
the seven (magnitude $\times$ direction) bins ($n=134$ total);
(B) per-bin accuracy across the 8 memory baselines.}
\label{fig:appA6}
\end{figure}

T8 is observed post-treatment lab change tracking, not causal
treatment-response estimation; concurrent-medication annotations were
deliberately not retained in the final evaluation parquet, so the seven
magnitude $\times$ direction bins are a difficulty proxy rather than a
clinical-effect estimate. The headline pattern is that large/medium bins
climb from $0.29$--$0.65$ at \texttt{no-ctx} to $0.69$--$0.94$ at
\texttt{full-context}, but the small bins are categorically harder:
\textbf{S$-$} and \textbf{S$+$} sit at ${\sim}0$ across every baseline
including \texttt{full-context} (\Cref{fig:appA6}), while \textbf{S$0$} is rescued only by
\texttt{structured timeline} ($1.00$) and \texttt{full-context} ($0.75$).

\subsection{T9 silent-evidence kind: lab vs.\ other}
\begin{figure*}[t]
\centering
\includegraphics[width=\textwidth]{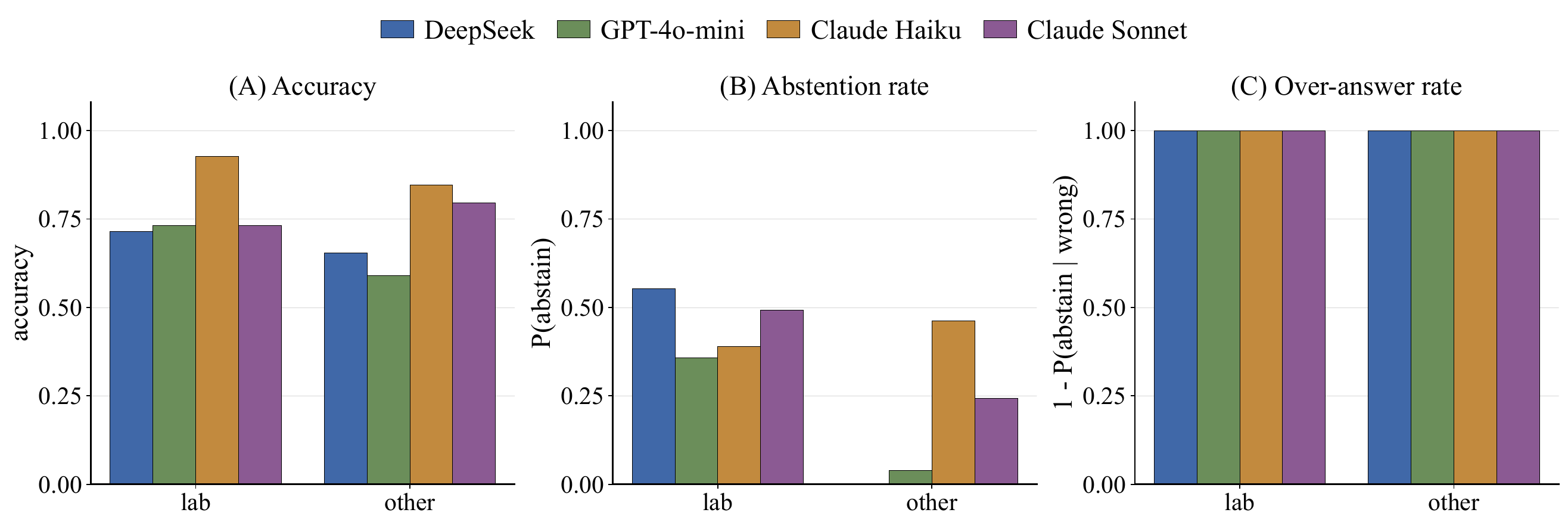}
\caption{T9 silent-evidence abstention stratified by anchor kind
(lab vs.\ other) under full-context. (A) accuracy; (B) abstention rate;
(C) over-answer rate when wrong. $n=1{,}147$ lab vs.\ $n=353$ other.}
\label{fig:appA7}
\end{figure*}

The active T9 pool is structurally degenerate along the
primary-or-secondary axis: every question has \texttt{anchor\_count}~$=$~1
and \texttt{primary\_or\_secondary}~$=$~``primary''. The cleanest available
stratification is therefore the silent-evidence kind (lab vs.\
other), and \Cref{fig:appA7} reports accuracy, abstention rate, and
over-answer rate on that split. The T9 pool splits
$1{,}147$ lab-anchored vs.\ $353$ other-anchored, an
$\approx 76\%/24\%$ ratio.

\section{Cohort, retrieval, and compression footprints}
\label{app:cohort}
\label{app:appA10}  

This appendix characterises the static building blocks of the benchmark:
the patient cohort it draws from, the auxiliary retrieval index it lets
the \texttt{dense} baseline rely on, and the compressed-memory structures
the \texttt{Mem0} / \texttt{A-Mem} / \texttt{LLM summary} baselines build
upstream of answer time. The point is to make clear that the headline
accuracy differences come from representation choice, not from
asymmetric access to the underlying clinical evidence.

\subsection{Patient cohort}
\begin{figure}[htbp]
\centering
\includegraphics[width=\columnwidth]{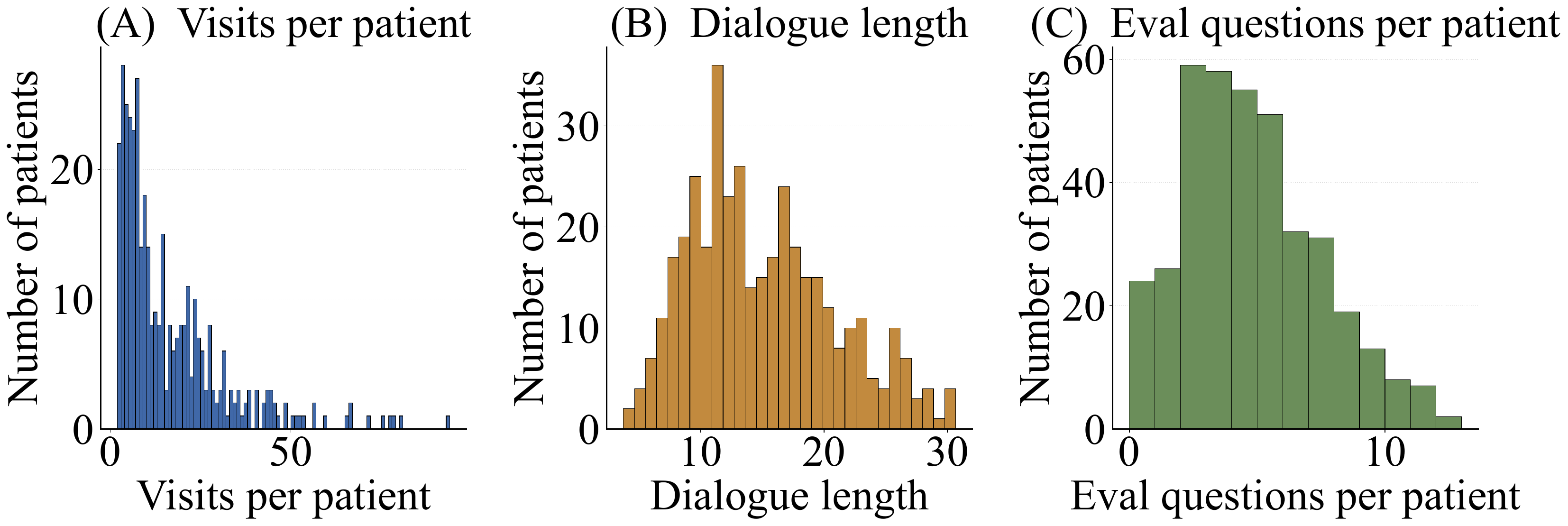}
\caption{Cohort statistics for the $n=385$ verified dialogues.
(A) visits per patient, (B) dialogue length, (C) eval questions per
patient.}
\label{fig:appA9}
\end{figure}

The 6{,}271-question evaluation set is drawn from $n=385$ verified
dialogues. \Cref{fig:appA9} reports the three distributional cuts that
matter: visits per patient (median 10, mean 16.0, max 93), dialogue length
in thousands of characters (median 13.7\,k, mean 15.0\,k, max 30.7\,k),
and stratified-eval questions per patient (median 4, mean 4.3, max 12).
The vast majority of the 385 dialogues contribute at least one question to
the eval set.

\subsection{BGE-M3 dense-retrieval hit rate}
\begin{figure}[htbp]
\centering
\includegraphics[width=\columnwidth]{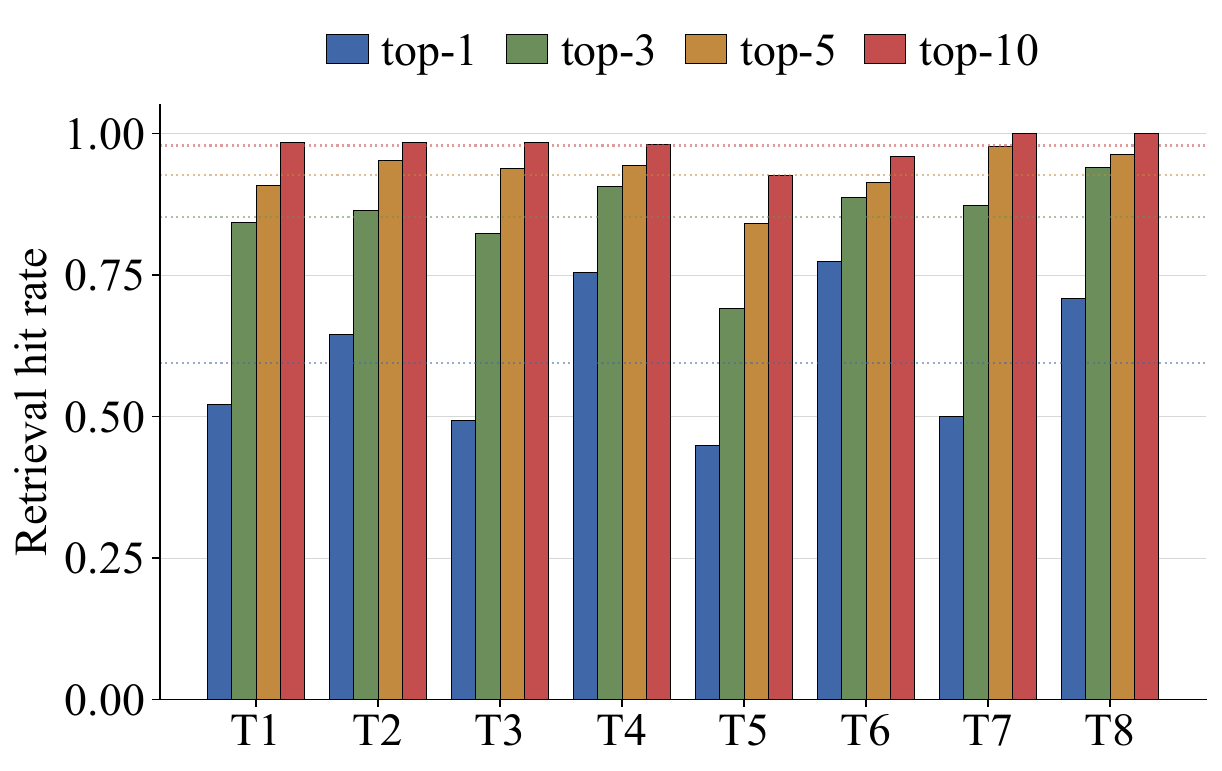}
\caption{BGE-M3 dense-retrieval hit rate by task on 4{,}771
retrieval-eligible questions. Per-task eligible $n$: T1 1500, T2 889, T3
600, T4 53, T5 495, T6 700, T7 400, T8 134. Dotted lines mark the overall
hit rate at each $k$.}
\label{fig:appA10}
\end{figure}

The \texttt{dense} baseline relies on BGE-M3 to embed visits and rank them
against the query. \Cref{fig:appA10} reports top-$k$ visit-hit rate per
task on the 1{,}177 retrieval-eligible questions (T9 silent-evidence is
excluded by construction). The overall headline is
hit@1 $=$ 0.595, hit@3 $=$ 0.852, hit@5 $=$ 0.926, hit@10 $=$ 0.979,
which is sufficient quality that the retrieval step is not the bottleneck
on any of the tasks where \texttt{dense} loses to \texttt{full-context}.

\subsection{Compression footprints}
\begin{figure}[htbp]
\centering
\includegraphics[width=\columnwidth]{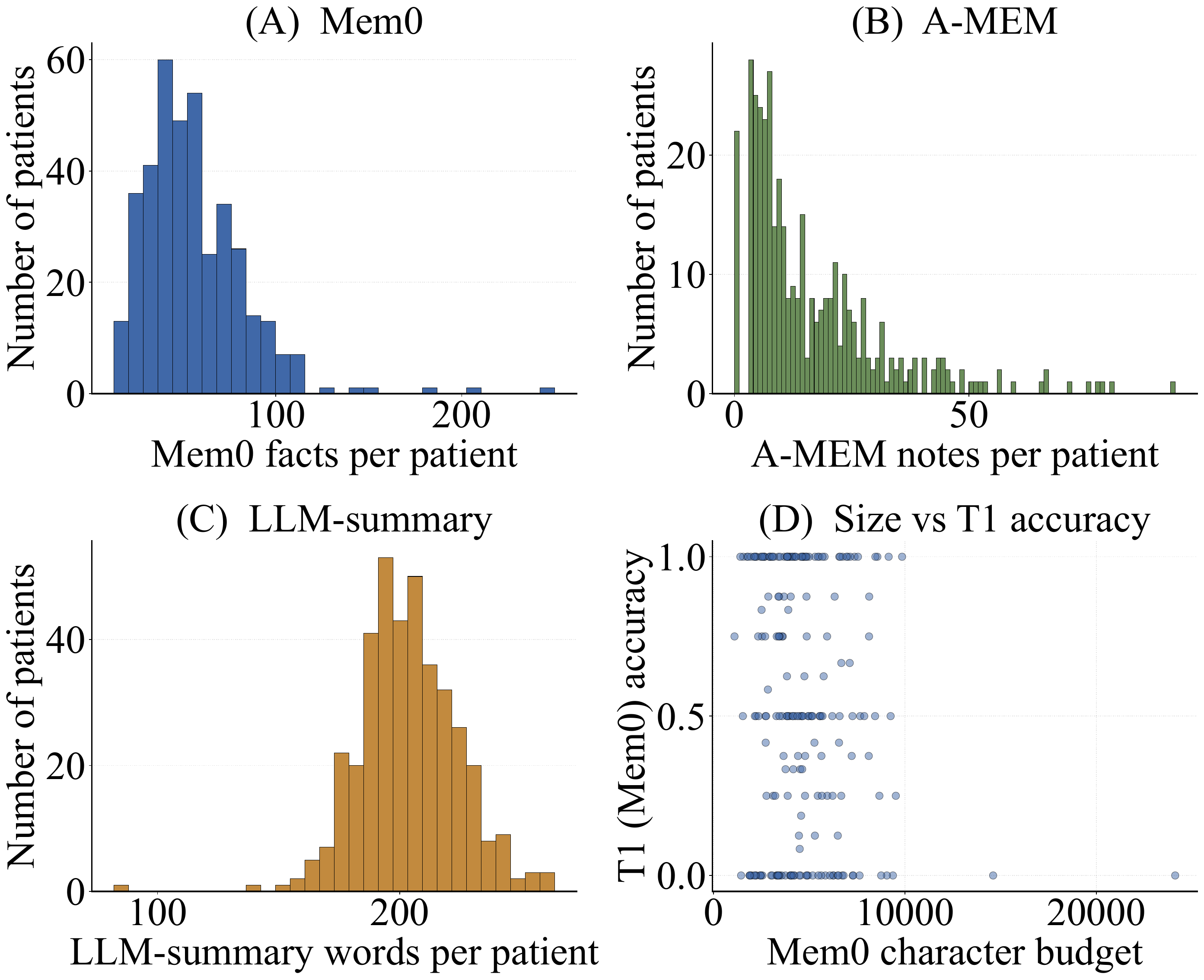}
\caption{Per-patient compression footprints. (A) \texttt{Mem0} facts
per patient, (B) \texttt{A-Mem} notes per patient, (C) \texttt{LLM-summary}
words per patient, (D) total \texttt{Mem0} size vs T1 accuracy
($n=210$, Pearson $r=-0.129$).}
\label{fig:appA11}
\label{app:appA11}  
\end{figure}

The three compressed-memory strategies produce structures of very
different sizes. \Cref{fig:appA11} reports the per-patient footprint of
each, plus a size-vs-accuracy scatter that disposes of the hypothesis
that headline accuracy differences are driven by raw representation size.
Pearson $r(\text{\texttt{Mem0} size}, \text{\texttt{Mem0} T1 accuracy}) = -0.129$
on $n=210$ patients: bigger is not better, and the compressed
representations are bottlenecked by what they encode, not by
how much.

\section{Resource consumption}
\label{app:resources}
\label{app:appA8}  

Tables and figures in this appendix attach a dollar / token / cache cost
to each of the 32 (memory baseline $\times$ backbone) cells. The
takeaway is the same one the main paper draws: most of the
representation-driven accuracy gains do not require the most expensive
cell, and the cost ranges across the grid are wide enough that
efficiency--accuracy trade-offs are a real decision rather than a marginal
one.

\begin{figure}[htbp]
\centering
\includegraphics[width=\columnwidth]{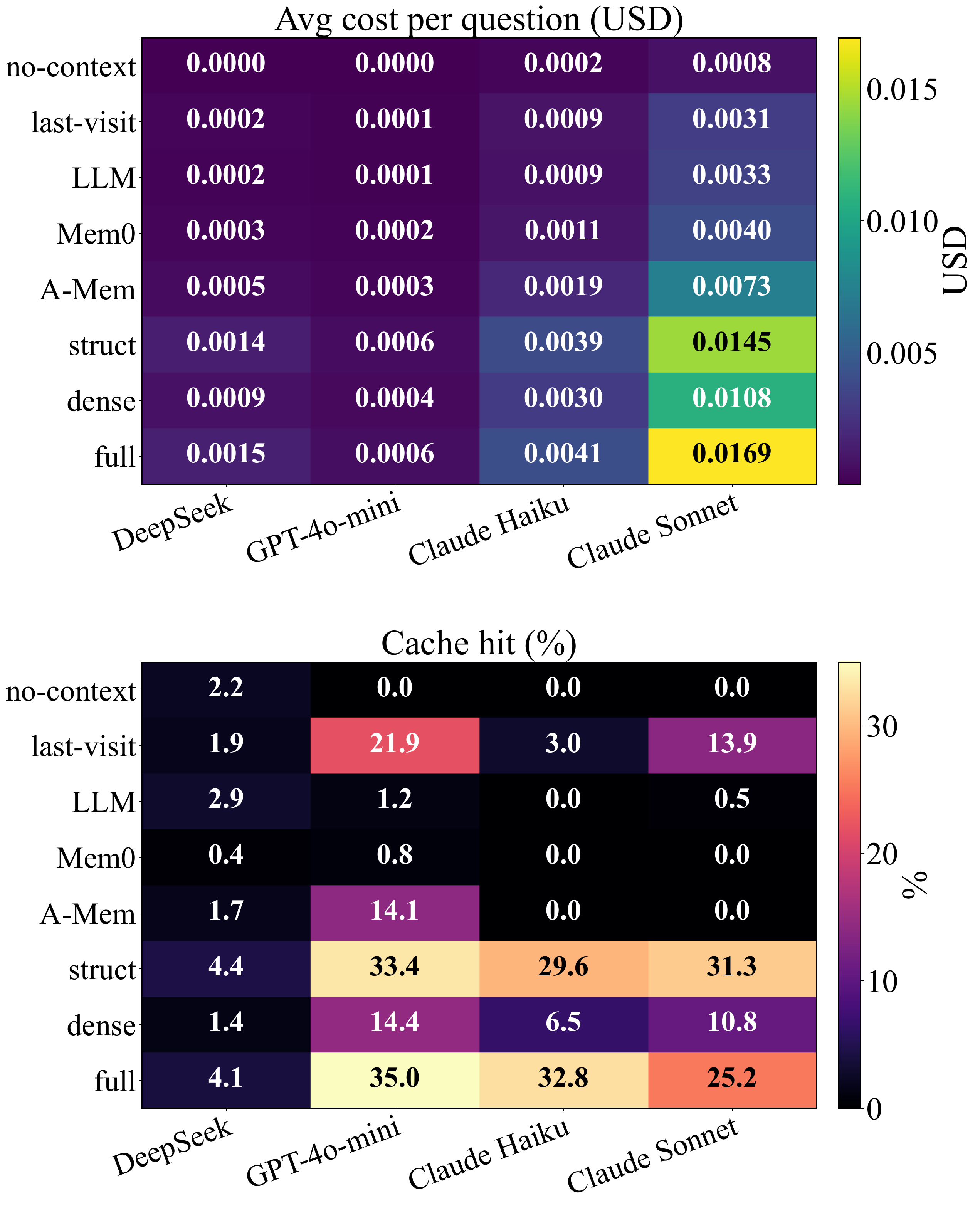}
\caption{Companion visualisation to \Cref{tab:appA8_cost_tokens_cache}:
avg cost per question (top) and cache hit \% (bottom) per (strategy,
backbone) cell. Sonnet $\times$ \texttt{full-context} is the most
expensive cell (\$0.0169/q); DeepSeek $\times$ \texttt{no-context blind}
is the cheapest (\$0.00003/q). Cache-hit range: 0\%--35\%.}
\label{fig:appA8}
\end{figure}

\Cref{tab:appA8_cost_tokens_cache} gives the per-cell figures and
\Cref{fig:appA8} visualises them.

\begin{table*}[t]
\centering
\small
\setlength{\tabcolsep}{4pt}
\begin{tabular}{l l r r r r r}
\toprule
Strategy & Backbone & Cost (USD) & \$/q & Input (M) & Output (K) & Cache hit \% \\
\midrule
no-context blind & DeepSeek-V3 & \$0.19 & \$0.0000 & 0.5 & 52.3 & 2.2\% \\
 & GPT-4o-mini & \$0.13 & \$0.0000 & 0.5 & 74.3 & 0.0\% \\
 & Haiku 4.5 & \$1.54 & \$0.0002 & 0.6 & 272.8 & 0.0\% \\
 & Sonnet 4.6 & \$5.27 & \$0.0008 & 0.6 & 238.1 & 0.0\% \\
last-visit-only & DeepSeek-V3 & \$1.45 & \$0.0002 & 5.3 & 40.4 & 1.9\% \\
 & GPT-4o-mini & \$0.71 & \$0.0001 & 5.5 & 81.9 & 21.9\% \\
 & Haiku 4.5 & \$5.80 & \$0.0009 & 6.4 & 208.1 & 3.0\% \\
 & Sonnet 4.6 & \$19.34 & \$0.0031 & 6.4 & 163.1 & 13.9\% \\
full-context & DeepSeek-V3 & \$9.62 & \$0.0015 & 36.8 & 51.3 & 4.1\% \\
 & GPT-4o-mini & \$3.92 & \$0.0006 & 37.7 & 78.8 & 35.0\% \\
 & Haiku 4.5 & \$25.76 & \$0.0041 & 44.7 & 137.9 & 32.8\% \\
 & Sonnet 4.6 & \$106.20 & \$0.0169 & 44.9 & 141.4 & 25.2\% \\
dense-retrieval & DeepSeek-V3 & \$5.42 & \$0.0009 & 20.1 & 52.2 & 1.4\% \\
 & GPT-4o-mini & \$2.68 & \$0.0004 & 20.2 & 75.2 & 14.4\% \\
 & Haiku 4.5 & \$18.61 & \$0.0030 & 24.0 & 131.4 & 6.5\% \\
 & Sonnet 4.6 & \$67.63 & \$0.0108 & 24.2 & 131.6 & 10.8\% \\
structured-timeline & DeepSeek-V3 & \$8.91 & \$0.0014 & 34.2 & 45.2 & 4.4\% \\
 & GPT-4o-mini & \$3.70 & \$0.0006 & 34.9 & 69.5 & 33.4\% \\
 & Haiku 4.5 & \$24.72 & \$0.0039 & 41.0 & 158.1 & 29.6\% \\
 & Sonnet 4.6 & \$90.83 & \$0.0145 & 41.2 & 133.2 & 31.3\% \\
LLM-summary & DeepSeek-V3 & \$1.29 & \$0.0002 & 4.7 & 50.3 & 2.9\% \\
 & GPT-4o-mini & \$0.78 & \$0.0001 & 4.9 & 78.9 & 1.2\% \\
 & Haiku 4.5 & \$5.49 & \$0.0009 & 5.9 & 187.9 & 0.0\% \\
 & Sonnet 4.6 & \$20.78 & \$0.0033 & 6.0 & 200.0 & 0.5\% \\
mem0 & DeepSeek-V3 & \$1.76 & \$0.0003 & 6.4 & 45.3 & 0.4\% \\
 & GPT-4o-mini & \$0.99 & \$0.0002 & 6.4 & 75.3 & 0.8\% \\
 & Haiku 4.5 & \$6.65 & \$0.0011 & 7.5 & 158.9 & 0.0\% \\
 & Sonnet 4.6 & \$25.16 & \$0.0040 & 7.7 & 145.0 & 0.0\% \\
amem & DeepSeek-V3 & \$3.17 & \$0.0005 & 11.7 & 48.1 & 1.7\% \\
 & GPT-4o-mini & \$1.62 & \$0.0003 & 12.0 & 76.1 & 14.1\% \\
 & Haiku 4.5 & \$11.99 & \$0.0019 & 14.1 & 168.1 & 0.0\% \\
 & Sonnet 4.6 & \$45.75 & \$0.0073 & 14.5 & 143.5 & 0.0\% \\
\midrule
\textbf{Total} & --- & \textbf{\$527.87} & --- & \textbf{532} & \textbf{3714} & \textbf{17.1\%} \\
\bottomrule
\end{tabular}
\caption{\textbf{Per-cell resource consumption over the 6{,}271-question evaluation set.} Each row is one (strategy, backbone) cell. Per-question cost ($\$/q$) is cost / 6{,}271. Total spend: \$527.87 across all 32 cells.}
\label{tab:appA8_cost_tokens_cache}
\end{table*}

\section{Pipeline-quality diagnostics}
\label{app:pipeline}
\label{app:diagnostic}  

This appendix documents the failure modes of the evaluation pipeline
itself. Two kinds of issue can in principle confound the headline
accuracies: a model returning no response at all, and the strict gold
parser failing to recover a prediction. We measure both, and we measure
where each one concentrates so the reader can see they are not driving
the comparisons in the main paper.

\subsection{Empty responses (D1)}
\begin{figure}[htbp]
\centering
\includegraphics[width=\columnwidth]{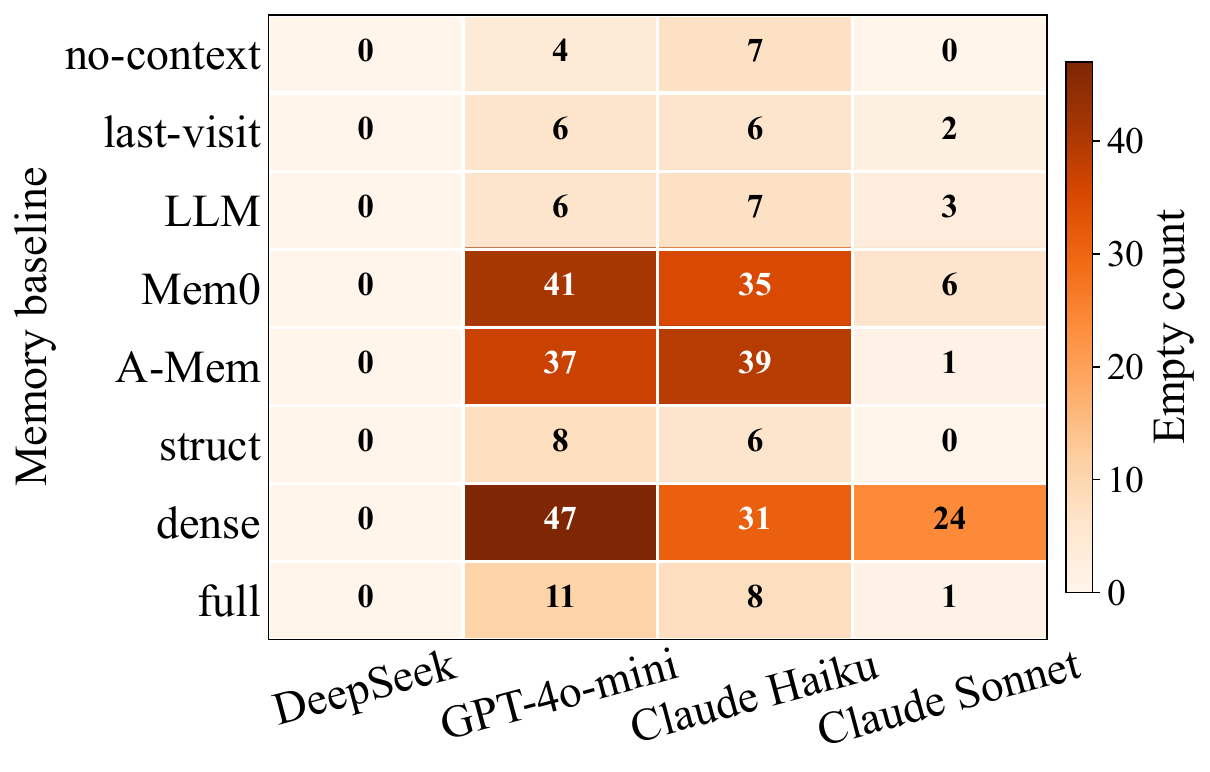}
\caption{D1: empty API responses per cell after the M1 retry pass
($\approx 0.7\%$ of 200{,}672 rows). DeepSeek's column is uniformly zero;
the mass sits on GPT-4o-mini and Haiku, with a small Sonnet residue.}
\label{fig:appD1}
\end{figure}

After the M1 retry pass, roughly $0.7\%$ of the 200{,}672 prediction
rows still have an empty \texttt{raw\_response}. They concentrate on
GPT-4o-mini and Claude Haiku under retrieval-style
contexts (\texttt{Mem0}, \texttt{A-Mem}, \texttt{dense});
DeepSeek's column is uniformly zero (\Cref{fig:appD1}). These rows are documented but
not excluded from the headline accuracy numbers.

\subsection{Parse failures (D2)}
\begin{figure}[htbp]
\centering
\includegraphics[width=\columnwidth]{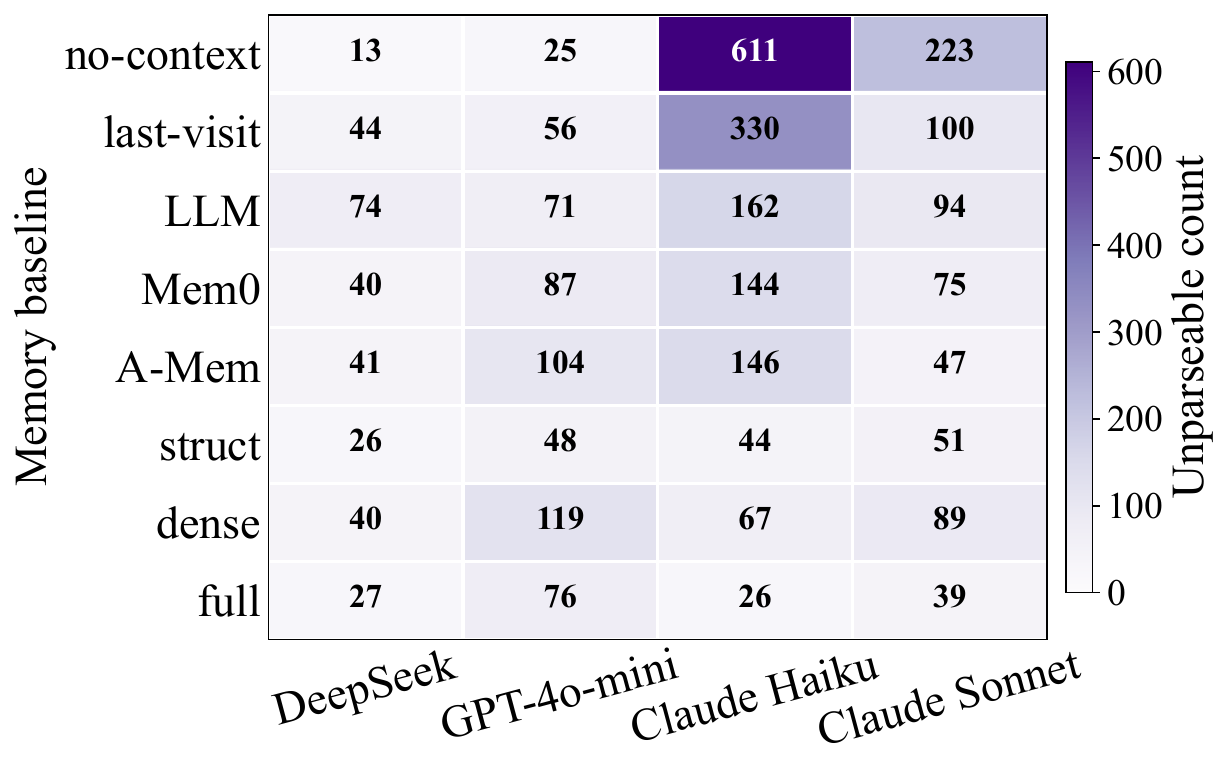}
\caption{D2: parse-failure rate (\texttt{pred is None}) per cell
across 8 memory baselines $\times$ 4 backbones (6{,}271 questions per
cell).}
\label{fig:appD2}
\end{figure}

A response can be present but un-parseable by the strict gold-format
matcher. About $6.5\%$ of the 200{,}672 rows fall into this bucket,
and they are dominated by Claude refusal-as-prose under
\texttt{no-context blind}; the Haiku $\times$ \texttt{no-context} cell is
the single largest contributor (\Cref{fig:appD2}). The pattern is concentrated on a small
number of cells and does not affect the comparative ordering between
memory strategies under richer contexts.

\subsection{Output length and truncation}
\begin{figure}[htbp]
\centering
\includegraphics[width=\columnwidth]{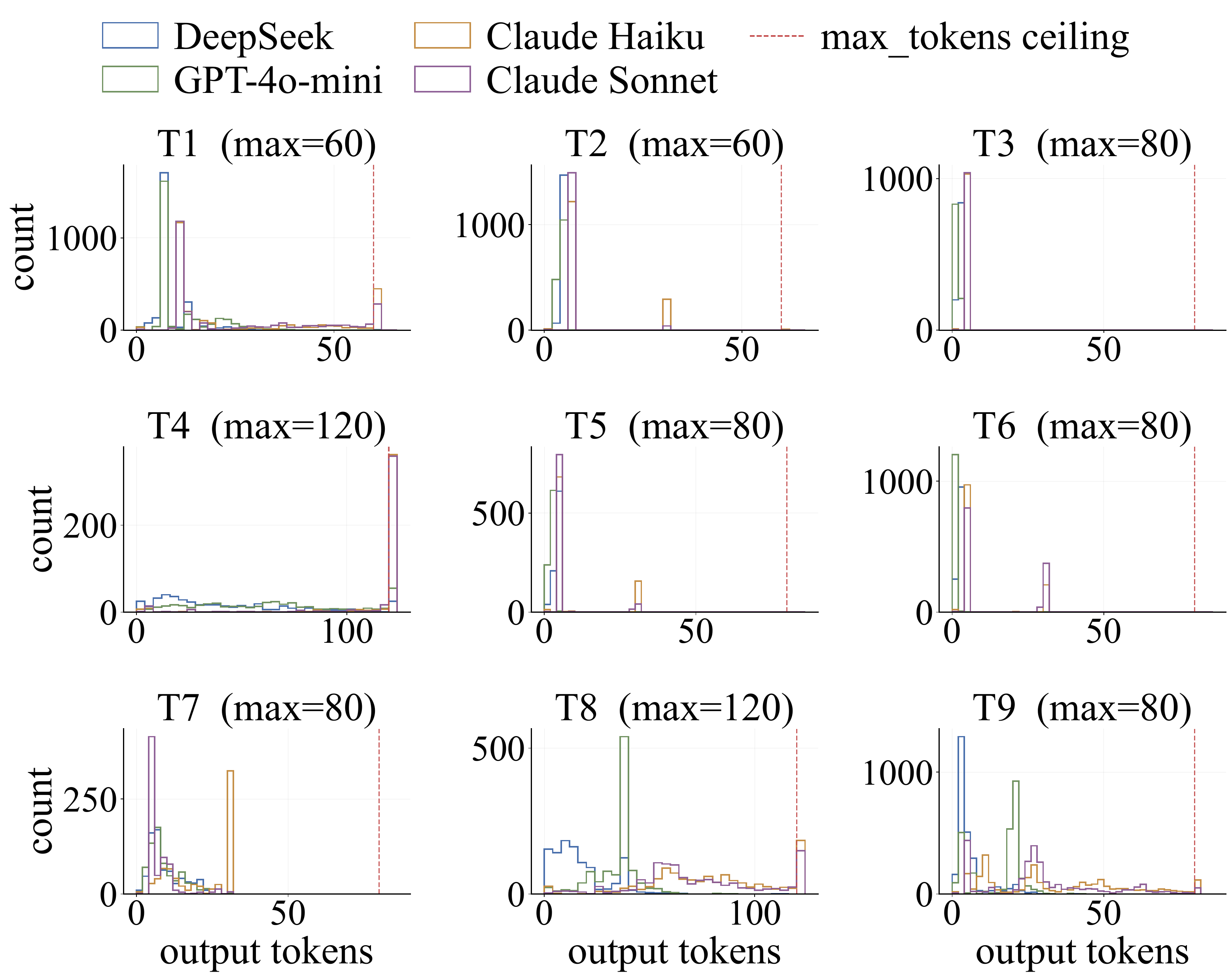}
\caption{D3: output-length histograms per task overlaid by backbone,
with the per-task \texttt{max\_tokens} ceiling marked.}
\label{fig:appD3}
\end{figure}

\begin{figure}[htbp]
\centering
\includegraphics[width=\columnwidth]{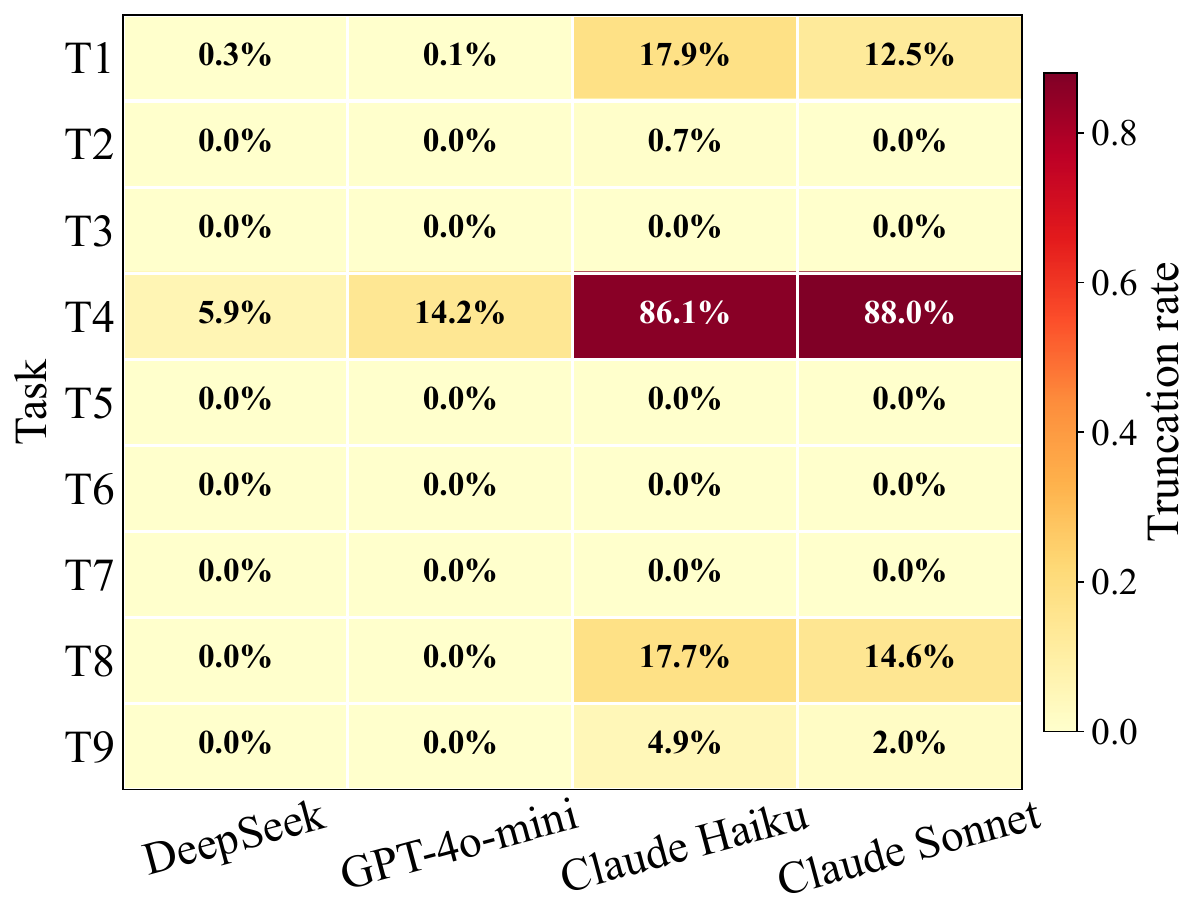}
\caption{D4: truncation rate per (task, backbone), mean across the 8
memory baselines. Per-task ceilings: T1 60, T2 60, T4 120, T8 120,
T9 80; others 80.}
\label{fig:appD4}
\end{figure}

The model is given a per-task \texttt{max\_tokens} budget. Two diagnostics
follow: the full output-length distribution per task overlaid by backbone
(\Cref{fig:appD3}), and the rate at which a response is truncated at
ceiling minus one token (\Cref{fig:appD4}). T4 is the only task where the
120-token ceiling binds in a meaningful way ($86$--$88\%$ truncation on
Haiku / Sonnet under the worst memory baseline); T1 and T8 are the next
largest binders. T3 / T5 / T6 / T7 are essentially un-truncated across the
board.

\section{Robustness and stratification}
\label{app:robustness}

The headline numbers in the main paper are point estimates averaged over
patients, diseases, and within-task difficulty tertiles. This appendix
reports the four sensitivity checks that we think matter most: per-patient
skew, per-disease drift, per-tertile calibration of the stratifier, and
bootstrap-CI width per cell.

\subsection{Per-patient accuracy distribution (D5)}
\begin{figure}[htbp]
\centering
\includegraphics[width=\columnwidth]{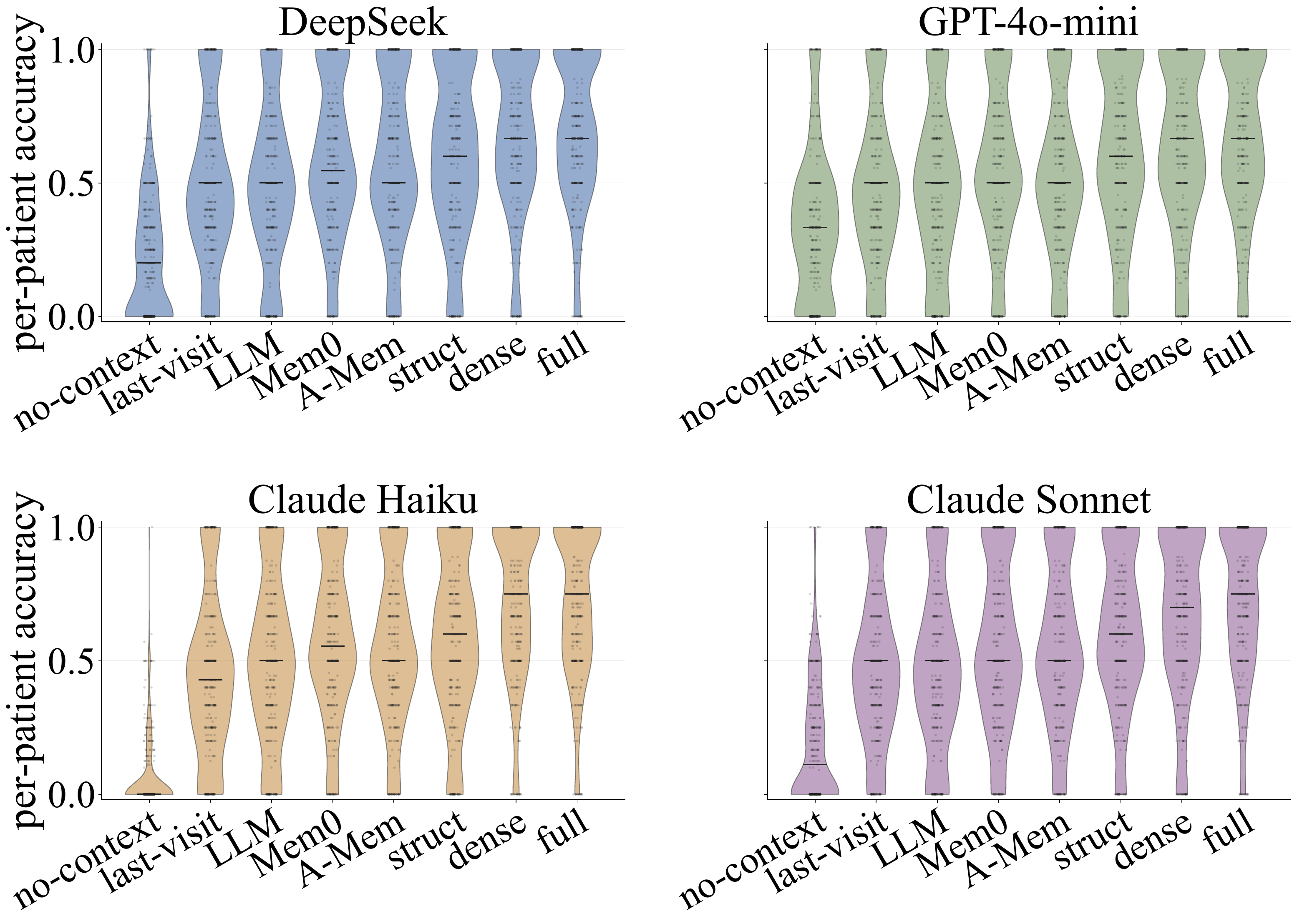}
\caption{D5: per-patient accuracy distribution as a violin per cell;
medians overlaid in black, patient-level scatter in grey.}
\label{fig:appD5}
\end{figure}

\Cref{fig:appD5} renders the per-cell distribution of per-patient mean
accuracy as a violin. The headline observation is that \texttt{full-context}
violins are uniformly narrower than the compressed-representation cells on
every backbone -- richer memory produces less between-patient skew, not
just a higher mean.

\subsection{Per-disease drift (D6)}
\begin{figure}[htbp]
\centering
\includegraphics[width=\columnwidth]{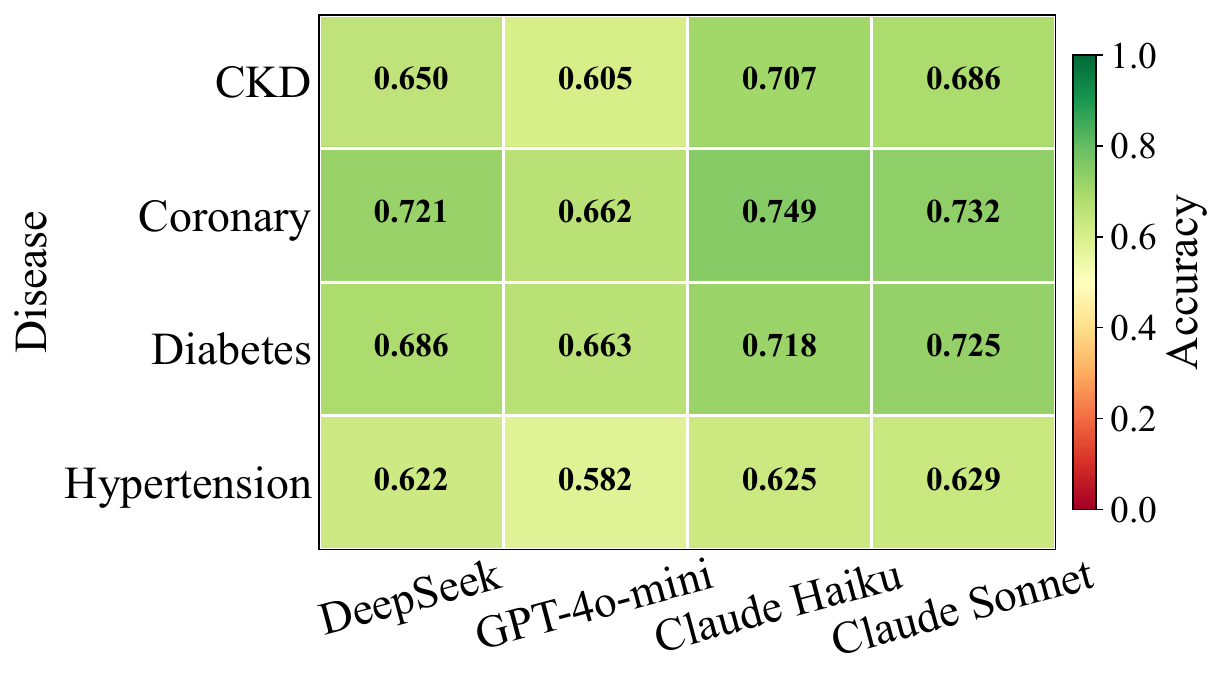}
\caption{D6: per-disease \texttt{full-context} accuracy by backbone,
pooled across T1--T9. Per-disease $n$ per backbone:
CKD 488, Coronary 358, Diabetes 404, Hypertension 251.}
\label{fig:appD6}
\end{figure}

The benchmark spans four condition families: CKD, Coronary, Diabetes, and
Hypertension. \Cref{fig:appD6} reports full-context accuracy pooled
across all nine tasks, by backbone $\times$ disease. The hypertension row
is consistently the weakest (mean $0.615$), but this is a task-mix bias --
the hypertension question pool over-indexes the longitudinal / comparison
families -- rather than evidence of a disease-specific knowledge gap.

\subsection{Per-tertile calibration of the stratifier (D7)}
\begin{figure}[htbp]
\centering
\includegraphics[width=\columnwidth]{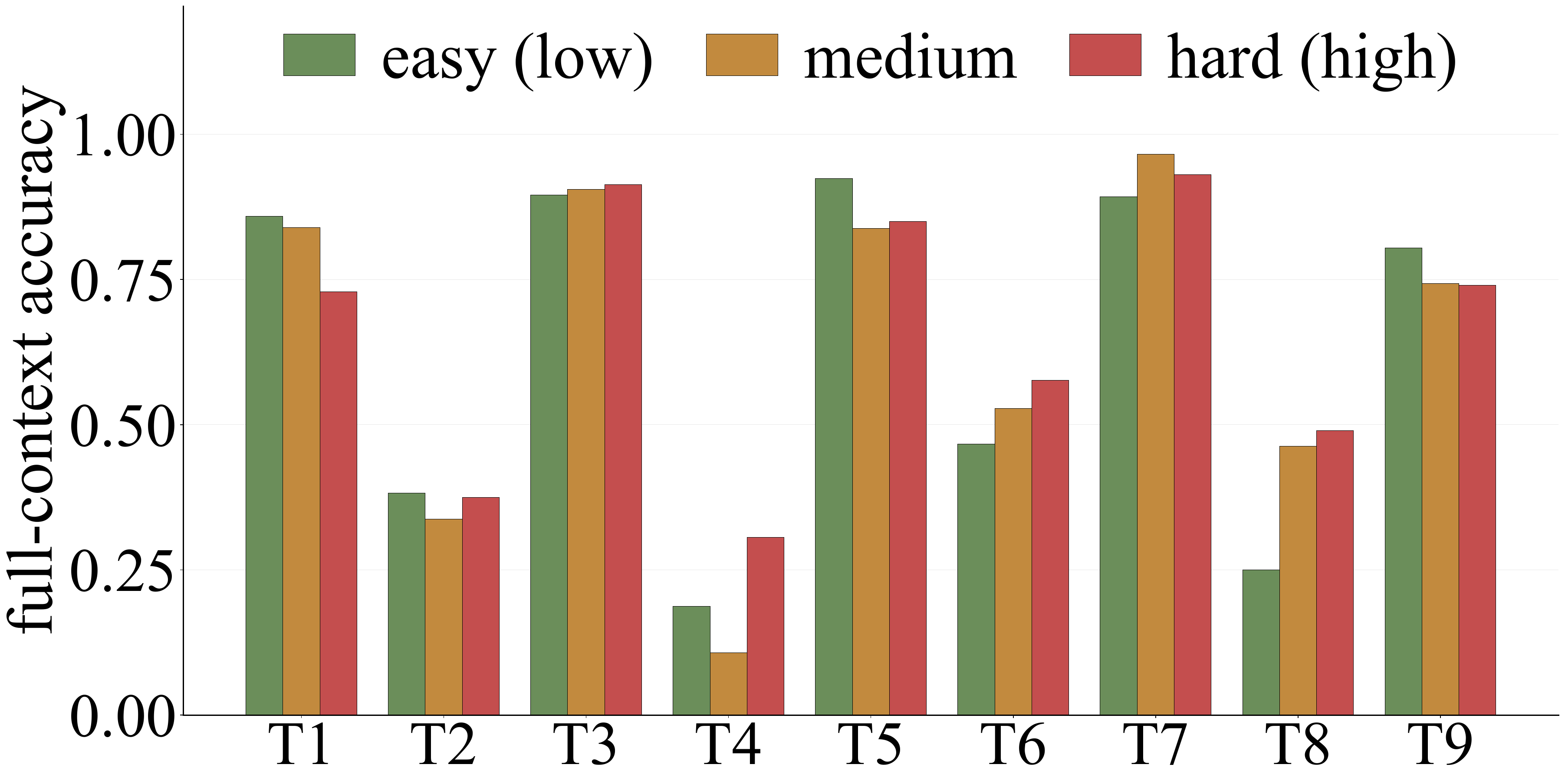}
\caption{D7: per-tertile \texttt{full-context} accuracy per task, pooled
across backbones. Patients are split into low / medium / high
complexity tertiles by the stratifier of \Cref{sec:design}; per-task
tertile sizes follow the per-task question counts reported there.}
\label{fig:appD7}
\end{figure}

Each question carries a difficulty tertile from the generation-time
stratifier. \Cref{fig:appD7} checks whether the tertile actually
predicts difficulty under \texttt{full-context}. T1 and T9 are
monotonically calibrated, while T8's ``high'' tertile is the easiest of
the three -- T8 tertile anti-alignment, which we flag in the main paper.

\subsection{Bootstrap CI widths per cell (D8)}
\begin{figure*}[t]
\centering
\includegraphics[width=\textwidth]{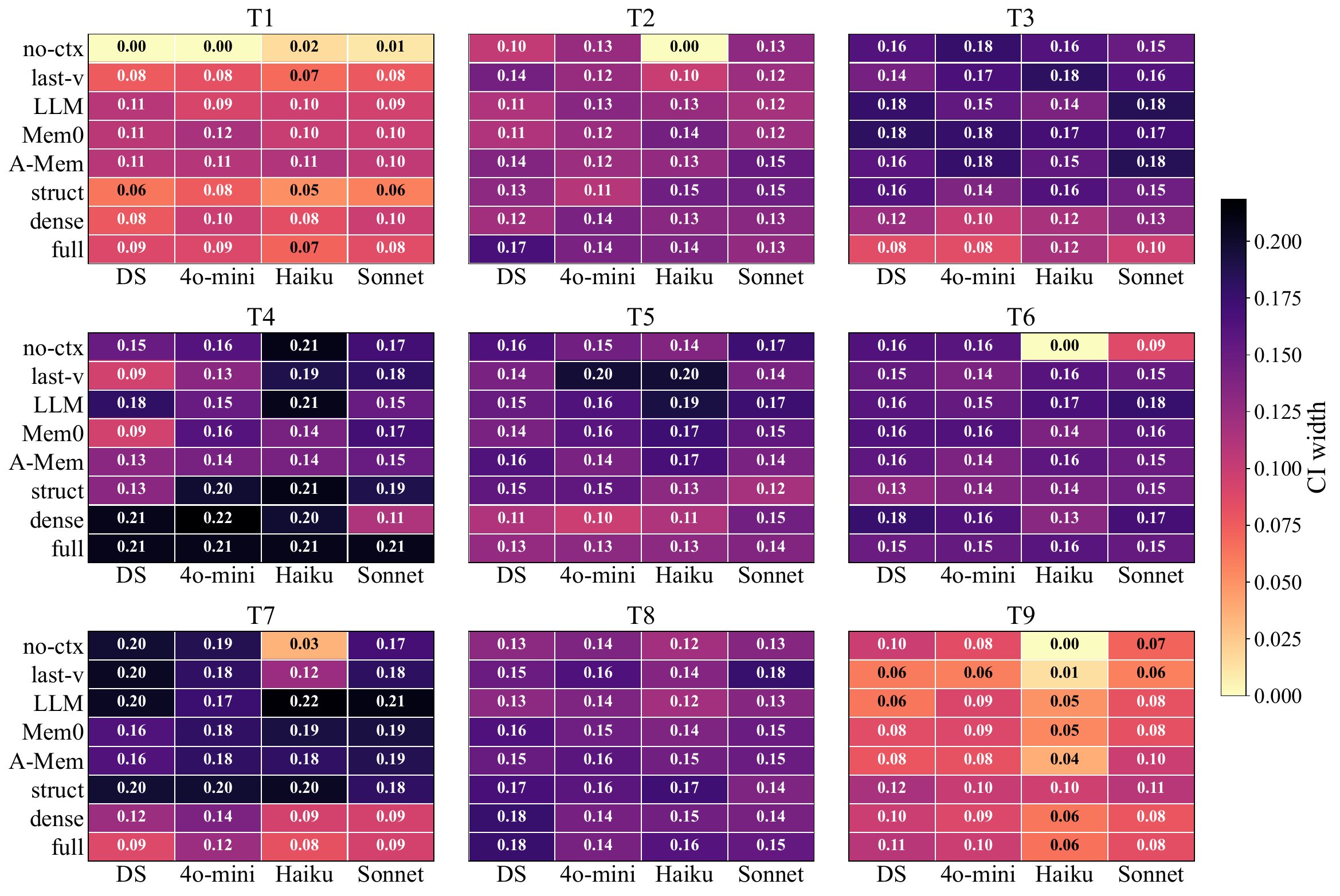}
\caption{D8: bootstrap 95\% CI width per (task, baseline, backbone) cell.
Shared \texttt{magma\_r} colour scale across all nine sub-panels.
Per-task questions/cell: T1 $n=1500$, T2 889, T3 600, T4 53,
T5 495, T6 700, T7 400, T8 134, T9 1500.}
\label{fig:appD8}
\end{figure*}

\Cref{fig:appD8} reports the bootstrap 95\% CI width for each of the
$9 \times 8 \times 4 = 288$ cells (100 resamples per cell, seed
20260519). The headline is that CI width is dominated by per-cell sample
size, not by memory strategy or backbone: T4 (smallest $n$) and T8 are
the widest by construction; T1 / T7 / T9 (the highest-resampled tasks)
are the tightest. Conclusion: the headline rankings in the main paper
are robust to within-cell sampling noise.

\section{Supplementary analyses: balanced metrics, sensitivity, and ablations}
\label{app:newresults}

This appendix collects the analyses referenced from \Cref{sec:design},
\Cref{sec:strategies}, \Cref{sec:results} and \Cref{sec:discussion}. No models
were rerun for the re-scoring analyses (\Cref{tab:appR-t5f1}--\Cref{tab:appR-t7parser},
\Cref{tab:appR-mde}, \Cref{tab:appR-subsets}); the ablations in
\Cref{tab:appR-prep}--\Cref{tab:appR-sumlen} were run on a stratified
20-patient subset of the 385-dialogue cohort.

\subsection{Class-balanced metrics for the skewed tasks}

Accuracy on T5 and T8 conflates two questions: whether a task beats a trivial
majority predictor, and whether it separates representation strategies.
Macro-F1 separates them: \Cref{tab:appR-t5f1} reports T5 and
\Cref{tab:appR-t8f1} reports T8.

\begin{table}[htbp]
\centering\small\setlength{\tabcolsep}{3.2pt}
\begin{tabular}{lccccc}
\toprule
strategy & DS & GPT & Hk & Sn & pooled \\
\midrule
\textit{full-context}       & 0.731 & 0.710 & 0.674 & 0.661 & 0.695 \\
\textit{dense-retrieval}    & 0.743 & 0.760 & 0.709 & 0.734 & 0.736 \\
\textit{structured-timeline}& 0.620 & 0.559 & 0.738 & 0.717 & 0.656 \\
\textit{Mem0}               & 0.524 & 0.482 & 0.466 & 0.491 & 0.491 \\
\textit{A-Mem}              & 0.580 & 0.491 & 0.593 & 0.599 & 0.565 \\
\textit{llm-summary}        & 0.578 & 0.428 & 0.468 & 0.446 & 0.476 \\
\textit{last-visit-only}    & 0.513 & 0.367 & 0.340 & 0.412 & 0.408 \\
\textit{no-context-blind}   & 0.247 & 0.348 & 0.155 & 0.425 & 0.308 \\
\bottomrule
\end{tabular}
\caption{T5 macro-F1 (majority-class baseline $=0.478$). \textit{full-context},
\textit{dense-retrieval}, and \textit{structured-timeline} exceed the baseline on
every backbone; the compressed strategies do not.}
\label{tab:appR-t5f1}
\end{table}

\begin{table}[htbp]
\centering\small\setlength{\tabcolsep}{3.2pt}
\begin{tabular}{lccccc}
\toprule
strategy & DS & GPT & Hk & Sn & pooled \\
\midrule
\textit{full-context}       & 0.504 & 0.377 & 0.498 & 0.547 & 0.485 \\
\textit{dense-retrieval}    & 0.478 & 0.378 & 0.476 & 0.479 & 0.455 \\
\textit{structured-timeline}& 0.513 & 0.465 & 0.538 & 0.569 & 0.523 \\
\textit{Mem0}               & 0.454 & 0.440 & 0.435 & 0.414 & 0.435 \\
\textit{A-Mem}              & 0.455 & 0.435 & 0.411 & 0.459 & 0.442 \\
\textit{llm-summary}        & 0.354 & 0.360 & 0.329 & 0.372 & 0.353 \\
\textit{last-visit-only}    & 0.399 & 0.454 & 0.328 & 0.359 & 0.389 \\
\textit{no-context-blind}   & 0.249 & 0.275 & 0.328 & 0.373 & 0.294 \\
\bottomrule
\end{tabular}
\caption{T8 macro-F1 (majority-class baseline $=0.247$; $n=134$ per cell).
\textit{structured-timeline} leads, followed by \textit{full-context} and
\textit{dense-retrieval}.}
\label{tab:appR-t8f1}
\end{table}

T6 is already discriminative under accuracy: its best cell reaches 0.609 against
a 0.423 majority baseline (\Cref{tab:main}). T5, T6 and T8 therefore do separate
representations; T9 remains a relative diagnostic probe.

\subsection{Statistical power per task}

\begin{table}[htbp]
\centering\small
\begin{tabular}{lccl}
\toprule
task & $n$ & MDE & assessment \\
\midrule
T1 / T9 & 1500 & \phantom{0}5.1 pp & well-powered \\
T2      & \phantom{0}889 & \phantom{0}6.6 pp & well-powered \\
T6      & \phantom{0}700 & \phantom{0}7.5 pp & well-powered \\
T3      & \phantom{0}600 & \phantom{0}8.1 pp & well-powered \\
T5      & \phantom{0}495 & \phantom{0}8.9 pp & well-powered \\
T7      & \phantom{0}400 & \phantom{0}9.9 pp & well-powered \\
T8      & \phantom{0}134 & 17.1 pp & weak \\
T4      & \phantom{00}53 & 27.2 pp & under-powered \\
\bottomrule
\end{tabular}
\caption{Conservative per-task minimum detectable effects (two-proportion
approximation, $\alpha=0.05$, 80\% power, $p=0.5$). T8 has limited power for
moderate differences; T4 is clearly under-powered.}
\label{tab:appR-mde}
\end{table}

\subsection{Parser-strictness sensitivity}

T9 abstentions can be phrased in many ways, so we re-scored the existing T9
predictions with a lenient parser that also accepts synonymous or verbose
abstentions (e.g.\ ``no sodium value was reported on that date'', ``not
measured''); \Cref{tab:appR-t9parser} reports the result. For T7 we used a
lenient matcher accepting either the option label or the full option text
(\Cref{tab:appR-t7parser}).

\begin{table}[htbp]
\centering\small
\begin{tabular}{lccc}
\toprule
strategy & strict & lenient & $\Delta$ \\
\midrule
\textit{full-context}       & 0.763 & 0.838 & $+0.075$ \\
\textit{dense-retrieval}    & 0.798 & 0.832 & $+0.034$ \\
\textit{structured-timeline}& 0.535 & 0.791 & $+0.256$ \\
\textit{Mem0}               & 0.823 & 0.842 & $+0.019$ \\
\textit{A-Mem}              & 0.825 & 0.841 & $+0.016$ \\
\textit{llm-summary}        & 0.814 & 0.880 & $+0.066$ \\
\textit{last-visit-only}    & 0.924 & 0.930 & $+0.005$ \\
\textit{no-context-blind}   & 0.311 & 0.353 & $+0.042$ \\
\bottomrule
\end{tabular}
\caption{T9 abstention accuracy under strict and lenient parsers, full T9
evaluation set (1{,}500 questions per backbone, pooled over backbones).
\textit{last-visit-only} stays above \textit{full-context} (0.930 vs.\ 0.838);
the gap narrows from 0.161 to 0.092, so parser strictness changes the magnitude
but not the direction of SP3.}
\label{tab:appR-t9parser}
\end{table}

\begin{table}[htbp]
\centering\small
\begin{tabular}{lccc}
\toprule
strategy & strict & lenient & $\Delta$ \\
\midrule
\textit{full-context}       & 0.933 & 0.933 & $0.000$ \\
\textit{dense-retrieval}    & 0.907 & 0.907 & $0.000$ \\
\textit{structured-timeline}& 0.613 & 0.613 & $0.000$ \\
\textit{Mem0}               & 0.642 & 0.642 & $0.000$ \\
\textit{A-Mem}              & 0.581 & 0.581 & $0.000$ \\
\textit{llm-summary}        & 0.448 & 0.448 & $0.000$ \\
\textit{last-visit-only}    & 0.294 & 0.294 & $0.000$ \\
\textit{no-context-blind}   & 0.177 & 0.177 & $0.000$ \\
\bottomrule
\end{tabular}
\caption{T7 accuracy (4-option MCQ) under strict and lenient matching: identical
throughout. Models answer as ``option label $+$ option text'', so the strict
parser already captures valid answers and incorrect responses are genuine errors
rather than parser misses.}
\label{tab:appR-t7parser}
\end{table}

\subsection{T9 over-answer composition}

On the T9 diagnostic subset the strict parser scored 307 \textit{full-context}
responses as non-abstentions. Of these, 36\% fabricated a specific value, 28\%
gave a bare Yes/No, and 36\% fell into an ``other'' category---but 93\% of that
``other'' bucket were verbose abstentions the strict parser missed. Discounting
those parser misses, fabricated values account for roughly 54\% of genuine
over-answers, making plausible-but-unsupported detail the largest observed error
type behind SP3.

\subsection{Backbone-matched memory preparation}

We rebuilt \textit{Mem0} and \textit{A-Mem} with each non-DeepSeek answering
backbone, so that GPT-4o-mini, Haiku and Sonnet each prepared and answered from
their own memory. The preparation model is the only changed factor.

\begin{table}[htbp]
\centering\small\setlength{\tabcolsep}{3.5pt}
\begin{tabular}{lccc}
\toprule
cell & DS-prepared & matched & $\Delta$ \\
\midrule
GPT-4o-mini $\times$ \textit{Mem0}  & 0.532 & 0.340 & $-0.192$ \\
Haiku $\times$ \textit{Mem0}        & 0.571 & 0.327 & $-0.244$ \\
Sonnet $\times$ \textit{Mem0}       & 0.526 & 0.372 & $-0.154$ \\
GPT-4o-mini $\times$ \textit{A-Mem} & 0.474 & 0.346 & $-0.128$ \\
Haiku $\times$ \textit{A-Mem}       & 0.513 & 0.365 & $-0.147$ \\
Sonnet $\times$ \textit{A-Mem}      & 0.481 & 0.436 & $-0.045$ \\
\bottomrule
\end{tabular}
\caption{Overall accuracy with DeepSeek-V3--prepared versus backbone-matched
agentic memory on the stratified 20-patient subset. Matching prep to the
answering backbone does not improve accuracy in any of the six cells. The
corresponding T3 positive-class recall is reported in \Cref{tab:sp4-equal}.}
\label{tab:appR-prep}
\end{table}

\subsection{Dense-retrieval and summary-length ablations}

\Cref{tab:appR-topk} varies the retrieval depth and \Cref{tab:appR-sumlen}
the summary budget; neither knob removes the gap to \textit{full-context}.

\begin{table}[htbp]
\centering\small
\begin{tabular}{lccc}
\toprule
backbone & $K=3$ & $K=5$ & $K=10$ \\
\midrule
DeepSeek-V3 & 0.474 & 0.513 & 0.487 \\
GPT-4o-mini & 0.410 & 0.397 & 0.391 \\
Haiku~4.5   & 0.423 & 0.487 & 0.385 \\
Sonnet~4.6  & 0.468 & 0.506 & 0.558 \\
\bottomrule
\end{tabular}
\caption{\textit{dense-retrieval} accuracy across top-$K$ on the stratified
20-patient subset, all nine tasks. Larger $K$ gives no consistent gain, matching
the retrieval diagnostics in \Cref{app:appA10}: hit rate rises from 0.852 at
$K{=}3$ to 0.979 at $K{=}10$ without a corresponding accuracy gain. $K{=}5$ is
therefore not the sole source of the remaining retrieval errors.}
\label{tab:appR-topk}
\end{table}

\begin{table}[htbp]
\centering\small\setlength{\tabcolsep}{3.2pt}
\begin{tabular}{lcccc}
\toprule
& \multicolumn{2}{c}{all 9 tasks} & \multicolumn{2}{c}{aggregation} \\
\cmidrule(lr){2-3}\cmidrule(lr){4-5}
backbone & 500 & 1000 & 500 & 1000 \\
\midrule
DeepSeek-V3 & 0.465 & 0.393 & 0.211 & 0.259 \\
GPT-4o-mini & 0.392 & 0.309 & 0.235 & 0.185 \\
Haiku~4.5   & 0.405 & 0.372 & 0.235 & 0.210 \\
Sonnet~4.6  & 0.399 & 0.415 & 0.258 & 0.304 \\
\bottomrule
\end{tabular}
\caption{\textit{llm-summary} accuracy at a 500- versus 1{,}000-token budget on
the stratified 20-patient subset; ``aggregation'' pools T2, T6 and T8. Doubling
the budget lowers overall accuracy for three of four backbones and moves
aggregation accuracy in both directions within 0.185--0.304. The 500-token limit
is not the sole explanation for SP1.}
\label{tab:appR-sumlen}
\end{table}

\subsection{Macro-accuracy under task-subset ablation}

\begin{table}[htbp]
\centering\small\setlength{\tabcolsep}{3.2pt}
\begin{tabular}{lccc}
\toprule
strategy & all 9 & w/o T5,T8,T9 & w/o T3 \\
\midrule
\textit{full-context}        & 0.645 & 0.628 & 0.612 \\
\textit{dense-retrieval}     & 0.612 & 0.580 & 0.585 \\
\textit{structured-timeline} & 0.541 & 0.508 & 0.545 \\
\textit{Mem0}                & 0.493 & 0.421 & 0.493 \\
\textit{A-Mem}               & 0.485 & 0.408 & 0.484 \\
\textit{llm-summary}         & 0.435 & 0.365 & 0.428 \\
\textit{last-visit-only}     & 0.403 & 0.306 & 0.385 \\
\textit{no-context-blind}    & 0.226 & 0.200 & 0.193 \\
\bottomrule
\end{tabular}
\caption{Macro-accuracy averaged over the included tasks and the four backbones,
recomputed from the existing predictions. The strategy ordering is unchanged
when the exploratory tasks (T5, T8, T9) or the T3 probe are excluded, so the
headline comparison does not depend on them.}
\label{tab:appR-subsets}
\end{table}

\subsection{Substrate pilot: dialogue versus a bare event table}

As a check on what the dialogue layer contributes, we ran a small
\textit{full-context} pilot in which the model received either the dialogue or a
bare structured-event table built from the same source events. The table
retained the events, but the narrative relations targeted by T3, T4 and T5 were
not explicitly represented in it, so those tasks could not be evaluated
equivalently from the table alone. This delimits what the dialogue substrate
adds; it is not a validation of real-note realism, and a controlled comparison
against a visit-organized table is left to future work.

\onecolumn
\section{Case studies}
\label{app:cases}

The main paper carries four illustrative failure-mode case studies; we
reprint them here so the appendix is self-contained.

\label{app:appA12cases}
\begin{figure*}[htbp]
\centering
\setlength{\tabcolsep}{3pt}
\setlength{\fboxsep}{3pt}
\renewcommand{\arraystretch}{0.92}
\fontsize{8.4}{9.2}\selectfont
\begin{tabularx}{\textwidth}{@{}X X@{}}
\fbox{\begin{minipage}[t][0.11\textheight][t]{0.47\textwidth}
\textbf{Case 1 -- T3: Controlled Evidence--Problem Linkage}\\[0.55em]
\textbf{Q:} Did the physician explicitly attribute the Creatinine 1.7 mg/dL (2183-01-13) to a diagnosis of\ldots{}\\[0.55em]
\textcolor{green!50!black}{(full-context, DeepSeek):} \hfill \textcolor{green!50!black}{yes \quad [correct]}\\[0.25em]
\textcolor{red!70!black}{(mem0, DeepSeek):} \hfill \textcolor{red!70!black}{no \quad [incorrect]}\\[0.55em]
\textit{Lesson: Compressed representations lose the encounter-local linkage sentence and collapse to constant-No.}
\end{minipage}}
&
\fbox{\begin{minipage}[t][0.11\textheight][t]{0.47\textwidth}
\textbf{Case 2 -- T9: Abstention over Unstated Facts}\\[0.55em]
\textbf{Q:} Based on the dialogue, what was the patient's Potassium on 2138-04-03?\\[0.55em]
\textcolor{green!50!black}{(last-visit-only, Sonnet):} \hfill \textcolor{green!50!black}{insufficient \quad [correct]}\\[0.25em]
\textcolor{red!70!black}{(full-context, Sonnet):} \hfill \textcolor{red!70!black}{the patient's Pota\ldots{} \quad [incorrect]}\\[0.55em]
\textit{Lesson: More context can hurt abstention; full-context over-answers.}
\end{minipage}}
\\[0.35em]
\fbox{\begin{minipage}[t][0.11\textheight][t]{0.47\textwidth}
\textbf{Case 3 -- T2: Multi-Visit Trend Classification}\\[0.55em]
\textbf{Q:} How did the patient's Potassium change overall from 2126-02-06 to 2126-12-06? Answer \texttt{<direction>;<magnitude>}.\\[0.55em]
\textcolor{green!50!black}{(full-context, Sonnet):} \hfill \textcolor{green!50!black}{decreased;weak \quad [correct]}\\[0.25em]
\textcolor{red!70!black}{(full-context, Haiku):} \hfill \textcolor{red!70!black}{stable;none \quad [incorrect]}\\[0.55em]
\textit{Lesson: Even with full chart access, multi-visit trend aggregation remains backbone-sensitive.}
\end{minipage}}
&
\fbox{\begin{minipage}[t][0.11\textheight][t]{0.47\textwidth}
\textbf{Case 4 -- T6: Cross-Patient Comparison}\\[0.55em]
\textbf{Q:} Comparing patient A and patient B, which patient has more diagnoses documented in the dialogue? Answer A, B, or similar.\\[0.55em]
\textcolor{green!50!black}{(full-context, Sonnet):} \hfill \textcolor{green!50!black}{B \quad [correct]}\\[0.25em]
\textcolor{red!70!black}{(mem0, Sonnet):} \hfill \textcolor{red!70!black}{A \quad [incorrect]}\\[0.55em]
\textit{Lesson: Agentic memory representations can confuse facts across patients when retrieval is keyed on similar attributes.}
\end{minipage}}
\end{tabularx}
\caption{Four illustrative failure modes drawn from the master results table: 
representational loss (Case~1, T3), uncritical answering (Case~2, T9), 
reasoning-capacity ceiling (Case~3, T2 on Haiku), 
cross-document aggregation breakdown (Case~4, T6). 
Green = correct prediction; red = incorrect.}
\label{fig:9}
\end{figure*}

\section{Full per-task strategy-by-backbone matrix}
\label{app:fullmatrix}
\label{app:appA1}  

For completeness, the full $9 \times 8 \times 4 = 288$-cell accuracy
matrix from which all per-task figures in the main paper and in this
appendix are derived is reproduced in \Cref{tab:appA1_full_matrix}.

{\footnotesize
\setlength{\tabcolsep}{5pt}%
\renewcommand{\arraystretch}{0.92}%
\begin{longtable}{l l l r r c}
    \caption{Complete per-task accuracy across all 32 (history representation strategy, backbone) cells, on the 6{,}271-question stratified evaluation set (200{,}672 evaluations total). \texttt{n} is the per-task sample size; per-task sizes are T1=1500, T2=889, T3=600, T4=53, T5=495, T6=700, T7=400, T8=134, T9=1500. The rightmost column marks rows from the stratified evaluation set (every row is).} \label{tab:appA1_full_matrix} \\
    \toprule
    Task & Strategy & Backbone & $n$ & Accuracy & Subset \\
    \midrule
    \endfirsthead
    \multicolumn{6}{l}{\textit{(continued from previous page)}} \\
    \toprule
    Task & Strategy & Backbone & $n$ & Accuracy & Subset \\
    \midrule
    \endhead
    \midrule
    \multicolumn{6}{r}{\textit{(continued on next page)}} \\
    \endfoot
    \bottomrule
    \endlastfoot
    T1 & no-context (blind) & DeepSeek & 1500 & 0.000 & \checkmark \\
    T1 & no-context (blind) & GPT-4o-mini & 1500 & 0.000 & \checkmark \\
    T1 & no-context (blind) & Claude Haiku & 1500 & 0.006 & \checkmark \\
    T1 & no-context (blind) & Claude Sonnet & 1500 & 0.003 & \checkmark \\
    T1 & last-visit only & DeepSeek & 1500 & 0.160 & \checkmark \\
    T1 & last-visit only & GPT-4o-mini & 1500 & 0.148 & \checkmark \\
    T1 & last-visit only & Claude Haiku & 1500 & 0.127 & \checkmark \\
    T1 & last-visit only & Claude Sonnet & 1500 & 0.120 & \checkmark \\
    T1 & LLM summary & DeepSeek & 1500 & 0.355 & \checkmark \\
    T1 & LLM summary & GPT-4o-mini & 1500 & 0.361 & \checkmark \\
    T1 & LLM summary & Claude Haiku & 1500 & 0.352 & \checkmark \\
    T1 & LLM summary & Claude Sonnet & 1500 & 0.355 & \checkmark \\
    T1 & Mem0 & DeepSeek & 1500 & 0.515 & \checkmark \\
    T1 & Mem0 & GPT-4o-mini & 1500 & 0.515 & \checkmark \\
    T1 & Mem0 & Claude Haiku & 1500 & 0.509 & \checkmark \\
    T1 & Mem0 & Claude Sonnet & 1500 & 0.488 & \checkmark \\
    T1 & A-Mem & DeepSeek & 1500 & 0.491 & \checkmark \\
    T1 & A-Mem & GPT-4o-mini & 1500 & 0.420 & \checkmark \\
    T1 & A-Mem & Claude Haiku & 1500 & 0.426 & \checkmark \\
    T1 & A-Mem & Claude Sonnet & 1500 & 0.485 & \checkmark \\
    T1 & structured timeline & DeepSeek & 1500 & 0.886 & \checkmark \\
    T1 & structured timeline & GPT-4o-mini & 1500 & 0.886 & \checkmark \\
    T1 & structured timeline & Claude Haiku & 1500 & 0.920 & \checkmark \\
    T1 & structured timeline & Claude Sonnet & 1500 & 0.923 & \checkmark \\
    T1 & dense retrieval & DeepSeek & 1500 & 0.802 & \checkmark \\
    T1 & dense retrieval & GPT-4o-mini & 1500 & 0.738 & \checkmark \\
    T1 & dense retrieval & Claude Haiku & 1500 & 0.778 & \checkmark \\
    T1 & dense retrieval & Claude Sonnet & 1500 & 0.787 & \checkmark \\
    T1 & full context & DeepSeek & 1500 & 0.806 & \checkmark \\
    T1 & full context & GPT-4o-mini & 1500 & 0.781 & \checkmark \\
    T1 & full context & Claude Haiku & 1500 & 0.812 & \checkmark \\
    T1 & full context & Claude Sonnet & 1500 & 0.843 & \checkmark \\
    \midrule
    T2 & no-context (blind) & DeepSeek & 889 & 0.172 & \checkmark \\
    T2 & no-context (blind) & GPT-4o-mini & 889 & 0.333 & \checkmark \\
    T2 & no-context (blind) & Claude Haiku & 889 & 0.000 & \checkmark \\
    T2 & no-context (blind) & Claude Sonnet & 889 & 0.302 & \checkmark \\
    T2 & last-visit only & DeepSeek & 889 & 0.328 & \checkmark \\
    T2 & last-visit only & GPT-4o-mini & 889 & 0.354 & \checkmark \\
    T2 & last-visit only & Claude Haiku & 889 & 0.177 & \checkmark \\
    T2 & last-visit only & Claude Sonnet & 889 & 0.354 & \checkmark \\
    T2 & LLM summary & DeepSeek & 889 & 0.260 & \checkmark \\
    T2 & LLM summary & GPT-4o-mini & 889 & 0.333 & \checkmark \\
    T2 & LLM summary & Claude Haiku & 889 & 0.323 & \checkmark \\
    T2 & LLM summary & Claude Sonnet & 889 & 0.339 & \checkmark \\
    T2 & Mem0 & DeepSeek & 889 & 0.333 & \checkmark \\
    T2 & Mem0 & GPT-4o-mini & 889 & 0.297 & \checkmark \\
    T2 & Mem0 & Claude Haiku & 889 & 0.349 & \checkmark \\
    T2 & Mem0 & Claude Sonnet & 889 & 0.391 & \checkmark \\
    T2 & A-Mem & DeepSeek & 889 & 0.333 & \checkmark \\
    T2 & A-Mem & GPT-4o-mini & 889 & 0.323 & \checkmark \\
    T2 & A-Mem & Claude Haiku & 889 & 0.365 & \checkmark \\
    T2 & A-Mem & Claude Sonnet & 889 & 0.380 & \checkmark \\
    T2 & structured timeline & DeepSeek & 889 & 0.297 & \checkmark \\
    T2 & structured timeline & GPT-4o-mini & 889 & 0.214 & \checkmark \\
    T2 & structured timeline & Claude Haiku & 889 & 0.292 & \checkmark \\
    T2 & structured timeline & Claude Sonnet & 889 & 0.406 & \checkmark \\
    T2 & dense retrieval & DeepSeek & 889 & 0.354 & \checkmark \\
    T2 & dense retrieval & GPT-4o-mini & 889 & 0.266 & \checkmark \\
    T2 & dense retrieval & Claude Haiku & 889 & 0.427 & \checkmark \\
    T2 & dense retrieval & Claude Sonnet & 889 & 0.432 & \checkmark \\
    T2 & full context & DeepSeek & 889 & 0.375 & \checkmark \\
    T2 & full context & GPT-4o-mini & 889 & 0.255 & \checkmark \\
    T2 & full context & Claude Haiku & 889 & 0.349 & \checkmark \\
    T2 & full context & Claude Sonnet & 889 & 0.474 & \checkmark \\
    \midrule
    T3 & no-context (blind) & DeepSeek & 600 & 0.500 & \checkmark \\
    T3 & no-context (blind) & GPT-4o-mini & 600 & 0.492 & \checkmark \\
    T3 & no-context (blind) & Claude Haiku & 600 & 0.500 & \checkmark \\
    T3 & no-context (blind) & Claude Sonnet & 600 & 0.500 & \checkmark \\
    T3 & last-visit only & DeepSeek & 600 & 0.562 & \checkmark \\
    T3 & last-visit only & GPT-4o-mini & 600 & 0.538 & \checkmark \\
    T3 & last-visit only & Claude Haiku & 600 & 0.554 & \checkmark \\
    T3 & last-visit only & Claude Sonnet & 600 & 0.508 & \checkmark \\
    T3 & LLM summary & DeepSeek & 600 & 0.500 & \checkmark \\
    T3 & LLM summary & GPT-4o-mini & 600 & 0.492 & \checkmark \\
    T3 & LLM summary & Claude Haiku & 600 & 0.500 & \checkmark \\
    T3 & LLM summary & Claude Sonnet & 600 & 0.500 & \checkmark \\
    T3 & Mem0 & DeepSeek & 600 & 0.492 & \checkmark \\
    T3 & Mem0 & GPT-4o-mini & 600 & 0.477 & \checkmark \\
    T3 & Mem0 & Claude Haiku & 600 & 0.500 & \checkmark \\
    T3 & Mem0 & Claude Sonnet & 600 & 0.500 & \checkmark \\
    T3 & A-Mem & DeepSeek & 600 & 0.500 & \checkmark \\
    T3 & A-Mem & GPT-4o-mini & 600 & 0.477 & \checkmark \\
    T3 & A-Mem & Claude Haiku & 600 & 0.477 & \checkmark \\
    T3 & A-Mem & Claude Sonnet & 600 & 0.500 & \checkmark \\
    T3 & structured timeline & DeepSeek & 600 & 0.508 & \checkmark \\
    T3 & structured timeline & GPT-4o-mini & 600 & 0.508 & \checkmark \\
    T3 & structured timeline & Claude Haiku & 600 & 0.500 & \checkmark \\
    T3 & structured timeline & Claude Sonnet & 600 & 0.500 & \checkmark \\
    T3 & dense retrieval & DeepSeek & 600 & 0.831 & \checkmark \\
    T3 & dense retrieval & GPT-4o-mini & 600 & 0.846 & \checkmark \\
    T3 & dense retrieval & Claude Haiku & 600 & 0.869 & \checkmark \\
    T3 & dense retrieval & Claude Sonnet & 600 & 0.769 & \checkmark \\
    T3 & full context & DeepSeek & 600 & 0.938 & \checkmark \\
    T3 & full context & GPT-4o-mini & 600 & 0.931 & \checkmark \\
    T3 & full context & Claude Haiku & 600 & 0.869 & \checkmark \\
    T3 & full context & Claude Sonnet & 600 & 0.885 & \checkmark \\
    \midrule
    T4 & no-context (blind) & DeepSeek & 53 & 0.132 & \checkmark \\
    T4 & no-context (blind) & GPT-4o-mini & 53 & 0.075 & \checkmark \\
    T4 & no-context (blind) & Claude Haiku & 53 & 0.189 & \checkmark \\
    T4 & no-context (blind) & Claude Sonnet & 53 & 0.113 & \checkmark \\
    T4 & last-visit only & DeepSeek & 53 & 0.057 & \checkmark \\
    T4 & last-visit only & GPT-4o-mini & 53 & 0.113 & \checkmark \\
    T4 & last-visit only & Claude Haiku & 53 & 0.170 & \checkmark \\
    T4 & last-visit only & Claude Sonnet & 53 & 0.113 & \checkmark \\
    T4 & LLM summary & DeepSeek & 53 & 0.151 & \checkmark \\
    T4 & LLM summary & GPT-4o-mini & 53 & 0.170 & \checkmark \\
    T4 & LLM summary & Claude Haiku & 53 & 0.189 & \checkmark \\
    T4 & LLM summary & Claude Sonnet & 53 & 0.113 & \checkmark \\
    T4 & Mem0 & DeepSeek & 53 & 0.038 & \checkmark \\
    T4 & Mem0 & GPT-4o-mini & 53 & 0.132 & \checkmark \\
    T4 & Mem0 & Claude Haiku & 53 & 0.094 & \checkmark \\
    T4 & Mem0 & Claude Sonnet & 53 & 0.151 & \checkmark \\
    T4 & A-Mem & DeepSeek & 53 & 0.075 & \checkmark \\
    T4 & A-Mem & GPT-4o-mini & 53 & 0.113 & \checkmark \\
    T4 & A-Mem & Claude Haiku & 53 & 0.094 & \checkmark \\
    T4 & A-Mem & Claude Sonnet & 53 & 0.132 & \checkmark \\
    T4 & structured timeline & DeepSeek & 53 & 0.057 & \checkmark \\
    T4 & structured timeline & GPT-4o-mini & 53 & 0.208 & \checkmark \\
    T4 & structured timeline & Claude Haiku & 53 & 0.245 & \checkmark \\
    T4 & structured timeline & Claude Sonnet & 53 & 0.170 & \checkmark \\
    T4 & dense retrieval & DeepSeek & 53 & 0.208 & \checkmark \\
    T4 & dense retrieval & GPT-4o-mini & 53 & 0.226 & \checkmark \\
    T4 & dense retrieval & Claude Haiku & 53 & 0.151 & \checkmark \\
    T4 & dense retrieval & Claude Sonnet & 53 & 0.057 & \checkmark \\
    T4 & full context & DeepSeek & 53 & 0.226 & \checkmark \\
    T4 & full context & GPT-4o-mini & 53 & 0.283 & \checkmark \\
    T4 & full context & Claude Haiku & 53 & 0.189 & \checkmark \\
    T4 & full context & Claude Sonnet & 53 & 0.245 & \checkmark \\
    \midrule
    T5 & no-context (blind) & DeepSeek & 495 & 0.252 & \checkmark \\
    T5 & no-context (blind) & GPT-4o-mini & 495 & 0.393 & \checkmark \\
    T5 & no-context (blind) & Claude Haiku & 495 & 0.168 & \checkmark \\
    T5 & no-context (blind) & Claude Sonnet & 495 & 0.383 & \checkmark \\
    T5 & last-visit only & DeepSeek & 495 & 0.785 & \checkmark \\
    T5 & last-visit only & GPT-4o-mini & 495 & 0.579 & \checkmark \\
    T5 & last-visit only & Claude Haiku & 495 & 0.374 & \checkmark \\
    T5 & last-visit only & Claude Sonnet & 495 & 0.636 & \checkmark \\
    T5 & LLM summary & DeepSeek & 495 & 0.832 & \checkmark \\
    T5 & LLM summary & GPT-4o-mini & 495 & 0.673 & \checkmark \\
    T5 & LLM summary & Claude Haiku & 495 & 0.636 & \checkmark \\
    T5 & LLM summary & Claude Sonnet & 495 & 0.626 & \checkmark \\
    T5 & Mem0 & DeepSeek & 495 & 0.804 & \checkmark \\
    T5 & Mem0 & GPT-4o-mini & 495 & 0.729 & \checkmark \\
    T5 & Mem0 & Claude Haiku & 495 & 0.720 & \checkmark \\
    T5 & Mem0 & Claude Sonnet & 495 & 0.748 & \checkmark \\
    T5 & A-Mem & DeepSeek & 495 & 0.776 & \checkmark \\
    T5 & A-Mem & GPT-4o-mini & 495 & 0.692 & \checkmark \\
    T5 & A-Mem & Claude Haiku & 495 & 0.738 & \checkmark \\
    T5 & A-Mem & Claude Sonnet & 495 & 0.776 & \checkmark \\
    T5 & structured timeline & DeepSeek & 495 & 0.748 & \checkmark \\
    T5 & structured timeline & GPT-4o-mini & 495 & 0.776 & \checkmark \\
    T5 & structured timeline & Claude Haiku & 495 & 0.869 & \checkmark \\
    T5 & structured timeline & Claude Sonnet & 495 & 0.860 & \checkmark \\
    T5 & dense retrieval & DeepSeek & 495 & 0.869 & \checkmark \\
    T5 & dense retrieval & GPT-4o-mini & 495 & 0.888 & \checkmark \\
    T5 & dense retrieval & Claude Haiku & 495 & 0.860 & \checkmark \\
    T5 & dense retrieval & Claude Sonnet & 495 & 0.841 & \checkmark \\
    T5 & full context & DeepSeek & 495 & 0.860 & \checkmark \\
    T5 & full context & GPT-4o-mini & 495 & 0.897 & \checkmark \\
    T5 & full context & Claude Haiku & 495 & 0.850 & \checkmark \\
    T5 & full context & Claude Sonnet & 495 & 0.841 & \checkmark \\
    \midrule
    T6 & no-context (blind) & DeepSeek & 700 & 0.358 & \checkmark \\
    T6 & no-context (blind) & GPT-4o-mini & 700 & 0.311 & \checkmark \\
    T6 & no-context (blind) & Claude Haiku & 700 & 0.000 & \checkmark \\
    T6 & no-context (blind) & Claude Sonnet & 700 & 0.099 & \checkmark \\
    T6 & last-visit only & DeepSeek & 700 & 0.424 & \checkmark \\
    T6 & last-visit only & GPT-4o-mini & 700 & 0.470 & \checkmark \\
    T6 & last-visit only & Claude Haiku & 700 & 0.470 & \checkmark \\
    T6 & last-visit only & Claude Sonnet & 700 & 0.424 & \checkmark \\
    T6 & LLM summary & DeepSeek & 700 & 0.444 & \checkmark \\
    T6 & LLM summary & GPT-4o-mini & 700 & 0.391 & \checkmark \\
    T6 & LLM summary & Claude Haiku & 700 & 0.417 & \checkmark \\
    T6 & LLM summary & Claude Sonnet & 700 & 0.424 & \checkmark \\
    T6 & Mem0 & DeepSeek & 700 & 0.457 & \checkmark \\
    T6 & Mem0 & GPT-4o-mini & 700 & 0.424 & \checkmark \\
    T6 & Mem0 & Claude Haiku & 700 & 0.450 & \checkmark \\
    T6 & Mem0 & Claude Sonnet & 700 & 0.430 & \checkmark \\
    T6 & A-Mem & DeepSeek & 700 & 0.503 & \checkmark \\
    T6 & A-Mem & GPT-4o-mini & 700 & 0.464 & \checkmark \\
    T6 & A-Mem & Claude Haiku & 700 & 0.444 & \checkmark \\
    T6 & A-Mem & Claude Sonnet & 700 & 0.477 & \checkmark \\
    T6 & structured timeline & DeepSeek & 700 & 0.609 & \checkmark \\
    T6 & structured timeline & GPT-4o-mini & 700 & 0.530 & \checkmark \\
    T6 & structured timeline & Claude Haiku & 700 & 0.523 & \checkmark \\
    T6 & structured timeline & Claude Sonnet & 700 & 0.563 & \checkmark \\
    T6 & dense retrieval & DeepSeek & 700 & 0.470 & \checkmark \\
    T6 & dense retrieval & GPT-4o-mini & 700 & 0.391 & \checkmark \\
    T6 & dense retrieval & Claude Haiku & 700 & 0.437 & \checkmark \\
    T6 & dense retrieval & Claude Sonnet & 700 & 0.457 & \checkmark \\
    T6 & full context & DeepSeek & 700 & 0.543 & \checkmark \\
    T6 & full context & GPT-4o-mini & 700 & 0.497 & \checkmark \\
    T6 & full context & Claude Haiku & 700 & 0.510 & \checkmark \\
    T6 & full context & Claude Sonnet & 700 & 0.523 & \checkmark \\
    \midrule
    T7 & no-context (blind) & DeepSeek & 400 & 0.256 & \checkmark \\
    T7 & no-context (blind) & GPT-4o-mini & 400 & 0.244 & \checkmark \\
    T7 & no-context (blind) & Claude Haiku & 400 & 0.012 & \checkmark \\
    T7 & no-context (blind) & Claude Sonnet & 400 & 0.198 & \checkmark \\
    T7 & last-visit only & DeepSeek & 400 & 0.360 & \checkmark \\
    T7 & last-visit only & GPT-4o-mini & 400 & 0.302 & \checkmark \\
    T7 & last-visit only & Claude Haiku & 400 & 0.140 & \checkmark \\
    T7 & last-visit only & Claude Sonnet & 400 & 0.372 & \checkmark \\
    T7 & LLM summary & DeepSeek & 400 & 0.512 & \checkmark \\
    T7 & LLM summary & GPT-4o-mini & 400 & 0.477 & \checkmark \\
    T7 & LLM summary & Claude Haiku & 400 & 0.349 & \checkmark \\
    T7 & LLM summary & Claude Sonnet & 400 & 0.453 & \checkmark \\
    T7 & Mem0 & DeepSeek & 400 & 0.733 & \checkmark \\
    T7 & Mem0 & GPT-4o-mini & 400 & 0.640 & \checkmark \\
    T7 & Mem0 & Claude Haiku & 400 & 0.535 & \checkmark \\
    T7 & Mem0 & Claude Sonnet & 400 & 0.663 & \checkmark \\
    T7 & A-Mem & DeepSeek & 400 & 0.651 & \checkmark \\
    T7 & A-Mem & GPT-4o-mini & 400 & 0.616 & \checkmark \\
    T7 & A-Mem & Claude Haiku & 400 & 0.442 & \checkmark \\
    T7 & A-Mem & Claude Sonnet & 400 & 0.616 & \checkmark \\
    T7 & structured timeline & DeepSeek & 400 & 0.640 & \checkmark \\
    T7 & structured timeline & GPT-4o-mini & 400 & 0.616 & \checkmark \\
    T7 & structured timeline & Claude Haiku & 400 & 0.512 & \checkmark \\
    T7 & structured timeline & Claude Sonnet & 400 & 0.686 & \checkmark \\
    T7 & dense retrieval & DeepSeek & 400 & 0.895 & \checkmark \\
    T7 & dense retrieval & GPT-4o-mini & 400 & 0.860 & \checkmark \\
    T7 & dense retrieval & Claude Haiku & 400 & 0.942 & \checkmark \\
    T7 & dense retrieval & Claude Sonnet & 400 & 0.930 & \checkmark \\
    T7 & full context & DeepSeek & 400 & 0.942 & \checkmark \\
    T7 & full context & GPT-4o-mini & 400 & 0.872 & \checkmark \\
    T7 & full context & Claude Haiku & 400 & 0.965 & \checkmark \\
    T7 & full context & Claude Sonnet & 400 & 0.953 & \checkmark \\
    \midrule
    T8 & no-context (blind) & DeepSeek & 134 & 0.231 & \checkmark \\
    T8 & no-context (blind) & GPT-4o-mini & 134 & 0.254 & \checkmark \\
    T8 & no-context (blind) & Claude Haiku & 134 & 0.172 & \checkmark \\
    T8 & no-context (blind) & Claude Sonnet & 134 & 0.261 & \checkmark \\
    T8 & last-visit only & DeepSeek & 134 & 0.313 & \checkmark \\
    T8 & last-visit only & GPT-4o-mini & 134 & 0.351 & \checkmark \\
    T8 & last-visit only & Claude Haiku & 134 & 0.179 & \checkmark \\
    T8 & last-visit only & Claude Sonnet & 134 & 0.231 & \checkmark \\
    T8 & LLM summary & DeepSeek & 134 & 0.231 & \checkmark \\
    T8 & LLM summary & GPT-4o-mini & 134 & 0.254 & \checkmark \\
    T8 & LLM summary & Claude Haiku & 134 & 0.194 & \checkmark \\
    T8 & LLM summary & Claude Sonnet & 134 & 0.216 & \checkmark \\
    T8 & Mem0 & DeepSeek & 134 & 0.373 & \checkmark \\
    T8 & Mem0 & GPT-4o-mini & 134 & 0.358 & \checkmark \\
    T8 & Mem0 & Claude Haiku & 134 & 0.306 & \checkmark \\
    T8 & Mem0 & Claude Sonnet & 134 & 0.299 & \checkmark \\
    T8 & A-Mem & DeepSeek & 134 & 0.373 & \checkmark \\
    T8 & A-Mem & GPT-4o-mini & 134 & 0.336 & \checkmark \\
    T8 & A-Mem & Claude Haiku & 134 & 0.291 & \checkmark \\
    T8 & A-Mem & Claude Sonnet & 134 & 0.373 & \checkmark \\
    T8 & structured timeline & DeepSeek & 134 & 0.448 & \checkmark \\
    T8 & structured timeline & GPT-4o-mini & 134 & 0.396 & \checkmark \\
    T8 & structured timeline & Claude Haiku & 134 & 0.537 & \checkmark \\
    T8 & structured timeline & Claude Sonnet & 134 & 0.493 & \checkmark \\
    T8 & dense retrieval & DeepSeek & 134 & 0.388 & \checkmark \\
    T8 & dense retrieval & GPT-4o-mini & 134 & 0.276 & \checkmark \\
    T8 & dense retrieval & Claude Haiku & 134 & 0.403 & \checkmark \\
    T8 & dense retrieval & Claude Sonnet & 134 & 0.396 & \checkmark \\
    T8 & full context & DeepSeek & 134 & 0.440 & \checkmark \\
    T8 & full context & GPT-4o-mini & 134 & 0.269 & \checkmark \\
    T8 & full context & Claude Haiku & 134 & 0.463 & \checkmark \\
    T8 & full context & Claude Sonnet & 134 & 0.470 & \checkmark \\
    \midrule
    T9 & no-context (blind) & DeepSeek & 1500 & 0.355 & \checkmark \\
    T9 & no-context (blind) & GPT-4o-mini & 1500 & 0.685 & \checkmark \\
    T9 & no-context (blind) & Claude Haiku & 1500 & 0.000 & \checkmark \\
    T9 & no-context (blind) & Claude Sonnet & 1500 & 0.204 & \checkmark \\
    T9 & last-visit only & DeepSeek & 1500 & 0.883 & \checkmark \\
    T9 & last-visit only & GPT-4o-mini & 1500 & 0.895 & \checkmark \\
    T9 & last-visit only & Claude Haiku & 1500 & 0.994 & \checkmark \\
    T9 & last-visit only & Claude Sonnet & 1500 & 0.926 & \checkmark \\
    T9 & LLM summary & DeepSeek & 1500 & 0.796 & \checkmark \\
    T9 & LLM summary & GPT-4o-mini & 1500 & 0.775 & \checkmark \\
    T9 & LLM summary & Claude Haiku & 1500 & 0.941 & \checkmark \\
    T9 & LLM summary & Claude Sonnet & 1500 & 0.744 & \checkmark \\
    T9 & Mem0 & DeepSeek & 1500 & 0.806 & \checkmark \\
    T9 & Mem0 & GPT-4o-mini & 1500 & 0.818 & \checkmark \\
    T9 & Mem0 & Claude Haiku & 1500 & 0.941 & \checkmark \\
    T9 & Mem0 & Claude Sonnet & 1500 & 0.725 & \checkmark \\
    T9 & A-Mem & DeepSeek & 1500 & 0.818 & \checkmark \\
    T9 & A-Mem & GPT-4o-mini & 1500 & 0.784 & \checkmark \\
    T9 & A-Mem & Claude Haiku & 1500 & 0.966 & \checkmark \\
    T9 & A-Mem & Claude Sonnet & 1500 & 0.731 & \checkmark \\
    T9 & structured timeline & DeepSeek & 1500 & 0.556 & \checkmark \\
    T9 & structured timeline & GPT-4o-mini & 1500 & 0.664 & \checkmark \\
    T9 & structured timeline & Claude Haiku & 1500 & 0.481 & \checkmark \\
    T9 & structured timeline & Claude Sonnet & 1500 & 0.438 & \checkmark \\
    T9 & dense retrieval & DeepSeek & 1500 & 0.772 & \checkmark \\
    T9 & dense retrieval & GPT-4o-mini & 1500 & 0.710 & \checkmark \\
    T9 & dense retrieval & Claude Haiku & 1500 & 0.920 & \checkmark \\
    T9 & dense retrieval & Claude Sonnet & 1500 & 0.790 & \checkmark \\
    T9 & full context & DeepSeek & 1500 & 0.701 & \checkmark \\
    T9 & full context & GPT-4o-mini & 1500 & 0.698 & \checkmark \\
    T9 & full context & Claude Haiku & 1500 & 0.907 & \checkmark \\
    T9 & full context & Claude Sonnet & 1500 & 0.747 & \checkmark \\
\end{longtable}
}

\twocolumn

\end{document}